%% file: online_robust_adapt.tex
\RequirePackage{fix-cm}
\documentclass[smallcondensed]{svjour3}     
\smartqed  
\usepackage{graphicx}
\usepackage{url}
\usepackage{natbib}

\newtheorem{assumption}{Assumption}

\usepackage{amssymb}
\usepackage{mathtools}
\mathtoolsset{showonlyrefs,showmanualtags}

\usepackage{bbm}

\usepackage{multirow}
\usepackage{booktabs}

\usepackage{longtable}

\usepackage{subcaption}
\usepackage{here}

\usepackage{algorithm}
\usepackage{algorithmic}

\usepackage{graphicx}
\usepackage{textcomp}
\def\BibTeX{{\rm B\kern-.05em{\sc i\kern-.025em b}\kern-.08em
    T\kern-.1667em\lower.7ex\hbox{E}\kern-.125emX}}

\newcommand\mydef{\mathrel{\overset{\makebox[0pt]{\mbox{\normalfont\tiny\sffamily def}}}{=}}}

\def\transpose#1{#1^{\top}}

\newcommand{\argmin}{\operatornamewithlimits{argmin}}
\def\d#1{\mathrm{d}#1}

\usepackage[T1]{fontenc}
\usepackage[utf8]{inputenc}

\begin{document}

\title{Online Robust and Adaptive Learning from Data Streams}


\author{Shintaro Fukushima  \and
        Atsushi Nitanda  \and
        Kenji Yamanishi
}


\institute{Shintaro Fukushima \at
              Graduate School of Information Science and Technology, The University of Tokyo \\
              7-3-1 Hongo, Bunkyo-ku, Tokyo, Japan \\ 
              TOYOTA MOTOR CORPORATION\\
              1-6-1 Otemachi, Chiyoda-ku, Tokyo, Japan \\
              \email{sfukushim@gmail.com \, 
                     s\_fukushima@mail.toyota.co.jp} 
           \and
           Atsushi Nitanda \at
           Faculty of Computer Science and Systems Engineering, 
           Kyushu Institute of Technology \\
           680-4 Kawazu, Iizuka-shi, Fukuoka, Japan \\
           \email{nitanda@ai.kyutech.ac.jp}
           \and 
           Kenji Yamanishi \at
           Graduate School of Information Science and Technology, The University of Tokyo \\
           7-3-1 Hongo, Bunkyo-ku, Tokyo, Japan \\
           \email{yamanishi@gcc.e.u-tokyo.ac.jp}
}

\date{Received: date / Accepted: date}

\maketitle

\begin{abstract}
In online learning from non-stationary data streams, 
it is necessary to learn robustly to outliers 
and to adapt quickly to changes in the underlying data generating mechanism. 
In this paper, we refer to the former attribute of online learning algorithms 
as robustness and to the latter as adaptivity. 
There is an obvious tradeoff between the two attributes. 
It is a fundamental issue to quantify and evaluate the tradeoff because it provides important information on the data generating mechanism. 
However, no previous work has considered the tradeoff quantitatively. 
We propose a novel algorithm called the stochastic approximation-based robustness-adaptivity algorithm (SRA) to evaluate the tradeoff. 
The key idea of SRA is to update parameters of distribution or sufficient statistics with the biased stochastic approximation scheme, 
while dropping data points with large values of the stochastic update. 
We address the relation between the two parameters: one is the step size of the stochastic approximation, 
and the other is the threshold parameter of the norm of the stochastic update. 
The former controls the adaptivity and the latter does the robustness. 
We give a theoretical analysis for the non-asymptotic convergence of SRA 
in the presence of outliers, 
which depends on both the step size and 
threshold parameter. 
Because SRA is formulated on the majorization-minimization principle, 
it is a general algorithm that includes many algorithms, 
such as the online EM algorithm and stochastic gradient descent. 
Empirical experiments for both synthetic and real datasets demonstrated 
that SRA was superior to previous methods. 
As SRA is based on assumptions 
that 
an abrupt change occurs in parameters 
or sufficient statistics 
of true distribution 
and 
that noisy distribution obeys the uniform distribution, 
it would be expected for SRA to be theoretically guaranteed in more general settings, 
such as incremental~(gradual) changes 
and other noisy distributions.  

\end{abstract}

\input{1_Introduction}

\input{2_Preliminaries}

\input{3_Proposed_Algorithm}

\input{4_Convergence_Analysis}

\input{5_Experiments}

\input{6_Conclusion}


%
%

\bibliographystyle{spbasic}      
\bibliography{online_robust_adapt}   


\newpage

\appendix

\input{A_Proofs}

\input{B_Dependency_on_Hyper_Parameters}

\end{document}

%% file: 1_Introduction.tex
\section{Introduction}

\subsection{Purpose of this paper}
\label{subsection:purpose_of_this_paper}

\begin{figure}[b]
\begin{center}
\includegraphics[width=\linewidth]{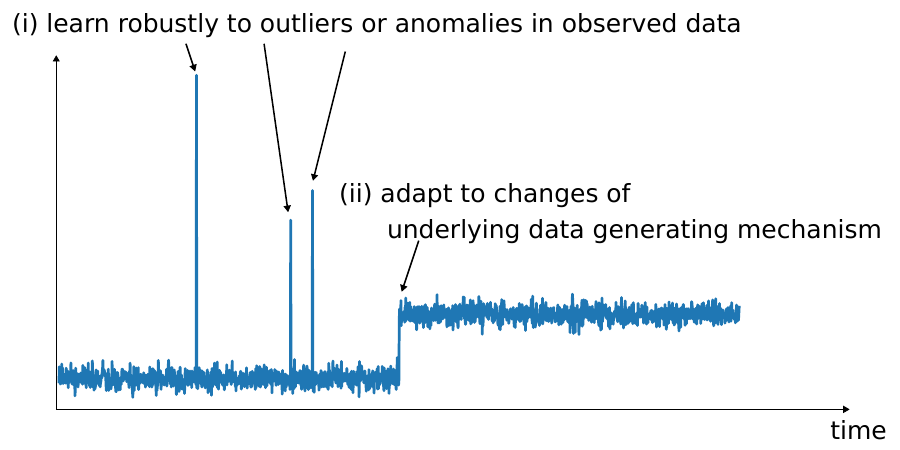}
\caption{Illustration of the concepts of the robustness and adaptivity of online learning algorithms.}
\label{fig:illustration_of_the_concepts_of_robustness_and_adaptivity_of_change_detection_algorithms}
\end{center}
\end{figure}

This study 
is concerned with online learning from data streams. 
We consider a situation where 
each datum arrives in an online fashion. 
In such a situation, 
we aim to 
(i) learn robustly to outliers or anomalies in the observed data. 
(ii) adapt to the changes in the underlying data-generating mechanism. 
In (i), if a data point is an outlier, 
we would like to learn with as little influence by the outlier as possible. 
In this paper, 
we refer to such an attribute of online learning algorithms 
as {\em robustness}. 
In contrast, 
with regard to (ii), 
it is desirable to adapt to the changes in the data-generating mechanism. 
We refer to such an attribute of online learning algorithms 
as {\em adaptivity}. 
Figure~\ref{fig:illustration_of_the_concepts_of_robustness_and_adaptivity_of_change_detection_algorithms} 
illustrates the concepts of the robustness and adaptivity. 

A tradeoff exists between the robustness and adaptivity: 
the robustness generally decreases if we try to adapt to the changes. 
Conversely, 
the adaptivity decreases if we try to learn robustly. 
Although many online learning algorithms 
have been introduced 
and some studies have addressed this issue \citep{Tsay1988,Gama2014,Chu2004,Huang2016,Odakura2018,Cejnek2018,Fearnhead2019,Guo2019}, 
to the best of our knowledge, 
no algorithm has quantitatively considered 
the tradeoff between the robustness and adaptivity. 

This study proposes 
an online learning algorithm 
that considers the tradeoff between the robustness and adaptivity. 
We introduce a novel algorithm, called 
the stochastic approximation-based robustness--adaptivity (SRA) algorithm, 
to provide a theoretical analysis for non-asymptotic convergence of 
SRA in the presence of outliers 
and to demonstrate its effectiveness for both synthetic and real datasets. 
The key idea of SRA is 
to update parameters of distribution or sufficient statistics 
with the stochastic approximation (SA) \citep{Robbins1951} 
while dropping points with large values of stochastic updates (drift terms).

\subsection{Related work}
\label{subsection:related_work}

This study is concerned with the robustness and adaptivity of 
online learning algorithms. Moreover, 
we briefly review 
studies related to the SA \citep{Robbins1951} and 
the online expectation--maximization (EM) algorithm \citep{Cappe2009,Karimi2019colt} 
because SRA uses both of them. 

\subsubsection{Robustness and adaptivity of online learning algorithms} 

The robustness and adaptivity of online learning algorithms 
have often been discussed 
in the context of the concept drift \citep{Gama2014,Chu2004,Huang2016,Cejnek2018}. 
Yamanishi et al. proposed 
an online learning algorithm, 
called the sequentially discounting EM algorithm (SDEM) \citep{Yamanishi2004}. 
Although SDEM can handle complicated distributions, 
it is prone to noise and can easily overfit the data. 
Odakura 
proposed an online nonstationary robust learning algorithm 
\citep{Odakura2018}. 
This algorithm independently introduces two parameters to control the robustness and adaptivity, respectively. 
Fearnhead and Rigaill 
proposed an algorithm for change detection 
that is robust in the presence of outliers \citep{Fearnhead2019}. 
The key idea of the algorithm is to 
adapt existing penalized cost approaches 
to detect changes such that the loss function 
is less sensitive to outliers. 
Guo proposed an algorithm 
based on an online sequential extreme learning machine 
for robust and adaptive learning \citep{Guo2019}.

\subsubsection{Online (stochastic) EM algorithms }

The EM algorithm \citep{Dempster1977} is a popular class of inference that minimizes loss function. 
The original EM algorithm  
does not scale to a large dataset 
because it requires the entire data at each iteration. 
To overcome this problem, 
several studies proposed online versions of 
the EM algorithm. 

Neal and Hinton  
proposed an EM algorithm in an incremental scheme 
referred to as the incremental EM (iEM) \citep{Neal1999}. 
Capp\'{e} and Moulines  
proposed the stochastic (online) EM (sEM) algorithm \citep{Cappe2009}, 
which updates the sufficient statistics 
in an SA scheme \citep{Robbins1951}. 
Chen et al. 
proposed the variance reduced sEM (sEM-VR) algorithm \citep{Chen2018}. 
Meanwhile, Karimi et al. showed 
non-asymptotic convergence bounds for the global convergence 
of iEM, sEM-VR, and the fast incremental EM \citep{Karimi2019neurips}. 

By contrast, 
only a few studies have considered 
the online EM algorithm 
in a situation where a fresh sample is drawn at each iteration. 
Capp\'{e} and Moulines 
proved the asymptotic convergence of 
the online EM algorithm \citep{Cappe2009}. 
Balakrishnan et al. 
analyzed the non-asymptotic convergence for a variant of the online EM algorithm \citep{Balakrishnan2017}, 
where the initial radius around the optimal parameter must 
be known in advance. 
Karimi et al. considered 
the SA scheme \citep{Robbins1951}, 
the stochastic update (drift term) of which depends on a state-dependent Markov chain. Moreover, the mean field is not necessarily of a gradient type, 
thereby covering an approximate second-order method and 
allowing an asymptotic bias for one-step updates \citep{Karimi2019colt}.  
They illustrated these settings using the online EM algorithm 
and the policy-gradient method for the average reward maximization 
in reinforcement learning.

\subsection{Significance of this paper}

In the context of Section~\ref{subsection:purpose_of_this_paper} and 
\ref{subsection:related_work}, 
the contributions of this paper are summarized below.

\subsubsection{Novel online learning algorithm 
               for tradeoff between robustness and adaptivity}

We propose a novel online learning algorithm, called SRA,
to consider the tradeoff between the robustness and adaptivity. 
Previous studies \citep{Chu2004,Huang2016,Cejnek2018,Yamanishi2004,Odakura2018,Fearnhead2019,Guo2019} 
considered only one of them, 
and even when both were considered, 
the relation between them was not clarified. 
This study considers both the robustness and adaptivity, 
and gives a theoretical analysis for the non-asymptotic convergence of SRA. 
To do so, 
we adopt the SA scheme \citep{Robbins1951} 
in a setting where outliers and change points may exist. 
As SRA is formulated on the majorization--minimization principle \citep{Lange2016,Mairal2015}, 
it is a general algorithm that includes many schemes, 
such as the online EM algorithm \citep{Cappe2009,Balakrishnan2017,Karimi2019colt} and stochastic gradient descent (SGD). 
Our approach is considered to be an extension of the work of \citep{Karimi2019colt}, 
but they presented convergence analysis of 
the biased SA in the absence of outliers and change points. 
By contrast, 
we consider convergence analysis in a setting where 
outliers and change points may exist. 
Our study is novel in that we show non-asymptotic convergence analysis in this broader setting and 
apply it to quantify and evaluate the tradeoff between the robustness and adaptivity of online learning algorithms. 

We present a detailed comparison 
between one of the promising 
previous studies and this study, 
to clarify the advantages of this study 
over previous ones. 
Fearnhead and Rigaill proposed 
a promising algorithm called F-RPOP 
for change detection in the presence of 
outliers~\citep{Fearnhead2019}. 
The key idea of F-RPOP is to find 
the optimal segmentation~(change points) 
of a data stream with 
dynamic programming 
under penalized cost criteria. 
More specifically, 
the authors defined the cost of a segment 
as the sum of losses 
at the time points in the segment 
using a loss function, 
and then found the optimal segmentation~(change points) 
by minimizing the cost with 
dynamic programming. 
The loss function has a segment-specific location parameter $\theta$. 
Some loss functions also have parameter $K$, 
that is, 
the tolerance threshold of distance 
between a data point and $\theta$. 
The functions include 
the Huber loss and the biweight loss. 
Therefore, 
$K$ controls the robustness. 
The authors also introduced a parameter 
$\beta$ in dynamic programming, 
which affects the number of change points. 
Therefore,  
$\beta$ controls the adaptivity. 
However, 
the relation between $K$ and 
$\beta$ is not 
clarified in \citep{Fearnhead2019} 
except one on the lengths of segments. 
Although empirical studies 
show good results with the robustness and adaptivity, 
it is necessary to tune the two parameters $K$ and $\beta$ separately 
without the knowledge of the relation 
on the robustness and adaptivity 
between them. 
By contrast, this
study quantitatively evaluates the tradeoff between the robustness and adaptivity. 
From an empirical point of view, 
this evaluation leads to the relation 
between 
two parameters that control the robustness 
and adaptivity. 
When a value of one parameter is given, 
the value of the other parameter is determined 
theoretically. 

Note that many studies already addressed the tradeoff between 
exploration and exploitation in bandit algorithms (e.g., \citep{Lattimore2018}). 
However, our problem setting is different from those in these studies. 
Bandit algorithms search for parameters 
independently of changes in the environment. 
In contrast, 
our SRA does not greatly change parameters 
when the change in the data-generating mechanism is moderate. 
It adapts to the changes of the data-generating mechanism. 
Therefore, 
although both our study and those concerned with bandit algorithms 
consider the tradeoff between global and local information, 
our motivation is different from that in other studies.

\subsubsection{Empirical demonstration of the proposed algorithm}

We evaluated the effectiveness of SRA 
on both synthetic and real datasets. 
We empirically showed characteristics of SRA 
by inspecting the dependencies on the parameters of SRA; 
these were consistent with those of the theoretical analysis. 
We also compared the performance of SRA 
with those of the previously proposed 
online learning algorithms~
\citep{Neal1999,Yamanishi2004,Cappe2009} 
and 
concep drift detection algorithms~
\citep{Bifet2007,Raab2020,Page1954}
, 
on important tasks, 
including change detection and anomaly detection. 
It was determined that SRA was superior to other algorithms.

%% file: 2_Preliminaries.tex
\section{Preliminaries}

In this section, 
we provide our problem setting 
and 
an important theoretical result of previous study: non-asymptotic convergence of SA~\cite{Karimi2019colt}. 

\subsection{Problem setting}
\label{subsection:problem_setting}

We consider a situation where each datum $y_{t} \in \mathbb{R}^{d}$ 
arrives in an online fashion at each time $t \in \mathbb{N}$. 
If no noise exists, 
we assume that $y_{t}$ is drawn from 
\begin{align}
y_{t} \sim f(y_{t}; \theta_{t}), 
\label{eq:ideal_data_generation}
\end{align}
where 
$f \in \mathcal{F}$  
is an element of 
a parametric class of distribution 
$\mathcal{F} = \{ f(y; \theta), \,
 \theta \in \Theta \}$, 
$\theta$ is a parameter, 
and 
$\Theta \subset \mathbb{R}^{p}$ is a parameter space associated. 
However, 
in the real world, 
data are sometimes contaminated by noise. 
In this case, 
we assume that $y_{t}$ is drawn from 
a mixture of probability density functions: 
\begin{align}
y_{t} \sim  \alpha f(y_{t}; \theta_{t}) 
            + (1-\alpha) f_{ \mathrm{noise} }(y_{t}; \xi ), 
\label{eq:real_data_generation}
\end{align}
where $\alpha$ denotes the mixture ratio ($0 < \alpha < 1$). 
Equation~\eqref{eq:real_data_generation} 
means that a datum is generated from a \textit{true} distribution 
with probability $\alpha$ 
and from a \textit{noisy} distribution with probability $1-\alpha$. 
$f_{\mathrm{noise}}$ is an element of 
a parametric class of data distributions 
$\mathcal{G} = \{ f_{\mathrm{noise}}(y; \xi), \, \xi \in \Xi \}$, 
where 
$\xi$ is a parameter 
and 
$\Xi \subset \mathbb{R}^{m}$ is a parameter space associated. 
This study 
addresses the convergence property of 
Equation~\eqref{eq:real_data_generation}. 

We assume that a change point $t^{\ast}$ is given, 
and each datum before and after the change point 
is drawn from different distributions. 
This means that 
$\theta_{t}$ in Equation~\eqref{eq:real_data_generation} varies as follows: 
\begin{align}
\theta_{t} = 
  \left\{
    \begin{array}{ll}
    \theta^{1} & (t=1, \dots, t^{\ast}-1), \\
    \theta^{2} & (t=t^{\ast}, \dots), 
    \end{array}
  \right.
\label{eq:real_data_abrupt_change}
\end{align}
where $\theta^{1} \neq \theta^{2}$. 
This implies a change abruptly occurs at $t^{\ast}$. 

\subsection{Non-asymptotic analysis of SA}
\label{subsection:nonasymptotic_analysis_of_SA}

Karimi et al. showed 
a convergence analysis \citep{Karimi2019colt} of 
the non-convex objective function
under the SA scheme 
\citep{Robbins1951} 
in Equation~\eqref{eq:ideal_data_generation}: 
\begin{align}
\theta_{t+1} = 
  \theta_{t} - \rho_{t+1} H_{ \theta_{t}}(Y_{t+1}), 
\label{eq:SA_scheme}
\end{align}
where $\theta_{t} \in \Theta \subset \mathbb{R}^{p}$ 
denotes the $t$-th iterate of parameters 
or the sufficient statistics of the distribution, 
$\rho_{t+1}$ is the step size. 
$Y_{t+1}$ denotes the random variable at $t+1$, 
and $y_{t+1}$ does its realization.  
$H_{ \theta_{t}}(Y_{t+1})$ is the 
stochastic update at time $t$. 
When $\{ y_{t} \}_{t=1}^{\infty}$ is an i.i.d. sequence of 
random vectors,
the mean field for the SA is defined as 
$
h(\theta_{t}) = \mathbb{E}[ 
                  H_{\theta_{t}}(Y_{t+1}) | 
                  \mathcal{F}_{t}
                ]
$,  
where $\mathcal{F}_{t}$ is the filtration 
generated by the random variables $( \theta_{0}, \{Y_{s} \}_{s=1}^{t} )$, 
at time $t$. 
When $\{ y_{t} \}_{t=1}^{\infty}$ is a state-dependent Markov chain, 
$h(\theta_{t}) = \int H_{\theta_{t}}(y) \, \pi_{\theta_{t}}(\d{y})$ 
under the assumption that 
$\int \| H_{\theta_{t}}(y) \| \, 
 \pi_{\theta_{t}}(\d{y}) < \infty$, 
where $\| \cdot \|$ denotes the norm of the vector in $\mathbb{R}^{p}$ 
and 
$\pi = \pi_{\theta}(y)$ is the true distribution. 
In this study, 
we consider the former case, 
that is, 
$\{y_{t}\}_{t=1}^{\infty}$ 
is an i.i.d sequence of random vectors. 
Karimi et al. assumed that 
$h$ is related to a smooth Lyapunov function 
$V : \mathbb{R}^{p} \rightarrow \mathbb{R}$, 
where $V(\theta) > -\infty$.  
This SA scheme in Equation~\eqref{eq:SA_scheme} aims to 
find a minimizer or a stationary point of the 
non-convex Lyapunov function $V$. 

For example, 
let us consider the online EM algorithm \citep{Cappe2009,Karimi2019colt} 
to the curved exponential family:
\begin{align}
f(Y, Z; \theta) = h(Y, Z) \exp{ 
                 \left( \langle S(Y, Z) | \phi(\theta) \rangle 
                   -\psi(\theta)
                 \right)
               }. 
\label{eq:curved_exponential_family}
\end{align}
Here, 
$\psi: \Theta \rightarrow \mathbb{R}$ is twice differentiable 
and convex. 
$\phi: \Theta \rightarrow \mathrm{S} \subset \mathbb{R}^{p}$ is 
concave and differentiable. 
$\mathrm{S}$ is a convex open subset of $\mathbb{R}^{p}$, 
$S$ denotes the sufficient statistics, 
and $\langle \cdot | \cdot \rangle$ denotes dot product. 
The Lyapunov function $V(s)$ is defined for the sufficient statistics $s$ as 
\begin{align}
V(s) \mydef 
  \mathrm{KL} (\pi, g(\cdot; \bar{\theta}(s)))
  + R( \bar{\theta}(s) ), 
\label{eq:online_em_lyapunov_function}
\end{align}
where $\mathrm{KL}$ is Kullback--Leibler (KL) divergence
between $\pi$ and $g_{\theta}$ defined as 
\begin{align}
\mathrm{KL} (\pi, g) &\mydef
  \mathbb{E}_{\pi} [ \log{ ( \pi(Y) / g(Y; \theta) ) } ], 
\end{align}
and $R: \Theta \rightarrow \mathbb{R}$ is 
a penalization term assumed to be twice differentiable \citep{Karimi2019colt}. 
$\bar{\theta}$ in Equation~\eqref{eq:online_em_lyapunov_function} is defined as 
the minimizer of the following loss function:  
\begin{align}
\ell(s; \theta) 
= \psi(\theta) + R(\theta) - \langle s | \phi(\theta) \rangle. 
\label{eq:loss_function_curved_exponential_family}
\end{align}
Therefore, $\bar{\theta}(s)$ is represented as 
\begin{align}
\bar{\theta}(s) 
  \mydef \argmin_{\theta} \, \ell(s; \theta)
  = \argmin_{\theta} \, 
    \left\{ 
      \psi(\theta) + R(\theta) - \langle s | \phi(\theta) \rangle
    \right\}. 
\label{eq:def_theta_bar}
\end{align}
Karimi et al. considered 
the following assumptions for $h$ and $V$.  
\begin{assumption} \citep{Karimi2019colt}
\begin{description}
\item [(a)]
$\forall \theta \in \Theta$, 
$\exists c_{0} \geq 0, \, c_{1} > 0$, 
s.t. $c_{0} + c_{1} \langle \nabla V(\theta) | h(\theta) \rangle 
      \geq \| h(\theta) \|^{2}$. 
\item [(b)]
$\forall \theta \in \Theta$, 
$\exists d_{0} > 0, d_{1} > 0$,  
s.t. $d_{0} + d_{1} \| h(\theta) \| \geq \| \nabla V(\theta) \|$. 
\item [(c)] 
The Lyapunov function $V$ is L-smooth: 
$\forall (\theta, \theta') \in \Theta^{2}, 
 \| \nabla V(\theta) - \nabla V(\theta') \| \leq 
 L \| \theta - \theta' \|$. 
\end{description}
\label{assumption:assumptions_of_Lyapunov_function}
\end{assumption}
Here, 
$\| h(\theta) \|$ denotes the norm of the mean field 
which takes on small values 
as the SA scheme in Equation~\eqref{eq:SA_scheme} converges. 
Assumption~\ref{assumption:assumptions_of_Lyapunov_function} (a) and (b) 
assume that the mean field $h(\theta)$ is 
indirectly related to the Lyapunov function $V(\theta)$, 
but it is not necessarily the same as $\nabla V(\theta)$. 
The constants $c_{0}$ and $d_{0}$ characterize the bias 
between the mean field and the gradient of the Lyapunov function. 
We note that the Lyapunov function $V$ can be a non-convex function under 
Assumption~\ref{assumption:assumptions_of_Lyapunov_function} (c).  

For any $n \geq 1$, 
we denote $N \in \{0, \dots, n\}$ 
as a discrete random variable 
independent of $ \{ \mathcal{F}_{n} \}_{n=1}^{\infty} $. 
When we adopt a randomized stopping rule in SA 
as in \citep{Ghadimi2013}, 
we define
$ P(N = \ell) 
\mydef 
  \rho_{\ell+1} / \sum_{k=0}^{n} \rho_{k+1}
$, 
where $N$ is the terminating iteration 
for Equation~\eqref{eq:SA_scheme}. 
We consider the following expectation:
\begin{align}
\mathbb{E} [ \| h(\theta_{N}) \|^{2} ]
= \sum_{k=1}^{n} P(N=k) \| h(\theta_{k}) \|^{2},
\label{eq:expectation_of_mean_field}
\end{align}
where $\theta_{k}$ is solved with Equation~\eqref{eq:SA_scheme}. 
The left side of Equation~\eqref{eq:expectation_of_mean_field} indicates the expectation of the norm of the mean field $h(\theta)$ 
when we consider the weights of the data points.

We then define the following noise vector:
\begin{align}
e_{t+1} \mydef H_{ \theta_{t} }( Y_{t+1} ) - h(\theta_{t}). 
\label{eq:def_noise_vector}
\end{align}
Equation~\eqref{eq:def_noise_vector} represents the difference between the stochastic update and mean field at time $t+1$. 

We assume the following:
\begin{assumption}
\citep{Karimi2019colt} 
The noise vectors have 
a Martingale difference sequence 
for any $t \in \mathbb{N}$, 
$\mathbb{E}[ e_{t+1} | \mathcal{F}_{t} ] = 0$, 
$\mathbb{E}[ \| e_{t+1} \|^{2} | \mathcal{F}_{t} ]
 \leq \sigma_{0}^{2} + 
      \sigma_{1}^{2} \| h(\theta_{t}) \|^{2}$ 
with $\sigma_{0}^{2}, \sigma_{1}^{2} \in [0, \infty)$. 
\label{assumption:martingale_difference_sequence}
\end{assumption}

The following theorem then holds: 
\begin{theorem} \citep{Karimi2019colt} 
If Assumption~\ref{assumption:assumptions_of_Lyapunov_function} (a), (c) 
and 
Assumption~\ref{assumption:martingale_difference_sequence} 
hold, 
and 
$\rho_{t+1} \leq 1/(2 c_{1} (1+\sigma_{1}^{2}) )$ 
for all $t \geq 0$, 
then we obtain the following inequality: 
\begin{align}
\mathbb{E} [ \| h(\theta_{N}) \|^{2} ]
\leq
\frac{ 2 c_{1} ( 
		 V_{0, n} + 
		 \sigma_{0}^{2} L 
		 \sum_{t=0}^{n} \rho_{t+1}^{2}
	   )
}{ \sum_{t=0}^{n} \rho_{t+1} } 
+ 2 c_{0},
\label{eq:upper_bound_karimi2019colt}
\end{align} 
where 
$V_{0, n} \mydef  
   \mathbb{E} [ V(\theta_{0}) - V(\theta_{n+1}) | \mathcal{F}_{n} ]$. 
\label{theorem:upper_bound_expectation_of_variance_of_mean_field}
\end{theorem}
In particular, 
when we set $\rho_{t} = 1/(2 c_{1} L(1 + \sigma_{0}^{2})\sqrt{t})$, 
the right-hand side of Equation~\eqref{eq:upper_bound_karimi2019colt} evaluates to $O(c_{0} + \log{n}/n)$. 
This means that the SA scheme in 
Equation \eqref{eq:SA_scheme} 
finds an $O(c_{0} + \log{n}/n)$ stationary point within $n$ iterations. 
Note that $c_{0}$ is the inevitable bias 
between the mean field $h(\theta)$ 
and the gradient of the Lyapunov function $\nabla V(\theta)$. 

%% file: 3_Proposed_Algorithm.tex
\section{Proposed algorithm}
\label{section:proposed_algorithm}

In this section, 
we introduce an online learning algorithm from data streams, 
called the 
SRA, 
to consider the tradeoff between the robustness and adaptivity. 
First, 
we describe SRA in Section~\ref{subsection:SRA} 
and its application to the online EM algorithm \citep{Cappe2009,Karimi2019colt} 
in Section~\ref{subsection:application_of_SRA_to_online_EM}. 
Because SRA is formulated on the majorization--minimization principle (e.g., \citep{Lange2016,Mairal2015}), 
it is widely applicable to a broad class of algorithms, 
such as SGD (e.g., \citep{Bottou2018}). 
We explain this point in Section~\ref{subsection:surrogate_functions_of_SRA}. 
The notations follows these in
Section~\ref{subsection:nonasymptotic_analysis_of_SA}, 
unless specifically defined. 

\subsection{SRA}
\label{subsection:SRA}

We consider the convergence property of Equation~\eqref{eq:real_data_generation} 
under the following SA scheme: 
\begin{align}
\theta_{t+1} &= \theta_{t} - \rho_{t+1} G_{\theta_{t}}(Y_{t+1}), 
\label{eq:truncated_SA_scheme}
\end{align}
where $\rho_{t+1}$ is the step size, 
as in Equation~\eqref{eq:SA_scheme}, 
and $G_{\theta_{t}}$ is defined 
for a given $\gamma > 0$ 
as 
\begin{align}
G_{\theta_{t}}(Y) &= 
  \begin{cases}
  H_{\theta_{t}}(Y) & ( \| H_{\theta_{t}}(Y) \| \leq \gamma), \\
  0 & ( \| H_{\theta_{t}}(Y) \| > \gamma). 
  \end{cases}
\label{eq:def_G}
\end{align}
We call the SA scheme in Equation~\eqref{eq:truncated_SA_scheme}
SRA,  
which is summarized in Algorithm~\ref{algorithm:SRA}. The computational cost of SRA is $O(1)$ at each time. 

\begin{algorithm}[h]
\caption{Stochastic approximation-based robustness--adaptivity algorithm (SRA)}
\label{algorithm:SRA}
\begin{algorithmic}[1]
\REQUIRE 
$\{ \rho_{t} \}$: step sizes for the SA scheme ($\rho_{t} > 0$). 
$\gamma$: threshold parameter for stochastic update ($\gamma > 0$). 
\STATE Initialize the parameters or the sufficient statistics $\theta$. 
\FOR {$t=1, \, \dots$}
  \STATE Receive $y_{t}$. 
  \STATE Calculate the stochastic update $ H_{\theta_{t-1}}(y_{t}) $. 
  \STATE Update the parameters or the sufficient statistics 
         with SA in \eqref{eq:truncated_SA_scheme} and \eqref{eq:def_G}. 
\ENDFOR
\end{algorithmic}
\end{algorithm}

Equation~\eqref{eq:truncated_SA_scheme} is 
different from 
Equation~\eqref{eq:SA_scheme} 
in that 
Equation~\eqref{eq:truncated_SA_scheme} 
does not update the parameters of the distribution 
or the sufficient statistics 
when $\| H_{\theta_{t}}(Y) \| > \gamma$. 
This means that 
SRA drops data points with large values of 
stochastic updates $H_{\theta_{t}}(Y_{t+1})$ 
and 
updates the parameters of the distribution or 
the sufficient statistics with SA. 
The former corresponds to the robustness, 
whereas the latter corresponds to the adaptivity of SRA. 
They are controlled by 
threshold parameter $\gamma$ 
and 
the step sizes $\{ \rho_{t} \}$, 
respectively. 
The step size is sometimes referred to 
as the discounting parameter (e.g., \citep{Yamanishi2004}).
Although the step size of the SA is generally different 
from the discounting parameter, 
it is related to the adaptivity with respect to introducing effects of new samples. 
The step size, in particular, introduces high adaptivity 
when the decrease rate is relatively small. 
Therefore, it is sufficient to discuss the step size 
with respect to adaptivity in the SA setting. 
The relation between $\{ \rho_{t} \}$ and $\gamma$, 
and the determination of the optimal values of $\{ \rho_{t} \}$ with $\gamma$ are 
addressed in Section~\ref{section:convergence_analysis_of_bounded_stochastic_update}. 
The former procedure of SRA is somewhat similar to 
the one in \citep{Hara2019}, 
while they inspected influential instances 
for models trained with SGD.

\subsection{Application to the online EM algorithm}
\label{subsection:application_of_SRA_to_online_EM}

Next, 
we consider SRA in the online EM setting \citep{Cappe2009}. 
The SA with the online EM algorithm is described as 
\begin{align}
\operatorname{E-step}: \hat{s}_{t+1}
&= \hat{s}_{t} - \rho_{t+1} ( \hat{s}_{t} - \bar{s}(Y_{t+1}; \hat{\theta}_{t}) ), 
\label{eq:SA_regularized_online_em_algorithm_estep} \\
\operatorname{M-step}: \hat{\theta}_{t+1} &= \bar{\theta}( \hat{s}_{t+1} ), 
\label{eq:SA_regularized_online_em_algorithm_mstep}
\end{align}
where $\hat{s}_{t}$ denotes estimated sufficient statistics at $t$. 
The E-step of the online EM algorithm updates the sufficient statistics, 
whereas the M-step updates the parameters. 
$\bar{s}(y; \theta)$ in Equation~\eqref{eq:SA_regularized_online_em_algorithm_estep} 
is defined as 
\begin{align}
\bar{s}(y; \theta)
\mydef 
 \mathbb{E}_{\theta} [ s(Y=y, Z) | Y=y ], 
\end{align}
where $Y$ and $Z$ are the observed and latent variables, respectively, 
and $s(Y, Z) \in \mathrm{S}$ denotes the complete-data sufficient statistics. 
We consider the curved exponential family in Equation~\eqref{eq:curved_exponential_family}. 
The negated complete data loglikelihood of 
Equation~\eqref{eq:curved_exponential_family} is defined in 
Equation~\eqref{eq:loss_function_curved_exponential_family}. 
In addition, $\bar{\theta}(s)$ in Equation~\eqref{eq:SA_regularized_online_em_algorithm_mstep} 
is defined in Equation~\eqref{eq:def_theta_bar}. 
Accordingly, 
Equation~\eqref{eq:truncated_SA_scheme} 
, 
\eqref{eq:SA_regularized_online_em_algorithm_estep}, 
and 
\eqref{eq:SA_regularized_online_em_algorithm_mstep} show 
that 
the stochastic update $H$ and 
its mean field $h$ are represented by 
\begin{align}
H_{ \hat{s}_{n} }( Y_{n+1} )
&= \hat{s}_{n} - \bar{s}( Y_{n+1}; \bar{\theta}(\hat{s}_{n}) ),
\label{eq:online_em_H} \\
h(\hat{s}_{n})
&= \mathbb{E}_{\pi} [ 
     H_{ \hat{s}_{n} } (Y_{n+1}) | \mathcal{F}_{n} 
   ] 
 = \hat{s}_{n} - 
   \mathbb{E}_{\pi} [ 
     \bar{s}( Y_{n+1}; \bar{\theta}( \hat{s}_{n} ) ) 
   ].
\end{align}
We use Equation~\eqref{eq:online_em_H} in 
Equation~\eqref{eq:def_G}. 
Please refer to \citep{Karimi2019colt} for application to the Gaussian mixture model (GMM).

\subsection{Surrogate functions of SRA}
\label{subsection:surrogate_functions_of_SRA}

Because SRA is formulated on the majorization--minimization principle (e.g., \citep{Lange2016,Mairal2015}), 
it is naturally applicable to a wider class of algorithms, 
such as SGD. 
For example, stochastic optimization with $L_{2}$-regularizer  
is described as
\begin{align}
\theta_{t+1} = 
  \argmin_{\theta}
    \left\{ 
    -\rho_{t+1}
      \langle
	\nabla \ell(\theta), \theta - \theta_{t} 
      \rangle
    +\frac{1}{2} \| \theta - \theta_{t} \|^{2}
    +\frac{\rho_{t+1}}{2} \lambda \| \theta_{t} \|^{2}
  \right\}
\, 
(t=1, \dots), 
\label{eq:stochastic_optimization_with_L2_regularizer}
\end{align}
where $\ell$ is a loss function, 
$\rho_{t+1} > 0$ is the learning rate, 
and $\lambda > 0$ is a penalty parameter. 
We obtain the solution of Equation~\eqref{eq:stochastic_optimization_with_L2_regularizer} as 
\begin{alignat}{2}
&&-(\theta_{t+1} - \theta_{t}) 
 &= \rho_{t+1} \nabla \ell (\theta_{t}) +
   \rho_{t+1} \lambda \theta_{t}, \\
&\Longleftrightarrow \,
&\theta_{t+1} &= (1 - \rho_{t+1} \lambda) \theta_{t} -
             \rho_{t+1} \nabla \ell (\theta_{t}), \\
&\Longleftrightarrow \, 
&\theta_{t+1} &= \theta_{t} - 
             \rho_{t+1} ( \lambda \theta_{t} + 
                   \nabla \ell(\theta_{t} ) ). 
\label{eq:stochastic_gradient_update}
\end{alignat}
The final equation in Equation~\eqref{eq:stochastic_gradient_update} 
corresponds to Equation~\eqref{eq:truncated_SA_scheme}, 
where 
$H_{\theta_{t}}(y_{t+1}) = 
     \lambda \theta_{t} + \nabla \ell(\theta_{t})$. 
Please refer to \citep{Ghadimi2013,Bottou2018} for details on stochastic optimization in the SA scheme.

%% file: 4_Convergence_Analysis.tex
\section{Convergence analysis} 
\label{section:convergence_analysis_of_bounded_stochastic_update} 

In this section, 
we present the convergence analysis of SRA. 
All the proofs are given in the Appendix~\ref{section:proofs_in_online_robust_and_adaptive_learning_from_data_streams}. 

\subsection{Upper bound of expectation of the mean field} 

We investigate
the convergence of Equation~\eqref{eq:truncated_SA_scheme}. 
In particular, 
our concern is on 
how Theorem~\ref{theorem:upper_bound_expectation_of_variance_of_mean_field} 
would be altered when each datum is generated from Equation~\eqref{eq:real_data_generation} 
instead of Equation~\eqref{eq:ideal_data_generation}. 
In this case, 
we define the following noise vector:
\begin{align}
\xi_{t+1} \mydef 
  G_{\theta_{t}}(Y_{t+1}) - h(\theta_{t}). 
\label{eq:def_noise_vector_biased}
\end{align}

We then address the convergence property of 
$\mathbb{E} [ \| h (\theta_{N}) \|^{2} ]$ 
under Equation~\eqref{eq:truncated_SA_scheme}, 
where $N \in \{ 0, \dots, n \}$ denotes a discrete random variable 
for any $n \geq 1$, 
and the expectation is calculated 
from Equation~\eqref{eq:expectation_of_mean_field} 
as in Section~\ref{subsection:nonasymptotic_analysis_of_SA}. 

The following lemma holds
with respect to the expectation of the dot product of
the gradient of the Lyapunov function
and the noise vector.
\begin{lemma} 
There exists $M > 0$, 
such that the following inequality holds for $k=0, \dots, n$: 
\begin{align}
\mathbb{E} [ 
  -\langle
    \nabla V(\theta_{k}) | \xi_{k+1} 
  \rangle
  | \mathcal{F}_{k}
] 
&\leq 
\| \nabla V(\theta_{k}) \|
  \int_{\gamma}^{\infty} 
    \exp{ \left( -\frac{ z^{2} }{ M^{2} } \right) } \, \d{z}.
\label{eq:inequality_expectation_lyapunov_function_gradient_noise_vector}
\end{align}
\label{lemma:inequality_expectation_lyapunov_function_gradient_noise_vector}
\end{lemma}
The proof of Lemma~\ref{lemma:inequality_expectation_lyapunov_function_gradient_noise_vector} is given in Appendix~\ref{subsection:proof_lemma_inequality_expectation_lyapunov_function_gradient_noise_vector}.  
The left-hand side of 
Equation~\eqref{eq:inequality_expectation_lyapunov_function_gradient_noise_vector}
represents the magnitude of the bias of $\gamma$. 
In contrast, 
on the right-hand side of 
Equation~\eqref{eq:inequality_expectation_lyapunov_function_gradient_noise_vector}, 
the sharper the distribution of $H$ is, 
the smaller $M$ becomes. 
As a result, 
the bound is improved. 
We address this point in the case where $H_{\theta_{k}}$ is bounded in the discussion of Corollary~\ref{corollary:difference_between_upper_bounds}.

We make the following assumption for the noise distribution $f_{\mathrm{noise}}$ in Equation~\eqref{eq:real_data_generation}:
\begin{assumption}
We assume that 
the noise distribution $f_{\mathrm{noise}}$ 
in Equation~\eqref{eq:real_data_generation} obeys 
the uniform distribution:
\begin{align}
f_{\mathrm{noise}}(y_{t}; \xi)
&= 1/(2U)^{d}, 
\end{align}
where $y_{t} \in [-U, U]^{d}$, 
$U \in \mathbb{R}$, 
and $d$ is the dimension of data. 
\label{assumption:noise_uniform_distribution}
\end{assumption}

Note that Assumption~\ref{assumption:noise_uniform_distribution} 
affects the results of the convergence analysis. 
In particular, 
we obtain the difference in the upper bounds 
by setting the threshold parameter $\gamma$ to be 
proportional to $dU^{2} - \gamma^{2}$ 
in Corollary~\ref{corollary:difference_between_upper_bounds} 
under certain assumptions. 

Because $U$ cannot be determined in advance, 
it should be carefully selected. 
However, 
we also note that 
$U$ does not affect the choice of $\rho$ 
in Equation~\eqref{eq:estimated_rho_by_gamma} 
in Corollary~\ref{corollary:rho_determined_by_gamma} 
if $\gamma < \sqrt{d} U$. 
In contrast,  
the convergence analysis 
in Equation~\eqref{eq:expectation_mean_field_upper_bound} 
is affected by $U$. 

Therefore, the following lemma holds: 
\begin{lemma}
If we consider Assumption~\ref{assumption:noise_uniform_distribution} 
and 
$\mathbb{E}[ \| e_{k+1} \|^{2} | \mathcal{F}_{k} ] 
 \leq \sigma_{0}^{2} + \sigma_{1}^{2} \| h(\theta_{k}) \|^{2}$, 
$\sigma_{0}^{2}, \sigma_{1}^{2} \in [0, \infty)$, 
the following inequality holds:
\begin{align}
\mathbb{E}[ \| \xi_{k+1} \|^{2} | \mathcal{F}_{k} ] 
&\leq 
 \alpha ( \sigma_{0}^{2} + (\sigma_{1}^{2}+1) \| h(\theta_{k}) \|^{2}) 
 + (1-\alpha) \min(dU^{2}, \gamma^{2}).
\label{eq:inequality_expectation_square_norm_noise_vector}
\end{align}
\label{lemma:inequality_expectation_square_norm_noise_vector}
\end{lemma}
The proof of Lemma~\ref{lemma:inequality_expectation_square_norm_noise_vector} is given in 
Appendix~\ref{subsection:proof_of_lemma_inequality_expectation_square_norm_noise_vector}. 
Note that 
the right-hand side of 
Equation~\eqref{eq:inequality_expectation_square_norm_noise_vector} 
represents the weighted sum of variances of the noise vector 
in Equation~\eqref{eq:def_noise_vector} 
from the true distribution as well as the noisy one. 
In particular, 
the first term on the right-hand side of 
Equation~\eqref{eq:inequality_expectation_square_norm_noise_vector} 
has an additional term $\| h(\theta_{k}) \|^{2}$ 
when compared with the noiseless case 
in Assumption~\ref{assumption:martingale_difference_sequence}. 
This indicates the bias of the noise vector 
by truncating $G_{\theta}(Y)$ in Equation~\eqref{eq:def_G}. 

The following theorem then holds: 
\begin{theorem}
Let us consider the SA scheme in Equation~\eqref{eq:truncated_SA_scheme}. 
If we assume that Assumption~\ref{assumption:noise_uniform_distribution} holds 
and 
$\mathbb{E} [ \| e_{k+1} \|^{2} | \mathcal{F}_{k} ] 
 \leq \sigma_{0}^{2} + \sigma_{1}^{2} \| h(\theta_{k}) \|^{2}, \, 
 \sigma_{0}^{2}, \sigma_{1}^{2} \in [0, \infty)$, 
$\rho_{k} < (1 - 2c_{1} d_{1} \int_{\gamma}^{\infty} \exp{ \left(- \frac{z^{2} }{ M^{2} } \right) } \d{z}) / (2c_{1} L( \sigma_{1}^{2} + 2 ) )$, 
the following inequality holds for $\gamma > 0$: 
\begin{align}
\mathbb{E}[ \| h(\theta_{N}) \|^{2} ] 
&= \frac{ 
     \sum_{k=0}^{n} 
       \rho_{k+1} \mathbb{E}[ \| h(\theta_{k}) \|^{2} | \mathcal{F}_{k}] 
   }{
     \sum_{k=0}^{n} \rho_{k+1}
   }  \\
&\leq 
  2 \left(
     c_{0} + 
     c_{1}(d_{0}+1)
     \int_{\gamma}^{\infty} 
       \exp{ \left(- \frac{z^{2}}{M^{2}} \right) } \, \d{z} 
  \right)  \\
&\quad 
  + 2 c_{1} \frac{ 
        V_{0, n} 
        +
        L ( \alpha \sigma_{0}^{2} + 
          (1-\alpha) \min(dU^{2}, \gamma^{2}) 
        )
        \sum_{k=0}^{n} \rho_{k+1}^{2} 
    }{
      \sum_{k=0}^{n} \rho_{k+1} 
    }.
\label{eq:expectation_mean_field_upper_bound}
\end{align}
where 
$V_{0, n} = \mathbb{E}[ V(\theta_{0}) - V(\theta_{n+1}) | \mathcal{F}_{n} ]$,  
$L$ is a constant that satisfies 
Assumption~\ref{assumption:assumptions_of_Lyapunov_function} (c), 
$d$ is the dimension of the data,  
and $\alpha$ is the mixture ratio in 
Equation~\eqref{eq:real_data_generation}. 
\label{theorem:tradeoff_between_discounting_factor_and_gradient_of_stochastic_update}
\end{theorem}
The proof of  Theorem~\ref{theorem:tradeoff_between_discounting_factor_and_gradient_of_stochastic_update} is given in 
Appendix~\ref{section:proof_of_theorem_tradeoff_between_discounting_factor_and_gradient_of_stochastic_update}. 
Note that $c_{0}$ is an inevitable bias term between the mean field and gradient of the Lyapunov function 
defined in Assumption~\ref {assumption:assumptions_of_Lyapunov_function} (a). 
This also appeared in Equation~\eqref{eq:upper_bound_karimi2019colt}. 
When we set $\rho_{k} = \rho = \mathrm{const}.$ 
in Equation~\eqref{eq:expectation_mean_field_upper_bound}, 
Theorem~\ref{theorem:tradeoff_between_discounting_factor_and_gradient_of_stochastic_update} 
is represented by 
\begin{align}
\mathbb{E} [ \| h(\theta_{N}) \|^{2} ]  
&\leq 2c_{0} 
     + 2 c_{1} (d_{0} + 1) 
       \int_{\gamma}^{\infty} 
         \exp{ \left( - \frac{z^{2}}{M^{2}} \right) } \d{z}  \\
&\quad 
 + \frac{ 2c_{1} V_{0, n} }{ \rho(n+1) } 
 + 2c_{1} \rho L 
   ( \alpha \sigma_{0}^{2} 
     +
     (1-\alpha) \min( dU^{2}, \gamma^{2} )
   ). 
\label{eq:tradeoff_between_const_discounting_factor_and_gradient_of_stochastic_update}
\end{align}
Equation~\eqref{eq:tradeoff_between_const_discounting_factor_and_gradient_of_stochastic_update} 
asserts that 
the SA scheme in Equation~\eqref{eq:truncated_SA_scheme} 
finds an $O( c_{0} + 1/\rho n + 
             \rho (\alpha + 
                   (1-\alpha) \min(dU^{2}, \gamma^{2})
             )
          )$ 
stationary point within $n$ iterations. 
Note that 
when $\rho_{k}$ is a constant 
or the decay rate of $\rho_{k}$ is small, 
whenever a change occurs according to Equation~\eqref{eq:real_data_abrupt_change}, 
the convergence rate of Equation~\eqref{eq:expectation_mean_field_upper_bound} 
is considered to be dependent on $c_{0}$, $c_{1}$, $d_{0}$, $M$, 
$L$, $\alpha$, $\sigma_{0}$, $U$, and $\gamma$. 
Consequently, it is independent of the change point $t^{\ast}$ 
in Equation~\eqref{eq:real_data_abrupt_change}. 
This means that when a change in the distribution occurs 
according to Equation~\eqref{eq:real_data_abrupt_change}, 
if the distribution satisfies  
the assumptions of Theorem~\ref{theorem:tradeoff_between_discounting_factor_and_gradient_of_stochastic_update},  
it converges at an almost constant rate 
irrespective of when a change occurs.  
In that sense, 
SRA is guaranteed to possess the adaptivity. 
In contrast, 
when we adopt decreasing step sizes, 
for example, 
the convergence rate deteriorates 
because the step sizes become small 
if the change happens later. 
In this case, the adaptivity decreases. 

Because $\alpha$, $c_{0}$, $c_{1}$, $\sigma_{0}$, and $L$ 
are generally unknown, 
we have to tune these parameters using, for example, cross validation. 

The following corollary holds with regard to
the relationship between the threshold parameter $\gamma$ 
and the step size $\rho$: 
\begin{corollary}
If $\gamma < \sqrt{d} U$, 
and we set $\rho_{k} = \rho = \mathrm{const.}$, 
the right-hand side of 
Equation~\eqref{eq:expectation_mean_field_upper_bound} 
is minimized by
\begin{align}
\rho
&= \frac{ (d_{0}+1) \exp{ \left(- \frac{ \gamma^{2} }{ M^{2} } \right)} }{ 2L(1-\alpha) \gamma }. 
\label{eq:estimated_rho_by_gamma}
\end{align}
\label{corollary:rho_determined_by_gamma}
\end{corollary}
The proof of Corollary~\ref{corollary:rho_determined_by_gamma} is given in Appendix~\ref{subsection:corollary_rho_determined_by_gamma}. 

\subsection{ Effect of $\gamma$ }

Next, 
we address  
how the upper bound of 
Equation~\eqref{eq:tradeoff_between_const_discounting_factor_and_gradient_of_stochastic_update}
behaves when $\gamma$ goes to infinity. 
The following corollary holds with regard to the expectation 
of the norm of the mean field $h(\theta)$: 
\begin{corollary}
The following inequality holds:
\begin{align}
\lim_{\gamma \rightarrow \infty} \, 
  \mathbb{E} [ \| h(\theta_{N}) \|^{2} ]  
\leq 
 2 c_{0} + 
 2 c_{1} \frac{ 
   V_{0,n} + 
   L(\alpha \sigma_{0}^{2} + 
     (1-\alpha) dU^{2}
   ) \sum_{k=0}^{n} \rho_{k+1}^{2} 
 }{ \sum_{k=0}^{n} \rho_{k+1} }. 
\label{eq:tradeoff_between_discounting_factor_and_gradient_of_stochastic_update_gamma_infty} 
\end{align}
\label{corollary:tradeoff_between_discounting_factor_and_gradient_of_stochastic_update_gamma_infty} 
\end{corollary}
The proof of Corollary~\ref{corollary:tradeoff_between_discounting_factor_and_gradient_of_stochastic_update_gamma_infty} 
is given in 
Appendix~\ref{subsection:proof_corollary_tradeoff_between_discounting_factor_and_gradient_of_stochastic_update_gamma_infty}. 
Note that 
Equation~\eqref{eq:tradeoff_between_discounting_factor_and_gradient_of_stochastic_update_gamma_infty} 
recovers Equation~\eqref{eq:upper_bound_karimi2019colt}, 
when $\alpha=1$ (noiseless case). 

The following corollary then holds with regard to the decrease in the upper bound by setting $\gamma$. 
\begin{corollary}
The difference of the upper bounds 
between 
Equation~\eqref{eq:tradeoff_between_discounting_factor_and_gradient_of_stochastic_update_gamma_infty} 
and 
Equation~\eqref{eq:expectation_mean_field_upper_bound} 
is 
calculated as 
\begin{align}
g(\gamma) 
&= 2c_{1} \frac{ 
  L(1-\alpha) \max(0, dU^{2} - \gamma^{2})  
  \sum_{k=0}^{n} \rho_{k+1}^{2} 
}{
  \sum_{k=0}^{n} \rho_{k+1}
}  \\
&\quad  
 -2 c_{1} (d_{0} + 1) 
 \int_{\gamma}^{\infty} 
   \exp{\left( - \frac{ z^{2} }{ M^{2} } \right)} \d{z}. 
\label{eq:difference_between_upper_bounds}
\end{align}
\label{corollary:difference_between_upper_bounds}
\end{corollary}
The proof of Corollary~\ref{corollary:difference_between_upper_bounds} is given in 
Appendix~\ref{subsection:corollary_difference_between_upper_bounds}. Equation~\eqref{eq:difference_between_upper_bounds} 
represents the effect of setting $\gamma$. 
The first term on the right-hand side of 
Equation~\eqref{eq:difference_between_upper_bounds} 
determines the decrease of the upper bound 
by setting $\gamma$ as the threshold parameter. 
In contrast, 
the second term appears on the right-hand side, as its demerit. 
As was mentioned after Lemma~\ref{lemma:inequality_expectation_lyapunov_function_gradient_noise_vector}, 
if $H_{\theta_{k}}$ is bounded, 
the cost of the second term disappears 
in a finite region. 
In such a case, 
the advantage of SRA becomes clearer. 
In fact, 
if $\| H_{\theta_{k}} \| \leq \gamma^{\ast}$ holds ($\gamma^{\ast} < \infty)$, 
we get the following inequality with Hoeffding's  inequality \citep{Vershynin2018}:
\begin{align}
P[ \| H_{\theta_{k}}(Y_{k+1}) \| \geq z ]
\leq \exp{ \left( -\frac{z^{2}}{(\gamma^{\ast} - \gamma)^{2}}
\right)}
\quad 
(\gamma \leq z \leq \gamma^{\ast}).
\end{align}
We then obtain the following equation for $\gamma < \gamma^{\ast}$:
\begin{align}
g(\gamma)
&= 2c_{1} 
   \frac{ L(1-\alpha) \max(0, dU^{2} - \gamma^{2}) \sum_{k=0}^{n} \rho_{k+1}^{2} 
   }{
     \sum_{k=0}^{n} \rho_{k+1}
   }  \\
&\quad 
   -2c_{1}(d_{0} + 1)
    \int_{\gamma}^{\gamma^{\ast}}
      \exp{ \left( - \frac{z^{2}}{(\gamma^{\ast} - \gamma)^{2}}
      \right) } \, \d{z}. 
\end{align}
Therefore, 
the following equation holds 
for $\gamma \geq \gamma^{\ast}$: 
\begin{align}
g(\gamma)
= 2c_{1}
  \frac{ 
    L(1-\alpha) \max(0, dU^{2} - \gamma^{2}) \sum_{k=0}^{n} \rho_{k+1}^{2}
  }{
    \sum_{k=0}^{n} \rho_{k+1}
  }. 
\label{eq:difference_when_gamma_over_threshold}
\end{align}
When $\gamma$ satisfies $\gamma^{\ast} \leq \gamma \leq \sqrt{d} U$, 
Equation~\eqref{eq:difference_when_gamma_over_threshold} 
shows that the effect of setting $\gamma$ 
is proportional to $dU^{2} - \gamma^{2}$.

%% file: 5_Experiments.tex
\section{Experiments}
\label{section:experiments}

In this section,  
we present the experimental results of SRA. 
We used a standard laptop 
with an Intel Core i9 
with 2.9~GHz $\times$ 6 Core 
and 
32GB of Ram. 
The source code is available at \url{https://github.com/s-fuku/robustadapt}. 
We conducted experiments on 
univariate synthetic datasets with abrupt and gradual changes 
in Section~\ref{subsection:univariate_synthetic_datasets}, 
and mutivariate synthetic datasets with abrupt and gradual changes 
in Section~\ref{subsection:multivariate_synthetic_datasets}. 
We examined the performance of SRA with respect to change detection, on real datasets, in Section~\ref{subsection:real_dataset_change_detection}, namely, 
the Well-log dataset~\citep{Ruanaidh1996} 
for univariate data stream, 
and the SKoltech Anomaly Benchmark~(SKAB) dataset~\citep{Katser2020} 
for multivariate one. 
We also investigated the performance of SRA with respect to anomaly detection, 
on real datasets in Section
\ref{subsection:real_dataset_anomaly_detection} 
for multivariate data streams: 
the SMTP and THYLOID datasets. 
Finally, 
we discuss the conclusions
based on the experiments in Section
\ref{subsection:concluding_remarks_experiments}.

\subsection{Univariate synthetic datasets}
\label{subsection:univariate_synthetic_datasets}

We generated 
univariate sequences 
with abrupt and gradual changes, 
from mixtures of 
true distribution and noisy one. 

\subsubsection{Datasets} 
\label{subsubsection:univariate_synthetic_datasets_generation}

We generated the following univariate sequences:
\begin{align}
y_{t} \sim f = \alpha f_{1} + (1 - \alpha) f_{2} \quad (t=1, \dots, 20000), 
\label{eq:data_generation_synthetic_dataset_one_dim}
\end{align}
where 
$f_{1} \in \mathcal{F}$ is an element of 
a parametric class of distribution 
$\mathcal{F} = \{ f(y; \theta), \theta \in \Theta \}$. 
$\theta$ is a parameter 
and 
$\Theta \subset \mathbb{R}^{p}$ 
is a parameter space associated. 
$f_{2}$ is an element of 
a parametric class of data distribution 
$\mathcal{G} = \{ f_{2}(y; \xi), \xi \in \Xi \}$, 
where 
$\xi$ is a parameter 
and 
$\Xi \subset \mathbb{R}^{m}$ is a parameter space associated, 
and 
Equation~\eqref{eq:data_generation_synthetic_dataset_one_dim} is equal to Equation~\eqref{eq:real_data_generation}. 
We generated the following two univariate datasets 
with abrupt and gradual changes:
\begin{itemize}
    \item \textbf{Abrupt Change} \\
    We set $f_{1}$ and $f_{2}$ in Equation~\eqref{eq:data_generation_synthetic_dataset_one_dim} as follows:
    \begin{align}
    f_{1} &= \frac{1}{2} \mathcal{N}(y; \mu_{1}, \sigma_{1}) +
         \frac{1}{2} \mathcal{N}(y; \mu_{2}, \sigma_{2}), \\
    f_{2} &= \mathrm{Uniform}(y; -U, U), \\
    \mu &= \left(
        \begin{array}{c}
        \mu_{1} \\ 
        \mu_{2}
        \end{array}
      \right) 
    = \begin{cases}
    \transpose{(0.5, -0.5)} & (t \leq  10000),  \\
    \transpose{(1.0, -1.0)} & (10001 \leq t \leq 20000),  
    \end{cases} \\
    \sigma_{1} &= \sigma_{2} = 0.1. 
    \label{eq:synthetic_univariate_data_generation_with_abrupt_change}
    \end{align}
    These sequences have a change point at $t=10001$, 
    where the mean changes abruptly. 
    
    \item \textbf{Gradual Change} \\
    We set $f_{1}$ and $f_{2}$ in Equation~\eqref{eq:data_generation_synthetic_dataset_one_dim} as follows:
    \begin{align}
    f_{1} &= \frac{1}{2} \mathcal{N}(y; \mu_{1}, \sigma_{1}) +
             \frac{1}{2} \mathcal{N}(y; \mu_{2}, \sigma_{2}), \\
    f_{2} &= \mathrm{Uniform}(y; -U, U), \\
    \mu &= \left(
        \begin{array}{c}
        \mu_{1} \\ 
        \mu_{2}
        \end{array}
      \right) 
        = \begin{cases} 
          \transpose{(0.5, -0.5)} & (t \leq  10000),  \\
          \transpose{(0.5, -0.5)} 
          + \frac{t-10000}{300}
          \transpose{(0.5, -0.5)} & (10001 \leq t \leq 10300), \\
          \transpose{(1.0, -1.0)} & (10301 \leq t \leq 20000),  
        \end{cases} \\
    \sigma_{1} &= \sigma_{2} = 0.1. 
    \label{eq:synthetic_univariate_data_generation_with_gradual_change}
    \end{align}
    These sequences have a change point at $t=10001$, 
    where the mean starts to change gradually up to $t=10300$. 
\end{itemize}

\begin{figure}[tb]
\centering
\includegraphics[width=\textwidth]{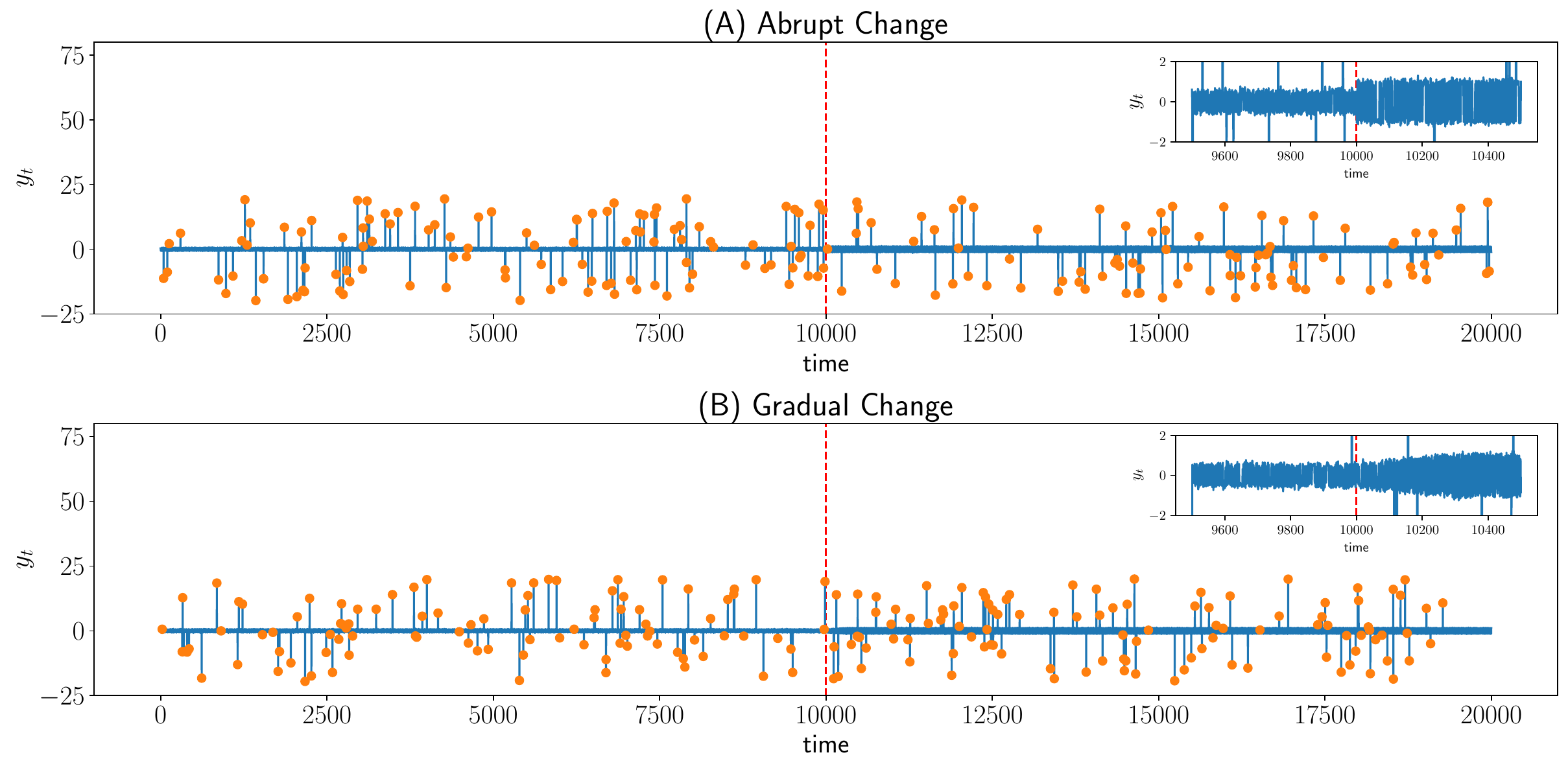}
\caption{Sample data streams of univariate synthetic datasets with abrupt and gradual changes. 
Each data point is drawn from 
Equation~\eqref{eq:synthetic_univariate_data_generation_with_abrupt_change} 
and 
\eqref{eq:synthetic_univariate_data_generation_with_gradual_change} 
for abrupt and gradual changes, 
respectively~($\alpha=0.99$ and $U=20$). 
Data points drawn from $f_{2}$ are marked with circles.  
Insets show the data streams between $t=9500$ and $t=10500$. 
(A) The data stream changes abruptly at $t=10001$ 
and 
(B) The data stream changes gradually from $t=10001$ up to $t=10300$. 
The red dashed lines indicate the change points at $t=10001$. 
}
\label{fig:sample_data_streams_of_univariate_synthetic_datasets_with_abrupt_and_gradual_changes}
\end{figure}

Figure~\ref{fig:sample_data_streams_of_univariate_synthetic_datasets_with_abrupt_and_gradual_changes} 
displays the sample data streams  
with abrupt and gradual changes. 
Each data point is drawn from Equation~\eqref{eq:synthetic_univariate_data_generation_with_abrupt_change} 
and 
~\eqref{eq:synthetic_univariate_data_generation_with_gradual_change} 
for abrupt and gradual changes, 
respectively. 
We set $\alpha=0.99$ and $U=20$. 
The data points drawn from $f_{2}$ are marked with circles, 
and 
we observe that most of the data points 
drawn from $f_{2}$ 
deviate from the ones drawn from $f_{1}$. 

\subsubsection{Methods for comparison}
\label{subsubsection:methods_for_comparison_univariate_synthetic_datasets}

We compared the performance of SRA with those of rival algorithms. 
We chose the following algorithms for comparison:
\begin{itemize}
\item \textbf{SDEM}~\citep{Yamanishi2004}: 
an online learning algorithm based on GMM. 
SDEM sequentially updates parameters 
and adapts to non-stationary changes 
with the discounting parameter. 
\item \textbf{iEM}~\citep{Neal1999}: 
an EM algorithm in an incremental scheme. 
As is pointed out in \citep{Yamanishi2004}, 
iEM is thought of as a version of SDEM, 
where the discounting parameter is set to $r=1/t$ at time $t$ 
with a fresh sample drawn each time. 
\item \textbf{sEM}~\citep{Cappe2009}: 
a stochastic~(online)~EM algorithm. 
sEM updates sufficient statistics 
in an SA scheme. 

\item \textbf{ADWIN}~\citep{Bifet2007}: 
an adaptive sliding window algorithm for detecting changes, 
and keeping updated statistics about a data stream. 
\item \textbf{KSWIN}~\citep{Raab2020}: 
a concept drift detection method 
based on the Kolmogorov-Smirnov~(KS) statistical test. 
\item \textbf{PH}~\citep{Page1954}: 
a change detection method for computing the observed values and their means up to the current moment. 

\end{itemize}
SDEM, iEM, and sEM are 
online learning algorithms, 
whereas 
ADWIN, KSWIN, and PH are 
concept drift detection algorithms~
(e.g., \citep{Goncalves2014}). 
We used the \texttt{scikit-multiflow} library\footnote{\url{https://scikit-multiflow.github.io/}}~\citep{Montiel2018} to implement the concept drift detection algorithms. 

\subsubsection{Evaluation metrics}
\label{subsubsection:evaluation_metrics_univariate_synthetic_datasets}

We defined 
two evaluation metrics: 
(i) 
the Area under the Curve~(AUC) 
to compare the performances of 
the online learning algorithms 
and 
concept drift detection algorithms, 
and 
(ii) 
the mean squared error~(MSE) 
to evaluate how well 
the online learning algorithms 
estimate the parameters. 

First, we defined AUC 
to compare the online learning algorithms 
and 
concept drift detection algorithms. 
SDEM, iEM, and sEM are online learning algorithms. 
For each algorithm, 
we calculated the change score as  
$s_{t} \mydef -\log{ f(y_{t}; \hat{\theta}_{t-1}) }$, 
where $\hat{\theta}_{t-1}$ is the parameter estimated at $t-1$. 
This change score has often been used 
in online change detection and anomaly detection~\citep{Yamanishi2004,Yamanishi2002,Fukushima2019}. 

In contrast, 
ADWIN, KSWIN, and PH are 
concept drift detection algorithms. 
We defined the change score of ADWIN 
as $s_{t} \mydef w_{t-1} - w_{t} $, 
where $w_{t}$ denotes the window width at time $t$. 
This score quantifies 
how much the window size shrinks between $t-1$ and $t$,  
and thus, 
how large the change is at time $t$. 
We defined the change score of KSWIN 
as $s_{t} \mydef 1 - p_{t}$, 
where $p_{t}$ is the $p$-value obtained from the KS statistical test.
We defined the change score of PH 
as $s_{t} \mydef g_{t} - G_{t}$, 
where 
$g_{0} = 0$, 
$g_{t} = g_{t-1} + y_{t} - \delta_{\mathrm{PH}}$, 
and 
$G_{t} = \min \{ g_{t}, G_{t-1} \}$. 
$\delta_{\mathrm{PH}}$ is a threshold parameter. 

We evaluated the performance of each algorithm for the training dataset 
in terms of detection delay and overdetection. 
We used the AUC score 
in terms of \textit{benefit} and \textit{false alarms} ~\citep{Fawcett1999,Yamanishi2016,Fukushima2019,Fukushima2020}. 
The AUC score was calculated as follows: 
we first fixed the threshold parameter $\epsilon$ 
and converted the change scores $\{ s_{t} \}$ 
to binary alarms $\{ \alpha_{t} \}$. 
That is, 
$\alpha_{t} = \mathbbm{1} \, (s_{t} > \epsilon)$, 
where $\mathbbm{1}(s)$ denotes a binary function that takes on the value of 1 if and only if $s$ is true. 
We let $T_{\mathrm{b}}$ be the maximum tolerant delay 
of change detection. 
In this experiment, 
we set $T_{\mathrm{b}}=100$. 
When the actual time of change was $t^{\ast}$, 
we defined the \textit{benefit} of an alarm at time $t$ as 
\begin{align}
b(t; t^{\ast}) =
  \begin{cases}
  1 - \frac{ | t - t^{\ast} |}{T_{\mathrm{b}}} &
  (0 \leq | t - t^{\ast} | < T_{\mathrm{b}}) \\
  0 & (\mathrm{otherwise})
  \end{cases}
\label{eq:def_benefit}
\end{align}
The number of \textit{false alarms} was calculated as 
\begin{align}
n(s_{t_{\mathrm{start}}}^{t_{\mathrm{end}}}) = \sum_{k=1}^{m} \alpha_{t_{k}} \mathbbm{1}(b(t_{k}, t^{\ast}) = 0),
\end{align}
where 
$t_{\mathrm{start}}$ and $t_{\mathrm{end}}$ are the starting and end time points 
for evaluation, respectively, 
and 
$s_{t_{\mathrm{start}}}^{t_{\mathrm{end}}} = s_{t_{\mathrm{start}}} \dots 
s_{t_{\mathrm{end}}}$ 
denotes a sequence of change scores 
within this period. 
We calculated the AUC based on the recall rate of the total benefit, 
$b/\sup_{\epsilon} b$, 
and the false alarm rate, 
$n/\sup_{\epsilon} n$, 
with $\epsilon$ varying. 

Next, we evaluated the performance of SRA 
and compared the online learning algorithms 
in terms of how well they estimated the parameters. 
We used the following mean squared errors~(MSE): 
\begin{align}
S_{\mathrm{eval}}
&= \frac{ \displaystyle
     \sum_{t=\tau_{\mathrm{start}}}^{\tau}
       \| \hat{\mu}_{t} - \mu_{t} \|^{2}
   }{
     \tau - \tau_{\mathrm{start}}
   }, 
&S_{\mathrm{tot}}
&= \frac{ \displaystyle
    \sum_{t=\tau+1}^{T}
      \| \hat{\mu}_{t} - \mu_{t} \|^{2}
  }{
    T - \tau
  }, \\
S_{\mathrm{bc}}
&= \frac{ \displaystyle
    \sum_{t=\tau+1}^{t^{\ast}-1}
      \| \hat{\mu}_{t} - \mu_{t} \|^{2}
  }{
    t^{\ast} - \tau
  }, 
&S_{\mathrm{ac}}
&= \frac{ \displaystyle
    \sum_{t=t^{\ast}+1}^{T}
      \| \hat{\mu}_{t} - \mu_{t} \|^{2}
  }{
    T - t^{\ast}
  }, 
\label{eq:def_mse_metrics}
\end{align}
where $\hat{\mu}_{t}$ is the estimated mean at $t$, 
$T$ is the sequence length ($T=20000$), 
$t^{\ast}$ is the change point~($t^{\ast}=10001$), 
and $\tau \in \mathbb{N}$ is a transient period. 

$S_{\mathrm{tot}}$, 
$S_{\mathrm{bc}}$, 
and 
$S_{\mathrm{ac}}$ 
represent the MSEs 
for the overall sequence, 
time points before the change point, 
and time points after the change point, 
respectively. 
$S_{\mathrm{eval}}$
represents the MSE 
between $t=\tau_{\mathrm{start}}$ and $t=\tau$. 
We set $\tau_{\mathrm{start}} = 500$, 
thus $S_{\mathrm{eval}}$ measures the MSE 
between $t=500$ and $t=999$. 
Each MSE excludes the transition period 
between $t=1$ and $t=\tau_{\mathrm{start}}-1$. 
We set $\tau=1000$ 
and the number of components of GMM to $K=2$ 
for each online learning algorithm 
throughout the following experiments 
for synthetic datasets. 
For each algorithm, 
we initialized the parameters 
or sufficient statistics with the data in the first 10 time steps.

\subsubsection{Result1: Tradeoff between $\gamma$ and $\rho$}
\begin{figure}[tb]
\begin{center}
\includegraphics[width=\linewidth]{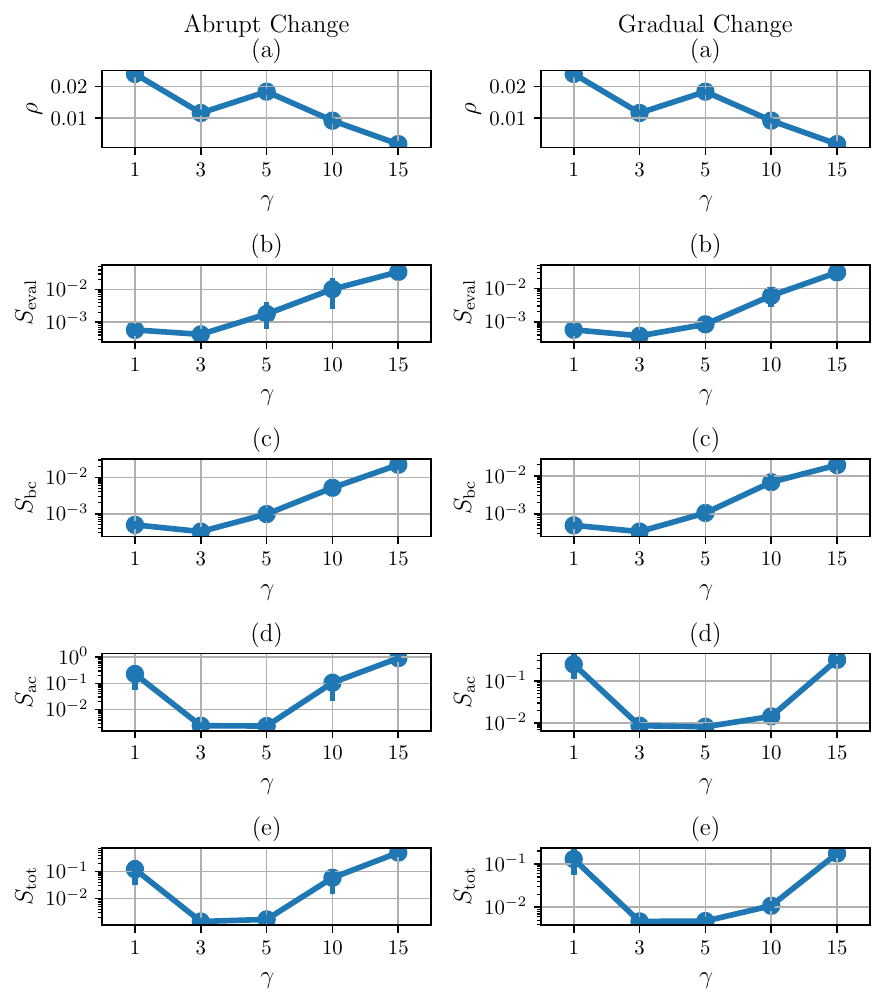}
\caption{Relation between the threshold parameter for stochastic update $\gamma$ and the step size $\rho$, and MSEs for both synthetic datasets with abrupt and gradual changes, respectively. 
   (a) $\rho$ estimated with Equation~\eqref{eq:estimated_rho_by_gamma}. 
   (b) MSE $S_{\mathrm{eval}}$ between $t=500$ and $t=999$ (evaluation). 
   (c) MSE $S_{\mathrm{bc}}$ between $t=1000$ and $t=10000$ (before the change). 
   (d) MSE $S_{\mathrm{ac}}$ between $t=10001$ and $t=20000$ (after the change). 
   (e) MSE $S_{\mathrm{tot}}$ between $t=1000$ and $t=20000$.  
}
\label{fig:experiment1_varying_rho}
\end{center}
\end{figure}

We empirically confirmed the tradeoff between 
the threshold parameter $\gamma$ and the step size $\rho$ 
for both datasets with abrupt and gradual changes, 
respectively. 
In practice, 
the hyperparameters $L$, $\alpha$, $d_{0}$, and $M$ must be tuned to determine $\rho$ by $\gamma$ in Equation~\eqref{eq:estimated_rho_by_gamma}. 
As $\beta \mydef (d_{0} + 1) / L(1-\alpha)$ is regarded as a parameter, 
we changed 
$\gamma \in \{ 1, 3, 5, 10, 15 \}$, 
$\beta \in \{ 0.1, 0.5, 1 \}$, 
and $M \in \{ 1, 5, 10 \}$ 
to estimate the optimal value of $\rho$ in Equation~\eqref{eq:estimated_rho_by_gamma}. 
We generated 10 data streams with $\alpha=0.99$ and $U=20$ 
according to 
Equation~\eqref{eq:data_generation_synthetic_dataset_one_dim}
using Equation~\eqref{eq:synthetic_univariate_data_generation_with_abrupt_change} 
and 
\eqref{eq:synthetic_univariate_data_generation_with_gradual_change} 
for abrupt and gradual changes, 
respectively. 
Figure~\ref{fig:experiment1_varying_rho} shows the estimated $\hat{\rho}$, 
$S_{\mathrm{eval}}$, 
$S_{\mathrm{bc}}$, $S_{\mathrm{ac}}$, and $S_{\mathrm{tot}}$, 
for datasets with abrupt and gradual changes. 
For each $\gamma$, 
the optimal combination of $\gamma$, $\beta$, $M$, 
and $\rho$ estimated in Equation~\eqref{eq:estimated_rho_by_gamma} 
was selected, 
which minimized $S_{\mathrm{eval}}$. 
Figure~\ref{fig:experiment1_varying_rho} shows that 
$S_{\mathrm{eval}}$, $S_{\mathrm{bc}}$, $S_{\mathrm{ac}}$, 
and $S_{\mathrm{tot}}$ 
were minimized when $\gamma=3$ 
for both datasets with abrupt and gradual changes, 
indicating that 
the choice of $\rho$ using Equation~\eqref{eq:estimated_rho_by_gamma} is reasonable 
because it provides both the robustness and adaptivity. 
The best combination of the hyperparameters was 
$(\gamma, \beta, M) = (3, 0.1, 5)$, 
and the estimated $\rho=0.0116$. 
We also observe from Figure~\ref{fig:experiment1_varying_rho} that 
$S_{\mathrm{eval}}$ and $S_{\mathrm{bc}}$ were not so different 
for $\gamma=1, 3$, 
whereas $S_{\mathrm{ac}}$ and $S_{\mathrm{tot}}$ were different. 
It indicates that $\gamma$ did not have much influence 
prior to the change point between $\gamma=1$ and $\gamma=3$. 
In contrast, after the change point, 
the mean of distribution changed to $\mu_{1, 2} = \pm 1$ 
from $\mu_{1, 2} = \pm 0.5$. 
Therefore, the difference between $\gamma=1$ and $\gamma=3$ 
became significant. 
In fact, 
the sufficient statistics $\hat{s}_{t}$ 
corresponds to mean $\mu_{t}$ in this case. 
The result with $\gamma=1$ indicates that 
the sufficient statistics led to a decrease in accuracy in the estimation of the mean by dropping more data points than $\gamma=3$. 
For $\gamma \geq 5$, 
the decrease in MSEs is observed in comparison with $\gamma=3$ 
even before the change point due to the influence of outliers.

\subsubsection{Result2: Dependency on $\alpha$}

We investigated the dependency of the upper bound 
in Equation~\eqref{eq:expectation_mean_field_upper_bound} on the ratio of the outlier. 
It is characterized as $(1-\alpha)$ in 
Equation~\eqref{eq:data_generation_synthetic_dataset_one_dim}. 
Based on Equation~\eqref{eq:difference_between_upper_bounds}, 
the smaller $\alpha$ is, 
the higher the upper bound of the expectation of the mean field is. 
In other words, the upper bound increases  
as the noisy data increase. 

\begin{figure}[tb]
\begin{center}
\includegraphics[width=\linewidth]{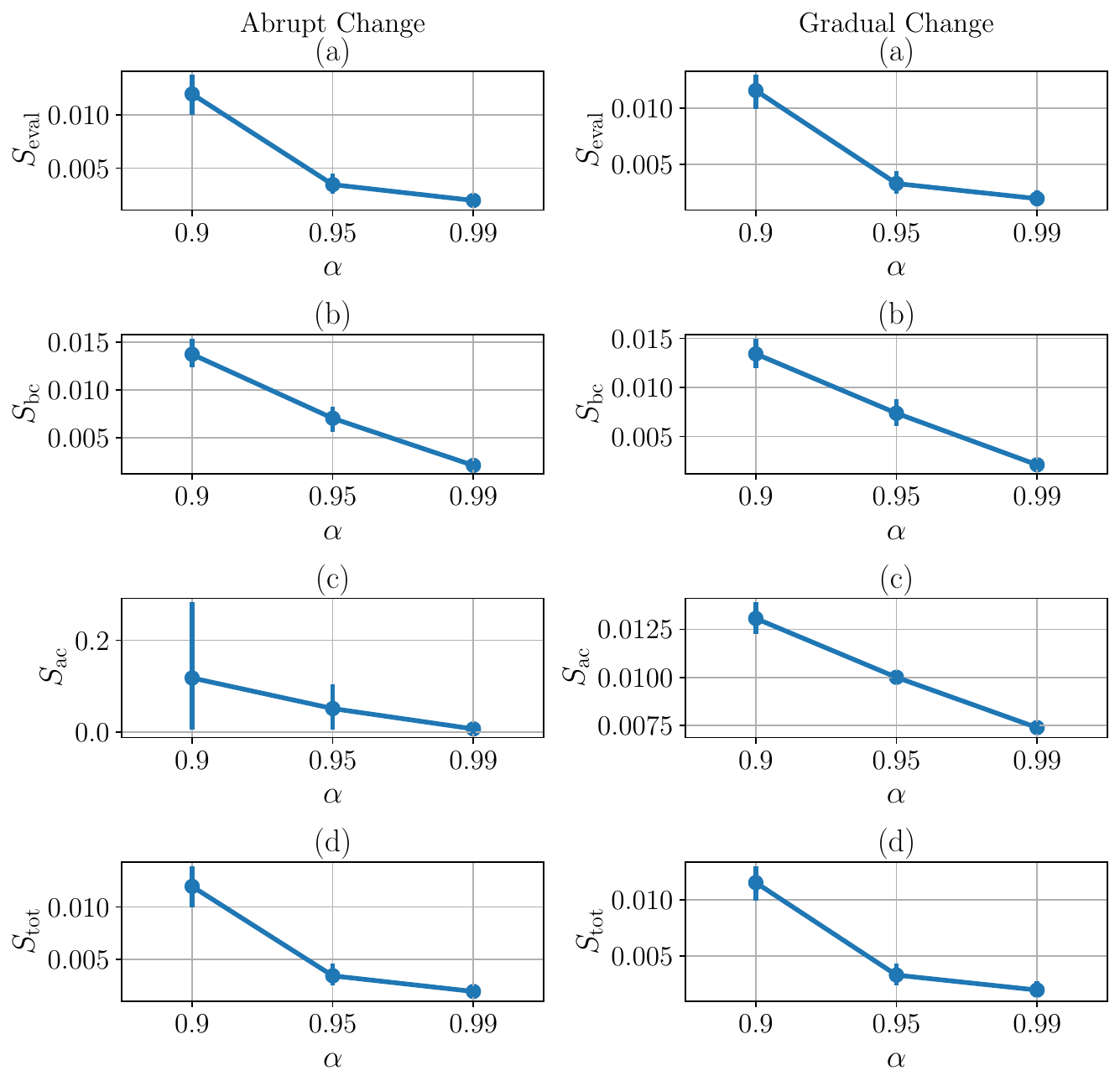}
\caption{Dependencies of the MSEs on $\alpha$ for both synthetic datasets with abrupt and gradual changes. 
  (a) MSE $S_{\mathrm{eval}}$ between $t=500$ and $t=999$ (evaluation). 
  (b) MSE $S_{\mathrm{bc}}$ between $t=1000$ and $t=10000$ (before the change). 
  (c) MSE $S_{\mathrm{ac}}$ between $t=10001$ and $t=20000$ (after the change). 
  (d) MSE $S_{\mathrm{tot}}$ between $t=1000$ and $t=20000$. 
}
\label{fig:experiment1_dependency_on_alpha}
\end{center}
\end{figure}

For $\alpha \in \{ 0.9, 0.95, 0.99 \}$, 
we set  
$\gamma = 3$, 
$M = 5$, 
and $\beta = 0.1 \times (1 - 0.99) / (1 -\alpha) = 10^{-3} /(1-\alpha)$. 
Thus, we used the best combination in the previous experiment, 
but $\beta$ is modified by the value of $\alpha$. 
Therefore, $\rho$ is also modified 
according to Equation~\eqref{eq:estimated_rho_by_gamma}. 
We generated 10 data streams with $\alpha=0.99$ and $U=20$ 
according to 
Equation~\eqref{eq:data_generation_synthetic_dataset_one_dim} 
using Equation~\eqref{eq:synthetic_univariate_data_generation_with_abrupt_change} 
and 
\eqref{eq:synthetic_univariate_data_generation_with_gradual_change} 
for abrupt and gradual changes, 
respectively. 
We then estimated the MSEs $S_{\mathrm{bc}}$, $S_{\mathrm{ac}}$, and $S_{\mathrm{tot}}$. 
Figure~\ref{fig:experiment1_dependency_on_alpha} shows $S_{\mathrm{tot}}$, 
$S_{\mathrm{bc}}$, and $S_{\mathrm{ac}}$. 
Each MSE decreased as $\alpha$ increased. 
This result is consistent with Equation~\eqref{eq:difference_between_upper_bounds}. 

Note that $S_{\mathrm{bc}}$ was much smaller 
for all the values of $\alpha$ 
in gradual changes 
than these in abrupt changes. 
That is, 
SRA is more adaptive to gradual changes 
than to abrupt ones. 
It indicates that 
it is easier for SRA 
to follow the gradual changes 
of the parameters of the distribution. 
It is intuitive and easily understandable. 

\subsubsection{Result3: Comparison with other algorithms}
\label{subsubsection:result_comparison_with_other_algorithms_synthetic_1D}

We compared the performance of SRA with those of rival algorithms. 

SDEM \citep{Yamanishi2004} and sEM \citep{Cappe2009} has the discounting parameter $r$ 
to adapt to new data. 
We chose $r_{\mathrm{SDEM}} \in \{ 0.0001$, $0.001$, $0.005$, $0.01 \}$ 
and $r_{\mathrm{sEM}} \in \{ 0.001$, $0.003$, $0.005 \}$. 
We set the parameters of SRA to 
$(\gamma$, $\beta$, $M$, $\rho)= (3$, $0.1$, $5$, $0.0116)$, 
that is, the best combination in the previous experiment.  
Then, we evaluated $S_{\mathrm{eval}}$, 
the MSE between $t=500$ and $t=999$, as before. 
As a result, 
the discounting parameters were chosen 
for SDEM and sEM as follows: 
$r_{\mathrm{SDEM}}=0.01$ and $r_{\mathrm{sEM}} = 0.005$~(abrupt change), 
and 
$r_{\mathrm{SDEM}}=0.01$ and $r_{\mathrm{sEM}} = 0.005$~(gradual change). 
We generated 10 data streams with $\alpha=0.99$ and $U=20$ 
according to 
Equation~\eqref{eq:data_generation_synthetic_dataset_one_dim} 
using Equation~\eqref{eq:synthetic_univariate_data_generation_with_abrupt_change} 
and 
\eqref{eq:synthetic_univariate_data_generation_with_gradual_change} 
for abrupt and gradual changes, 
respectively. 
Table~\ref{table:average_mses_on_the_univariate_synthetic_datasets} shows 
the average MSEs $S_{\mathrm{bc}}$, $S_{\mathrm{ac}}$, 
and $S_{\mathrm{tot}}$ for each algorithm. 
SRA was superior to other algorithms 
for the time periods after the change point and before it. 
This result indicates that SRA is better equipped in terms of both the robustness and adaptivity 
compared to other algorithms 
for both abrupt and gradual changes.

\begin{table}[tb]
\begin{center}
\caption{Average MSEs on the univariate synthetic datasets.}
\label{table:average_mses_on_the_univariate_synthetic_datasets}

\begin{subtable}{\linewidth}
\caption{Abrupt Change}
\centering
{\tabcolsep=0.8\tabcolsep
\begin{tabular}{rrrr}
 \toprule
 & \multicolumn{1}{c}{$S_{\mathrm{bc}}$} 
 & \multicolumn{1}{c}{$S_{\mathrm{ac}}$}
 & \multicolumn{1}{c}{$S_{\mathrm{tot}}$}  \\
 \midrule
 \multicolumn{1}{c}{SRA} & 
 $\mathbf{0.009 \pm 0.001}$ & 
 $\mathbf{0.010 \pm 0.001}$ & 
 $\mathbf{0.009 \pm 0.001}$ \\
 \multicolumn{1}{c}{SDEM} & 
 $0.507 \pm 0.001$ & 
 $2.013 \pm 0.003$ & 
 $1.300 \pm 0.002$ \\
 \multicolumn{1}{c}{iEM} & 
 $0.500 \pm 0.000$ & 
 $2.000 \pm 0.000$ & 
 $1.290 \pm 0.000$ \\
 \multicolumn{1}{c}{sEM} & 
 $0.515 \pm 0.002$ & 
 $2.025 \pm 0.004$ & 
 $1.310 \pm 0.002$ \\
 \bottomrule
\end{tabular}
}
\end{subtable}
\hfil

\begin{subtable}{\linewidth}
\caption{Gradual Change}
\centering
{\tabcolsep=0.8\tabcolsep
\begin{tabular}{rrrr}
 \toprule
 & \multicolumn{1}{c}{$S_{\mathrm{bc}}$} 
 & \multicolumn{1}{c}{$S_{\mathrm{ac}}$}
 & \multicolumn{1}{c}{$S_{\mathrm{tot}}$}  \\
 \midrule
 \multicolumn{1}{c}{SRA} & 
 $\mathbf{0.005 \pm 0.000}$ & 
 $\mathbf{0.002 \pm 0.000}$ & 
 $\mathbf{0.001 \pm 0.000}$ \\
 \multicolumn{1}{c}{SDEM} & 
 $1.010 \pm 0.985$ & 
 $3.946 \pm 3.916$ & 
 $0.978 \pm 0.793$ \\
 \multicolumn{1}{c}{iEM} & 
 $0.267 \pm 0.632$ & 
 $0.969 \pm 1.622$ & 
 $0.637 \pm 1.152$ \\
 \multicolumn{1}{c}{sEM} & 
 $0.029 \pm 0.005$ & 
 $0.315 \pm 0.029$ & 
 $0.309 \pm 0.024$ \\
 \bottomrule
\end{tabular}
}
\end{subtable}
\end{center}
\end{table} 

We then evaluated each algorithm with AUC. 
We chose 
$\gamma \in \{ 1$, $3$, $5$, $10$, $15 \}$, 
$\beta \in \{ 0.1$, $0.5$, $1 \}$, 
and 
$M \in \{ 1$, $5$, $10 \}$ 
for SRA. 
We chose the hyperparameters 
from among the values used 
in calculating MSEs 
for SDEM, iEM, and sEM. 
ADWIN~\citep{Bifet2007} 
has the confidence parameter $\delta_{\mathrm{ADWIN}}$. 
We chose $\delta_{\mathrm{ADWIN}} \in \{ 0.001$, $0.002$, $0.005$, $0.01$, $0.02$, $0.05$, $0.1$, $0.2$, $0.5 \}$. 
KSWIN has 
the probability $\alpha_{\mathrm{KSWIN}}$ 
for the test statistic of the KS-Test, 
the sliding window $w_{\mathrm{KSWIN}}$, 
and 
the statistic window $r_{\mathrm{KSWIN}}$. 
We chose $\alpha_{\mathrm{KSWIN}} \in \{ 0.001$, $0.005$, $0.01 \}$, 
$w_{\mathrm{KSWIN}} \in \{ 10$, $20 \}$, 
and 
$r_{\mathrm{KSWIN}} \in \{ 10$, $15 \}$. 
PH has the threshold parameter $\delta_{\mathrm{PH}}$. 
We chose $\delta_{\mathrm{PH}} \in \{ 0.001$, $0.01$, $0.1$, $1$, $10 \}$. 
We set the acceptable false alarm rate 
$\alpha_{\mathrm{PH}} = 0.05$. 

Table~\ref{table:average_aucs_on_the_synthetic_univariate_datasets} 
shows the average AUCs 
for each algorithm. 
The hyperparameters were chosen as follows: 
(i) Abrupt change: 
$(\gamma$, $\beta$, $M)=(3$, $0.1$, $10)$ for SRA, 
$r_{\mathrm{SDEM}}=0.0001$ for SDEM, 
$r_{\mathrm{sEM}}=0.01$ for sEM, 
$\delta_{\mathrm{ADWIN}}=0.5$ for ADWIN, 
$(\alpha_{\mathrm{KSWIN}}$, 
  $w_{\mathrm{KSWIN}}$, 
  $r_{\mathrm{KSWIN}})=(0.01$, $30$, $10)$ for KSWIN, 
and all the values for $\delta_{\mathrm{PH}} = 10$, 
$20$, $30$, $40$, $50$, $60$, $70$, $80$, and $90$ for PH. 
(ii) Gradual change: 
$(\gamma, \beta, M)=(1$, $0.5$, $5)$ for SRA, 
$r_{\mathrm{SDEM}}=0.0001$ for SDEM, 
$r_{\mathrm{sEM}}=0.1$ for sEM, 
$\delta_{\mathrm{ADWIN}}=0.5$ for ADWIN, 
$(\alpha_{\mathrm{KSWIN}}$, 
  $w_{\mathrm{KSWIN}}$, 
  $r_{\mathrm{KSWIN}})=(0.005$, $30$, $15)$ for KSWIN, 
and $\delta_{\mathrm{PH}} = 0.1$ for PH. 

We observed that SRA was superior to other algorithms 
for both abrupt and gradual changes. 
This result indicates that 
SRA is better equipped with both the robustness and adaptivity than other algorithms. 
As SRA has three parameters to be tuned, 
we also show the sensitivities of AUCs of SRA  
on hyperparameters in Appendix~\ref{section:dependency_on_hyperparameters} 
for both the datasets with abrupt and gradual changes. 
We observe that the performance of SRA is dependent on 
the values of the hyperparameters. 
Therefore, we should select them carefully. 
\begin{table}[tb]
\begin{center}
\caption{Average AUCs on the synthetic univariate datasets.}
\label{table:average_aucs_on_the_synthetic_univariate_datasets}
{\tabcolsep=0.8\tabcolsep
\begin{tabular}{rrr}
 \toprule
 & \multicolumn{1}{c}{Abrupt Change} 
 & \multicolumn{1}{c}{Gradual Change} \\
 \midrule
 \multicolumn{1}{c}{SRA} 
 & $\mathbf{0.717 \pm 0.021}$ 
 & $\mathbf{0.702 \pm 0.023}$  \\
 \multicolumn{1}{c}{SDEM} 
 & $0.500 \pm 0.000$ 
 & $0.524 \pm 0.074$  \\
 \multicolumn{1}{c}{iEM} 
 & $0.569 \pm 0.012$ 
 & $0.320 \pm 0.025$  \\
 \multicolumn{1}{c}{sEM} 
 & $0.584 \pm 0.087$ 
 & $0.481 \pm 0.087$  \\
 \multicolumn{1}{c}{ADWIN} 
 & $0.500 \pm 0.000$
 & $0.500 \pm 0.000$ \\
 \multicolumn{1}{c}{KSWIN}
 & $0.570 \pm 0.056$
 & $0.533 \pm 0.067$ \\
 \multicolumn{1}{c}{PH}
 & $0.446 \pm 0.201$
 & $0.416 \pm 0.174$ \\
 \bottomrule
\end{tabular}
}
\end{center}
\end{table} 

\begin{figure}[tb]
\centering
\includegraphics[width=\linewidth]{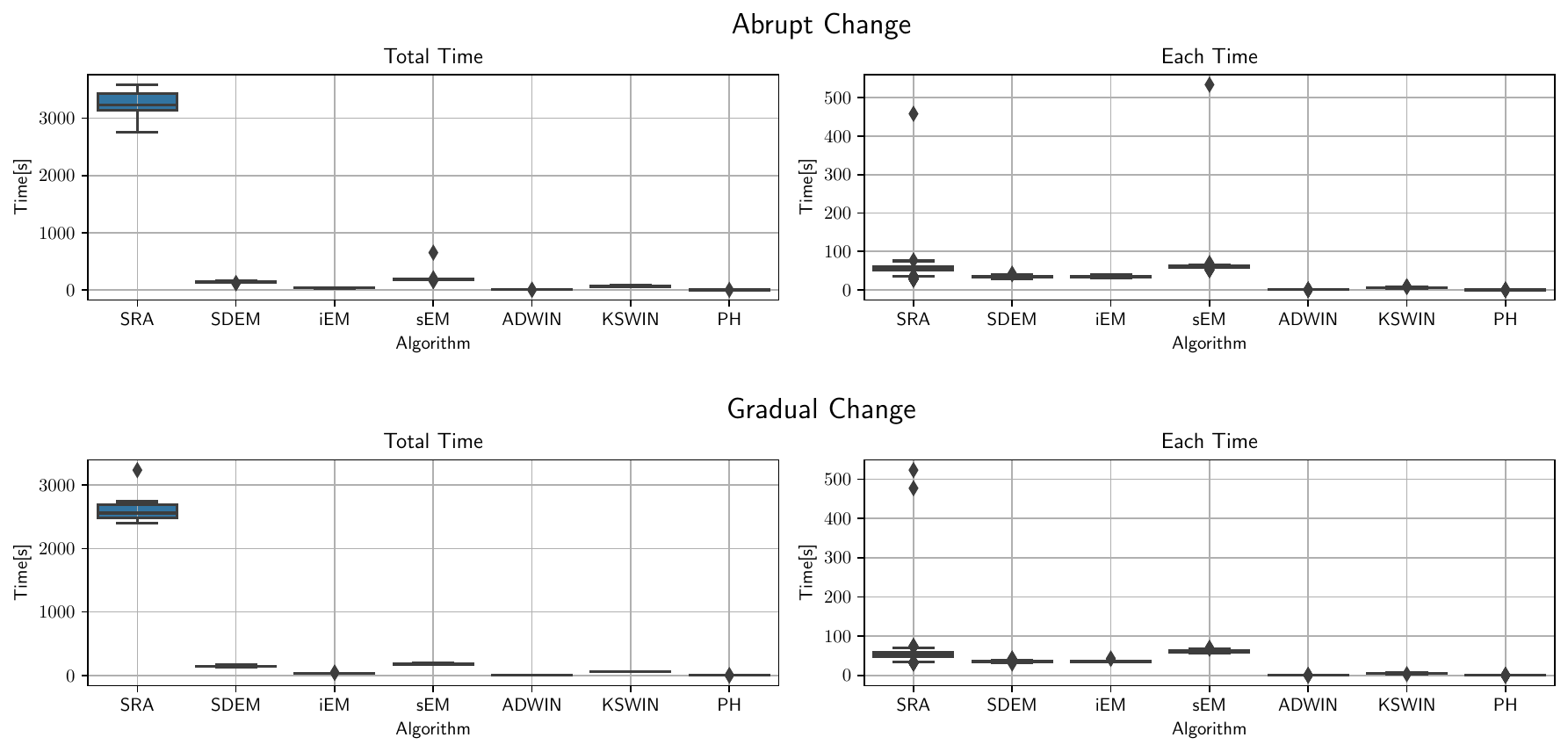}
\caption{Computation times for the algorithms on the synthetic univariate datasets with abrupt and gradual changes. The left-hand side of the figure displays the total time required for tuning the hyperparameters, 
whereas the right-hand side of the figure displays the time required for each set of hyperparameters.}
\label{fig:measured_times_for_the_algorithms_on_the_synthetic_univariate_daatasets}
\end{figure}

We further compared the algorithms described above in terms of computational cost.  
Figure~\ref{fig:measured_times_for_the_algorithms_on_the_synthetic_univariate_daatasets} displays 
the computation times for tuning hyperparameters 
for both datasets with abrupt and gradual changes. 
The left-hand side of Figure~\ref{fig:measured_times_for_the_algorithms_on_the_synthetic_univariate_daatasets} displays 
the total times required for tuning hyperparameters, 
whereas 
the right-hand side of  Figure~\ref{fig:measured_times_for_the_algorithms_on_the_synthetic_univariate_daatasets} displays 
each instance of time required by the set of hyperparameters for each algorithm. 
We repeatedly measured the computation time 
10 times. 
This result indicates that 
(i) SRA has a high computational cost in total 
because it has more hyperparameters to be tuned 
than other algorithms, 
and 
(ii) SRA and sEM have the highest computational cost 
when the algorithms are compared for a set of hyperparamters, 
followed by SDEM and iEM. 
In contrast, 
ADWIN, KSWIN, and PH 
require less computational cost. 
We conclude that 
the online learning algorithms 
have higher computational costs  
than the concept drift detection algorithms 
when tuning each set of hyperparameters. 
We attribute this to the fact that 
the online learning algorithms 
calculate probability densities 
of parametric distributions 
in calculating the change scores, 
which leads to high computational cost. 

In summary, 
SRA is superior to other algorithms in terms of accuracy 
in detecting changes 
for both datasets with abrupt and gradual changes. 
In contrast, 
SRA requires more time to tune the hyperparameters than rival algorithms 
because SRA has more hyperparameters. 
In addition, 
SRA has a relatively high computational cost 
because it calculates the probability densities of parametric distributions 
when calculating the change scores.

\subsection{Multivariate synthetic datasets}
\label{subsection:multivariate_synthetic_datasets}

We generated 
multivariate data streams  
with abrupt and gradual changes, 
which were generated from mixtures of 
true and noisy distributions. 
We focused on the comparison of SRA with the other algorithms 
described in 
Section~\ref{subsubsection:methods_for_comparison_univariate_synthetic_datasets}. 
The notations follow these in 
Section~\ref{subsection:univariate_synthetic_datasets} 
unless we specifically define them. 

\subsubsection{Datasets} 
We generated the following three-dimensional sequences:
\begin{align}
y_{t} \sim f = \alpha \, f_{1} + (1-\alpha) \, f_{2} 
\, \, \, 
(t = 1, \dots, 20000), 
\label{eq:data_generation_synthetic_dataset_three_dim}
\end{align}
where $f_{1}$ and $f_{2}$ denote the same as in Section~
\ref{subsubsection:univariate_synthetic_datasets_generation}. 
We generated the following two three-dimensional datasets with abrupt and gradual changes:
\begin{itemize}
    \item \textbf{Abrupt Change} \\
    We set $f_{1}$ and $f_{2}$ in Equation~\eqref{eq:data_generation_synthetic_dataset_three_dim} as follows:
    \begin{align}
    f_{1} &= \frac{1}{2} \mathcal{N}(y; \mu_{1}, \Sigma_{1}) +
         \frac{1}{2} \mathcal{N}(y; \mu_{2}, \Sigma_{2}), \, 
    f_{2} = \mathrm{Uniform}(y; -U, U), \\
    \mu &= \left(
        \begin{array}{c}
        \mu_{1} \\ 
        \mu_{2}
        \end{array}
      \right) 
    = \begin{cases}
      \transpose{
        (\transpose{(3, 2, 1)}, 
         \transpose{(-2, 3, 2)})
      } & (t \leq  10000),  \\
      \transpose{
        (\transpose{(6, 4, 2)},
         \transpose{(-4, 6, -4)})
      } & (10001 \leq t \leq 20000),  
    \end{cases} \\
    \Sigma_{1} &= \Sigma_{2} = 
    \left(
      \begin{matrix}
      1 & -0.8 & 0.2 \\
      -0.8 & 1 & 0.3 \\
      0.2 & 0.3 & 1
      \end{matrix}
    \right).   
    \label{eq:synthetic_multivariate_data_generation_with_abrupt_change}
    \end{align}
    These sequences have a change point at $t=10001$, 
    where the mean changes abruptly. 
    
    \item \textbf{Gradual Change} \\ 
    We set $f_{1}$ and $f_{2}$ in Equation~\eqref{eq:data_generation_synthetic_dataset_three_dim} as follows:
    \begin{align}
    f_{1} &= \frac{1}{2} \mathcal{N}(y; \mu_{1}, \Sigma_{1}) +
             \frac{1}{2} \mathcal{N}(y; \mu_{2}, \Sigma_{2}), \, 
    f_{2} = \mathrm{Uniform}(y; -U, U), \\
    \mu &= \left(
        \begin{array}{c}
        \mu_{1} \\ 
        \mu_{2}
        \end{array}
      \right) 
        = \begin{cases} 
          \transpose{
            (\transpose{(3, 4, 1)}, 
             \transpose{(-2, 3, -2)}
            )
          } & (t \leq  10000),  \\
          \frac{t-10000}{300}
          \transpose{
            (\transpose{(3, 4, 1)},
             \transpose{(-2, 3, -2)}
            )
          } & (10001 \leq t \leq 10300), \\
          \transpose{
            (\transpose{(6, 8, 2)},
             \transpose{(-4, 6, -4)}
            )
          } & (10301 \leq t \leq 20000),  
        \end{cases} \\
    \Sigma_{1} &= \Sigma_{2} = 
    \left(
      \begin{matrix}
      1 & -0.8 & 0.2 \\
      -0.8 & 1 & 0.3 \\
      0.2 & 0.3 & 1
      \end{matrix}
    \right). 
    \label{eq:synthetic_multivariate_data_generation_with_gradual_change}
    \end{align}
    These sequences have a change point at $t=10001$, 
    where the mean starts to change gradually up to  $t=10300$. 
\end{itemize}

\begin{figure}[H]
\centering
\includegraphics[width=\textwidth]{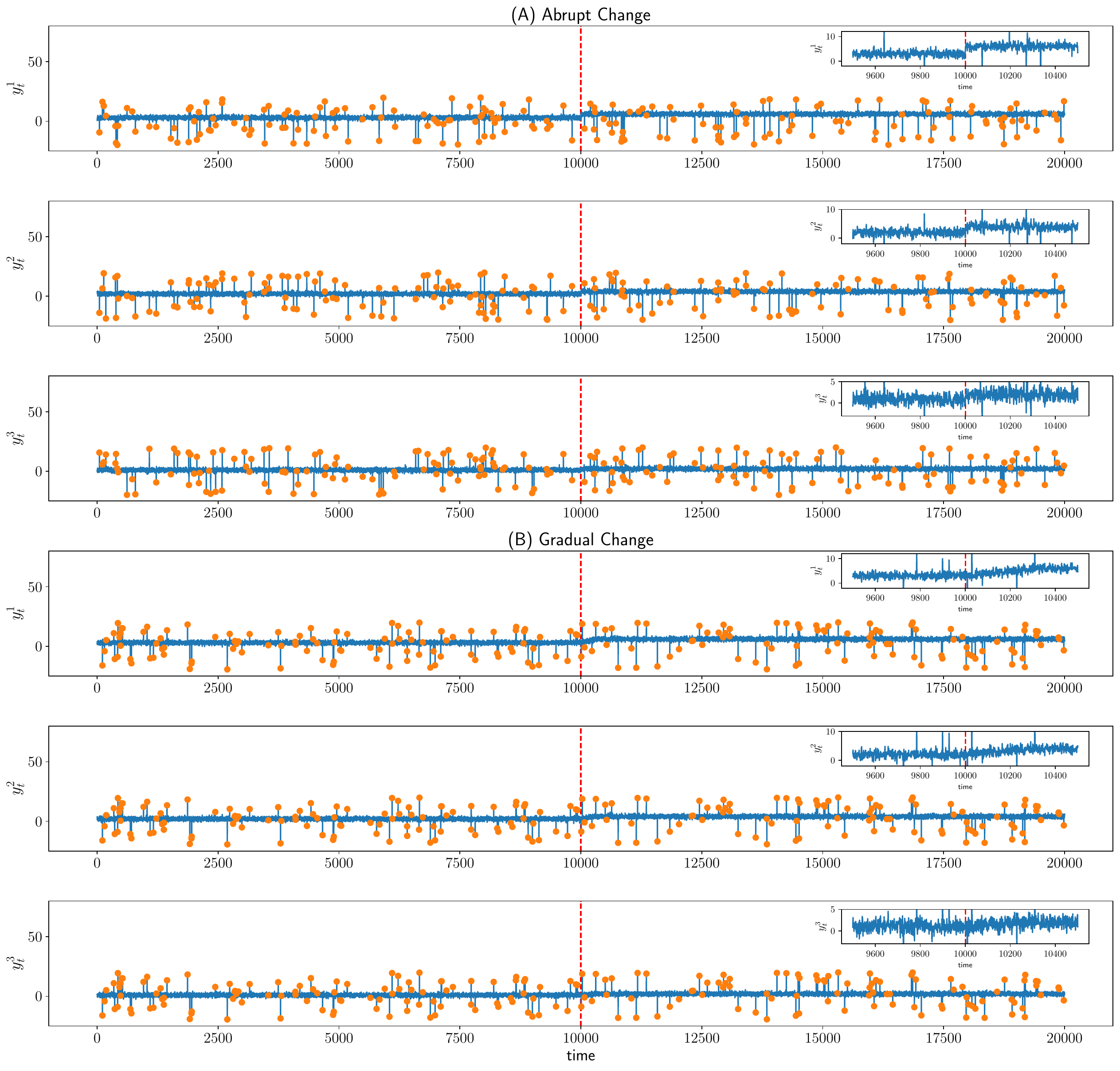}
\caption{Sample data streams of multivariate synthetic datasets with abrupt and gradual changes. 
Each data point 
$y_{t} = \transpose{(y_{t}^{1}, y_{t}^{2}, y_{t}^{3})}$ 
is drawn from 
Equation~\eqref{eq:synthetic_multivariate_data_generation_with_abrupt_change} 
and 
\eqref{eq:synthetic_multivariate_data_generation_with_gradual_change} 
for abrupt and gradual changes, 
respectively~($\alpha=0.99$ and $U=20$). 
Data points drawn from $f_{2}$ are marked with circles.  
Insets show the data streams between $t=9500$ and $t=10500$. 
(A) The data stream changes abruptly at $t=10001$,
and 
(B) The data stream changes gradually from $t=10001$ up to $t=10300$. 
The red dashed lines indicate the change points at $t=10001$.}
\label{fig:sample_data_streams_of_multivariate_synthetic_datasets_with_abrupt_and_gradual_changes}
\end{figure}

Figure~\ref{fig:sample_data_streams_of_multivariate_synthetic_datasets_with_abrupt_and_gradual_changes} 
illustrates sample data streams 
with abrupt and gradual changes. 
Each data point 
$y_{t} = \transpose{(y_{t}^{1}, y_{t}^{2}, y_{t}^{3})}$ 
is drawn from 
Equation~\eqref{eq:synthetic_multivariate_data_generation_with_abrupt_change} 
and 
\eqref{eq:synthetic_multivariate_data_generation_with_gradual_change} 
for abrupt and gradual changes, 
respectively. 
We set $\alpha=0.99$ and $U=20$. 
The data points drawn from $f_{2}$ are marked with circles, 
and 
we observe that most of the data points 
drawn from $f_{2}$ 
deviate from the ones drawn from $f_{1}$.

\subsubsection{Methods for Comparison}
\label{subsubsection:methods_for_comparison_multivariate_synthetic_datasets}

We compared the performance of SRA with those of rival algorithms 
described in  Section~\ref{subsubsection:methods_for_comparison_univariate_synthetic_datasets}:  
iEM~\citep{Neal1999}, 
sEM~\citep{Cappe2009}, 
ADWIN~\citep{Bifet2007}, 
KSWIN~\citep{Raab2020}, 
and PH~\citep{Page1954}. 

\subsubsection{Evaluation metrics}
\label{subsubsection:evaluation_metrics_multivariate_synthetic_datasets}

We used 
MSE 
and 
AUC 
as we did in 
Section~\ref{subsubsection:evaluation_metrics_univariate_synthetic_datasets}. 
Because ADWIN, KSWIN, and PH 
are designed for one-dimensional data streams, 
we calculated the change scores for each variable 
for these algorithms 
and then selected the variable 
that provided the best score.

\subsubsection{Result: Comparison with other algorithms}

We compared the performance of SRA with those of rival algorithms. 
We chose the hyperparameters 
with $S_{\mathrm{eval}}$ defined in 
Equation~\eqref{eq:def_mse_metrics} 
for each algorithm as follows: 
$\gamma \in \{ 1$, $3$, $5$, $7$, $10$, $15 \}$, 
$\beta \in \{ 0.1$, $0.5$, $1 \}$, 
and $M \in \{ 1$, $5$, $10 \}$ 
for SRA,
$r_{\mathrm{SDEM}} \in \{ 0.0001$, $0.001$, $0.005$, $0.01 \}$ 
for SDEM, 
$r_{\mathrm{sEM}} \in \{ 0.001$, $0.003$, $0.005 \}$ 
for sEM,  
$\delta_{\mathrm{ADWIN}} \in \{0.001$, $0.002$, $0.005$, 
$0.01$, $0.02$, $0.05$, $0.1$, $0.2$, $0.5 \}$ 
for ADWIN, 
$\alpha_{\mathrm{KSWIN}} \in \{0.001$, $0.005$, $0.01 \}$, 
$w_{\mathrm{KSWIN}} \in \{10$, $20 \}$, 
and 
$r_{\mathrm{KSWIN}} \in \{10$, $15 \}$ 
for KSWIN, 
and 
$\delta_{\mathrm{PH}} \in \{ 0.001$, $0.01$, $0.1$, $1$, $10 \}$ 
for PH. 

We first evaluated SRA and the other algorithms with MSE. 
We generated 10 data streams 
according to 
\eqref{eq:data_generation_synthetic_dataset_three_dim} 
with $\alpha=0.99$ and $U=20$ 
using Equation~\eqref{eq:synthetic_multivariate_data_generation_with_abrupt_change} 
and 
\eqref{eq:synthetic_multivariate_data_generation_with_gradual_change} 
for abrupt and gradual changes, 
respectively. 

Table~\ref{table:average_mses_on_the_multivariate_synthetic_datasets} shows 
the average MSEs $S_{\mathrm{bc}}$, $S_{\mathrm{ac}}$, 
and $S_{\mathrm{tot}}$ for each algorithm. 
The hyperparamters were chosen as follows: 
$(\gamma$, $\beta$, $M) = (7$, $0.5$, $5)$ for SRA, 
$r_{\mathrm{SDEM}} = 0.001$ for SDEM, 
$r_{\mathrm{sEM}} = 0.001$ for sEM. 
SRA was superior to other algorithms 
for the time periods after the change point and before it. 
This result indicates that SRA has higher robustness and adaptivity 
than other algorithms. 

\begin{table}[tb]
\begin{center}
\caption{Average MSEs on the multivariate synthetic datasets.}
\label{table:average_mses_on_the_multivariate_synthetic_datasets}
\begin{subtable}{\linewidth}
\caption{Abrupt Change}
\centering
{\tabcolsep=0.8\tabcolsep
\begin{tabular}{rrrr}
 \toprule
 & \multicolumn{1}{c}{$S_{\mathrm{bc}}$} 
 & \multicolumn{1}{c}{$S_{\mathrm{ac}}$}
 & \multicolumn{1}{c}{$S_{\mathrm{tot}}$}  \\
 \midrule
 \multicolumn{1}{c}{SRA} & 
 $\mathbf{0.026 \pm 0.002}$ & 
 $\mathbf{0.029 \pm 0.001}$ & 
 $\mathbf{0.026 \pm 0.002}$ \\
 \multicolumn{1}{c}{SDEM} & 
 $1.437 \pm 0.002$ & 
 $5.812 \pm 0.005$ & 
 $3.894 \pm 0.003$ \\
 \multicolumn{1}{c}{iEM} & 
 $1.473 \pm 0.001$ & 
 $5.836 \pm 0.001$ & 
 $3.527 \pm 0.001$ \\
 \multicolumn{1}{c}{sEM} & 
 $1.399 \pm 0.002$ & 
 $5.876 \pm 0.003$ & 
 $3.684 \pm 0.002$ \\
 \bottomrule
\end{tabular}
}
\end{subtable}
\hfil
\begin{subtable}{\linewidth}
\caption{Gradual Change}
\centering
{\tabcolsep=0.8\tabcolsep
\begin{tabular}{rrrr}
 \toprule
 & \multicolumn{1}{c}{$S_{\mathrm{bc}}$} 
 & \multicolumn{1}{c}{$S_{\mathrm{ac}}$}
 & \multicolumn{1}{c}{$S_{\mathrm{tot}}$}  \\
 \midrule
 \multicolumn{1}{c}{SRA} & 
 $\mathbf{0.016 \pm 0.002}$ & 
 $\mathbf{0.007 \pm 0.001}$ & 
 $\mathbf{0.003 \pm 0.000}$ \\
 \multicolumn{1}{c}{SDEM} & 
 $2.832 \pm 0.810$ & 
 $7.018 \pm 2.653$ & 
 $2.763 \pm 0.793$ \\
 \multicolumn{1}{c}{iEM} & 
 $0.659 \pm 0.596$ & 
 $2.675 \pm 1.513$ & 
 $1.514 \pm 1.029$ \\
 \multicolumn{1}{c}{sEM} & 
 $0.083 \pm 0.007$ & 
 $0.877 \pm 0.025$ & 
 $0.859 \pm 0.019$ \\
 \bottomrule
\end{tabular}
}
\end{subtable}
\end{center}
\end{table} 

Next, 
we evaluated each algorithm with AUC. 
Table~\ref{table:auc_comparison_for_synthetic_multivariate_datasets} shows 
the average AUCs for each algorithm. 
SRA was superior to other algorithms. 
The hyperparametes were chosen as follows: 
(i) Abrupt change: 
$(\gamma$, $\beta$, $M) = (7$, $0.5$, $5)$ 
for SRA, 
$r_{\mathrm{SDEM}} = 0.005$ 
for SDEM, 
$r_{\mathrm{sEM}} = 0.005$ 
for sEM, 
$\delta_{\mathrm{ADWIN}} = 0.5$ 
for ADWIN, 
$(\alpha_{\mathrm{KSWIN}}$, $w_{\mathrm{KSWIN}}$, 
$r_{\mathrm{KSWIN}})=(0.005$, 
$10$, $15)$ 
for KSWIN, 
and 
$\delta_{\mathrm{PH}} = 50$ 
for PH. 
(ii) Gradual change: 
$(\gamma$, $\beta$, $M) = (3$, $0.5$, $5)$ 
for SRA, 
$r_{\mathrm{SDEM}} = 0.001$ 
for SDEM, 
$r_{\mathrm{sEM}} = 0.001$ 
for sEM, 
$\delta_{\mathrm{ADWIN}} = 0.5$ 
for ADWIN, 
$(\alpha_{\mathrm{KSWIN}}$, $w_{\mathrm{KSWIN}})=(0.005$, 
$10)$ 
for KSWIN, 
and 
$\delta_{\mathrm{PH}} = 1$ 
for PH. 

\begin{table}
\begin{center}
\caption{Average AUCs on the synthetic multivariate datasets.}
\label{table:auc_comparison_for_synthetic_multivariate_datasets}
{\tabcolsep=0.8\tabcolsep
\begin{tabular}{rrr}
 \toprule
 & \multicolumn{1}{c}{Abrupt Change} 
 & \multicolumn{1}{c}{Gradual Change} \\
 \midrule
 \multicolumn{1}{c}{SRA} 
 & $\mathbf{0.753 \pm 0.015}$ 
 & $\mathbf{0.721 \pm 0.019}$  \\
 \multicolumn{1}{c}{SDEM} 
 & $0.551 \pm 0.009$ 
 & $0.532 \pm 0.051$  \\
 \multicolumn{1}{c}{iEM} 
 & $0.542 \pm 0.008$ 
 & $0.357 \pm 0.028$  \\
 \multicolumn{1}{c}{sEM} 
 & $0.612 \pm 0.091$ 
 & $0.527 \pm 0.089$  \\
 \multicolumn{1}{c}{ADWIN}  
 & $0.500 \pm 0.000$
 & $0.500 \pm 0.000$ \\
 \multicolumn{1}{c}{KSWIN}    
 & $0.538 \pm 0.052$
 & $0.515 \pm 0.062$ \\
 \multicolumn{1}{c}{PH}       
 & $0.482 \pm 0.183$
 & $0.459 \pm 0.139$ \\
 \bottomrule
\end{tabular}
}
\end{center}
\end{table} 

In summary, 
SRA is superior to other algorithms in terms of accuracy 
in detecting changes for both three-dimensional datasets 
with abrupt and gradual changes. 
Therefore, 
we conclude that 
SRA has higher robustness adaptivity than other algorithms 
for multivariate data streams. 

\subsection{Real dataset:~Change detection}
\label{subsection:real_dataset_change_detection}

We applied SRA to 
the Well-log dataset~\citep{Ruanaidh1996} 
and 
the SKoltech Anomaly Benchmark~(SKAB) dataset~\citep{Katser2020}. 
The Well-log dataset is composed of a one-dimensional sequence,
and 
the SKAB dataset is composed of eight-dimensional sequences. 

\subsubsection{Well-log dataset}
\label{sububsection:real_dataset_change_detection_welllog}

The Well-log dataset is a one-dimensional data stream 
consisting of 4050 nuclear magnetic resonance measurements 
during the drilling of a well. 
Since it was first studied~\citep{Ruanaidh1996}, 
it has become a benchmark dataset for univariate change detection. 
Although this dataset has been used in several studies
(e.g., \citep{Adams2007,Levyleduc2008,Ruggieri2016,Fearnhead2019}),
the outliers have often been removed before change detection, 
with the exception of only a few studies (e.g.,  \citep{Fearnhead2019}).

\begin{figure}[tb]
\begin{center}
\includegraphics[width=\textwidth]{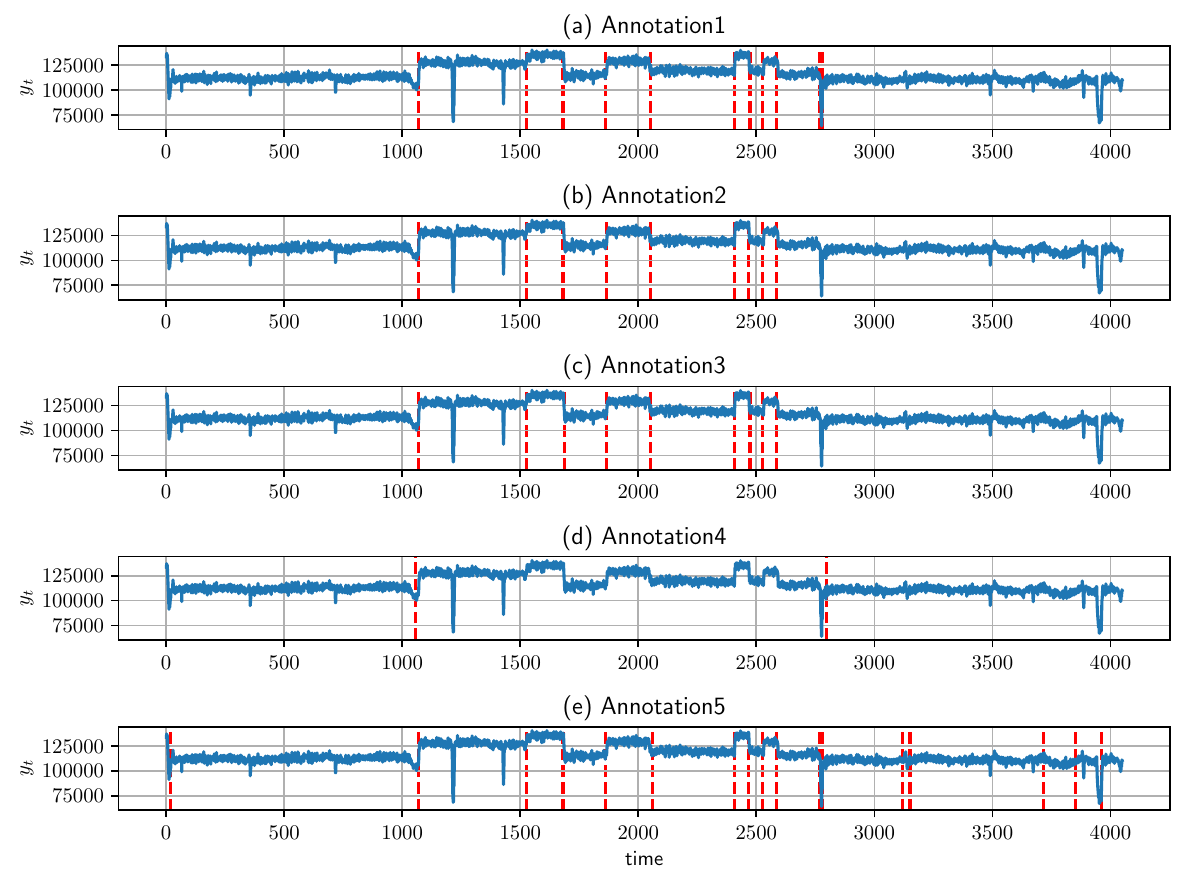}
\caption{Plot of the data stream and 
five sets of annotations of the change points for the Well-log dataset \citep{Burg2020}. 
The red dashed lines indicate change points.}
\label{fig:experiment4_plot_annotations}
\end{center}
\end{figure}

We applied SRA to the Well-log dataset for change detection. 
This dataset is available at \url{https://github.com/alan-turing-institute/rbocpdms/}. 
Figure~\ref{fig:experiment4_plot_annotations} shows
the annotated change points proposed by 
\citep{Burg2020}.
There are five sets of annotated changes, 
each provided by an annotator:
\begin{itemize}
    \item Annotation~1: $t=1069$, $1525$, $1681$, $1861$, $2053$, $2407$, $2473$, 
    $2527$, $2587$, $2767$, $2779$. 
    \item Annotation~2: $t=1069$, $1525$, $1681$, $1867$, $2053$, $2407$, $2467$, $2527$, $2587$. 
    \item Annotation~3: $t=1069$, $1525$, $1687$, $1867$, $2053$, $2407$, $2473$, $2527$, $2587$.
    \item Annotation~4: $t=1057$, $2797$.
    \item Annotation~5: $t=19$, $1069$, $1525$, $1681$, $1861$, $2059$, $2407$, $2467$, $2527$, $2587$, $2767$,
    $2779$, $3121$, $3151$, $3715$, $3853$, $3961$.
\end{itemize}

We used the first 1550 data points as the training dataset
and
the remaining points as the test dataset.
We chose SDEM~\citep{Yamanishi2004},
iEM~\citep{Neal1999},
sEM~\citep{Cappe2009}, 
ADWIN~\citep{Bifet2007}, 
KSWIN~\citep{Raab2020}, 
and PH~\citep{Page1954} 
for comparison.
SDEM, iEM, and sEM 
are online learning algorithms, 
and we employed the univariate normal distribution for these algorithms. 
ADWIN, KSWIN, and PH 
are concept drift detection algorithms. 
We calculated the change score $s_{t}$ 
for each algorithm in the same way as in 
Section~\ref{subsubsection:evaluation_metrics_univariate_synthetic_datasets}. 

We chose the hyperparameters of each algorithm with AUC scores between $t_{\mathrm{start}} = 20$ and $t_{\mathrm{end}} = 1550$: the hyperparameters of SRA were chosen among 
$\gamma \in \{ 2\times 10^{6}$, $3\times 10^{6}$, $4\times 10^{6} \}$,
$\beta = (d_{0} + 1)/L(1-\alpha) \in \{ 0.01 \gamma$, $0.03 \gamma$, $0.05 \gamma \}$,
and $M \in \{ \gamma$, $2\gamma$, $3\gamma \}$, 
and those of SDEM and sEM were chosen 
among 
$r_{\mathrm{SDEM}}, r_{\mathrm{sEM}} \in \{ 0.001$, $0.003$, $0.005$, $0.01$, $0.03$, $0.05$, $0.1 \}$. 
In contrast, 
the hyperparameters of 
the concept drift detection algorithms were chosen as follows: 
for ADWIN, 
$\delta_{\mathrm{ADWIN}} \in \{ 0.001$, $0.002$, $0.005$, $0.01$, $0.02$, $0.05$, $0.1$, $0.2$, $0.5 \}$, 
$\alpha_{\mathrm{KSWIN}} \in \{ 0.001$, 
$0.005$, $0.01\}$, 
$w_{\mathrm{KSWIN}} \in \{10$, $20\}$, 
and 
$r_{\mathrm{KSWIN}} \in \{10$, $15\}$ 
for KSWIN, 
and 
$\delta_{\mathrm{PH}} \in \{ 0.001$, 
$0.01$, 
$0.1$, 
$1$, 
$10 \}$ 
for PH. 
We initialized each parameter or sufficient statistics of each algorithm 10 times and selected the combination that provided the best performance on average. To initialize the parameter or sufficient statistics, we drew 20 initial points from the uniform distribution with a range of $[\min(y_{20}^{40}), \max(y_{20}^{40})]$, 
where $y_{20}^{40} = y_{20} \dots y_{40}$ is the sequence between $t=20$ and $t=40$. 

\begin{figure}[tb]
\centering
\includegraphics[width=\linewidth]{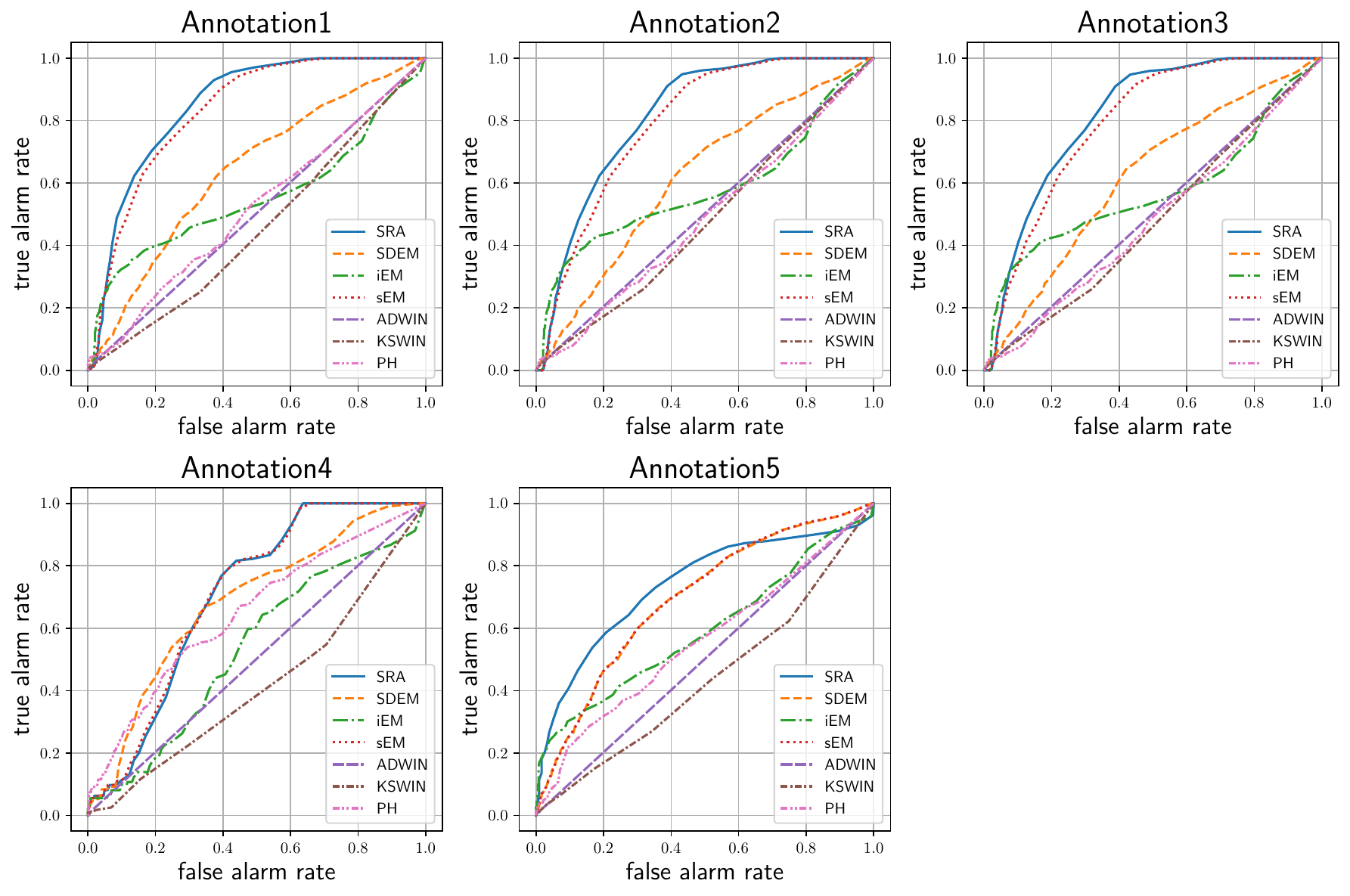}
\caption{ROC curves on the Well-log dataset for SRA, SDEM, iEM, sEM, and ADWIN.}
\label{fig:plot_roc_welllog}
\end{figure}

We applied each algorithm and calculated 
the AUC scores 
on the test dataset after the parameters were
determined on the training dataset. 
Figure~\ref{fig:plot_roc_welllog} 
displays 
the ROC curves
on the test dataset with the algorithms and annotations. 
We observe that:  
(i) When the false alarm rate is low, 
SRA is not better than iEM 
but is overwhelmingly better than the other algorithms 
when the false alarm rate is over $0.1$ 
for Annotation~1, 2, 3, 
and between $0.1$ and $0.6$ 
for Annotation~5. 
(ii) SRA is almost as good as sEM  
for Annotation~4. 
We infer that (ii) is due to 
Annotation~4 having only two change points,  
and that SRA and sEM detected all the change points in a similar way. 

Then, 
we calculated 
the AUC scores. 
Table~\ref{table:experiment4_result_change_detection} 
displays 
the AUCs 
for the algorithms. 
Because the estimated AUCs were the same 
for all the algorithms 
except SDEM, 
the most of standard deviations were zero. 
The ROC curves and AUCs were calculated 
in the same way as in 
Section~\ref{subsubsection:evaluation_metrics_univariate_synthetic_datasets}. 
We observe that SRA is superior to other algorithms
for each annotation.
The best combinations of hyperparameters were chosen as follows: 
for SRA, 
$(\gamma, \beta, M) = (2\times 10^{6}, 2\times 10^{4}, 4 \times 10^{6})$, 
$(3\times 10^{6}, 3 \times 10^{4}, 6\times 10^{6})$,  
$(4\times 10^{6}, 4\times 10^{4}, 8\times 10^{6})$ for Annotation~1, 2, and 3, 
$(\gamma, \beta, M) = (3 \times 10^{6}, 3\times 10^{4}, 12\times 10^{6})$, 
$(2 \times 10^{6}, 2 \times 10^{4}, 8 \times 10^{6})$, 
$(4 \times 10^{6}, 4 \times 10^{4}, 16 \times 10^{6})$
for Annotation~4, 
and
$(\gamma, \beta, M) = (2 \times 10^{6}, 2\times 10^{4}, 4\times 10^{6})$, 
$(3 \times 10^{6}, 3 \times 10^{4}, 6 \times 10^{6})$, 
$(4 \times 10^{6}, 4 \times 10^{4}, 8 \times 10^{6})$ 
for Annotation~5. 
For SDEM, 
$r_{\mathrm{SDEM}} = 0.05$ for all the annotations. 
For sEM, 
$r_{\mathrm{sEM}} = 0.003$ (Annotation~1, 2, and 3) 
$r_{\mathrm{sEM}} = 0.005$ (Annotation~4), 
and $r_{\mathrm{sEM}} = 0.05$ (Annotation~5). 
For ADWIN, 
$\delta_{\mathrm{ADWIN}} = 0.05$ 
for all the annotations.  
For KSWIN,
$(\alpha_{\mathrm{KSWIN}}$, $w_{\mathrm{KSWIN}}$, 
$r_{\mathrm{KSWIN}})=(0.005$, 
$30$, $15)$ for Annotation~1, 2, 
$(0.005$, $30$, $10)$ for Annotation~3, 
$(0.001$, $30$, $15)$ for Annotation~4, 
and 
$(0.001$, $30$, $10)$ for Annotation~5. 
For PH, 
$\delta_{\mathrm{PH}}=1$ 
for all the annotations.  

\begin{table}[tb]
\begin{center}
\caption{AUC scores on the test dataset of the Well-log dataset.}
{\tabcolsep=0.8\tabcolsep
\begin{tabular}{p{3em}p{6.5em}p{6.5em}p{6.5em}p{6.5em}p{6.5em}}
\toprule
  & 
  \multicolumn{1}{c}{Annotation~1} &
  \multicolumn{1}{c}{Annotation~2} &
  \multicolumn{1}{c}{Annotation~3} &
  \multicolumn{1}{c}{Annotation~4} &
  \multicolumn{1}{c}{Annotation~5} \\
\midrule
\multicolumn{1}{c}{SRA}  & 
  \multicolumn{1}{c}{$\mathbf{0.802 \pm 0.000}$} & 
  \multicolumn{1}{c}{$\mathbf{0.771 \pm 0.000}$} &
  \multicolumn{1}{c}{$\mathbf{0.772 \pm 0.000}$} & 
  \multicolumn{1}{c}{$\mathbf{0.700 \pm 0.000}$} &
  \multicolumn{1}{c}{$\mathbf{0.708 \pm 0.000}$} \\
\multicolumn{1}{c}{SDEM} & 
     \multicolumn{1}{c}{$0.629 \pm 0.003$} & 
     \multicolumn{1}{c}{$0.606 \pm 0.006$} & 
     \multicolumn{1}{c}{$0.608 \pm 0.006$} & 
     \multicolumn{1}{c}{$0.673 \pm 0.008$} &
     \multicolumn{1}{c}{$0.672 \pm 0.005$} \\
\multicolumn{1}{c}{iEM}  & 
  \multicolumn{1}{c}{$0.535 \pm 0.000$} & 
  \multicolumn{1}{c}{$0.553 \pm 0.000$} & 
  \multicolumn{1}{c}{$0.546 \pm 0.000$} & 
  \multicolumn{1}{c}{$0.538 \pm 0.000$} &
  \multicolumn{1}{c}{$0.578 \pm 0.000$} \\
\multicolumn{1}{c}{sEM}  & 
  \multicolumn{1}{c}{$0.779 \pm 0.000$} & 
  \multicolumn{1}{c}{$0.747 \pm 0.000$} & 
  \multicolumn{1}{c}{$0.747 \pm 0.000$} & 
  \multicolumn{1}{c}{$0.699 \pm 0.000$} &
  \multicolumn{1}{c}{$0.579 \pm 0.000$}  \\
\multicolumn{1}{c}{ADWIN} &
  \multicolumn{1}{c}{$0.503 \pm 0.000$} &
  \multicolumn{1}{c}{$0.502 \pm 0.000$} &
  \multicolumn{1}{c}{$0.502 \pm 0.000$} &
  \multicolumn{1}{c}{$0.502 \pm 0.000$} &
  \multicolumn{1}{c}{$0.501 \pm 0.000$}  \\
\multicolumn{1}{c}{KSWIN} &
  \multicolumn{1}{c}{$0.486 \pm 0.000$} &
  \multicolumn{1}{c}{$0.525 \pm 0.000$} &
  \multicolumn{1}{c}{$0.475 \pm 0.000$} &
  \multicolumn{1}{c}{$0.415 \pm 0.000$} &
  \multicolumn{1}{c}{$0.513 \pm 0.000$}  \\
\multicolumn{1}{c}{PH} &
  \multicolumn{1}{c}{$0.510 \pm 0.000$} &
  \multicolumn{1}{c}{$0.492 \pm 0.000$} &
  \multicolumn{1}{c}{$0.490 \pm 0.000$} &
  \multicolumn{1}{c}{$0.549 \pm 0.000$} &
  \multicolumn{1}{c}{$0.527 \pm 0.000$}  \\
\bottomrule
\end{tabular}
}
\label{table:experiment4_result_change_detection}
\end{center}
\end{table}

\begin{figure}[tb]
\begin{center}
\includegraphics[width=\linewidth]{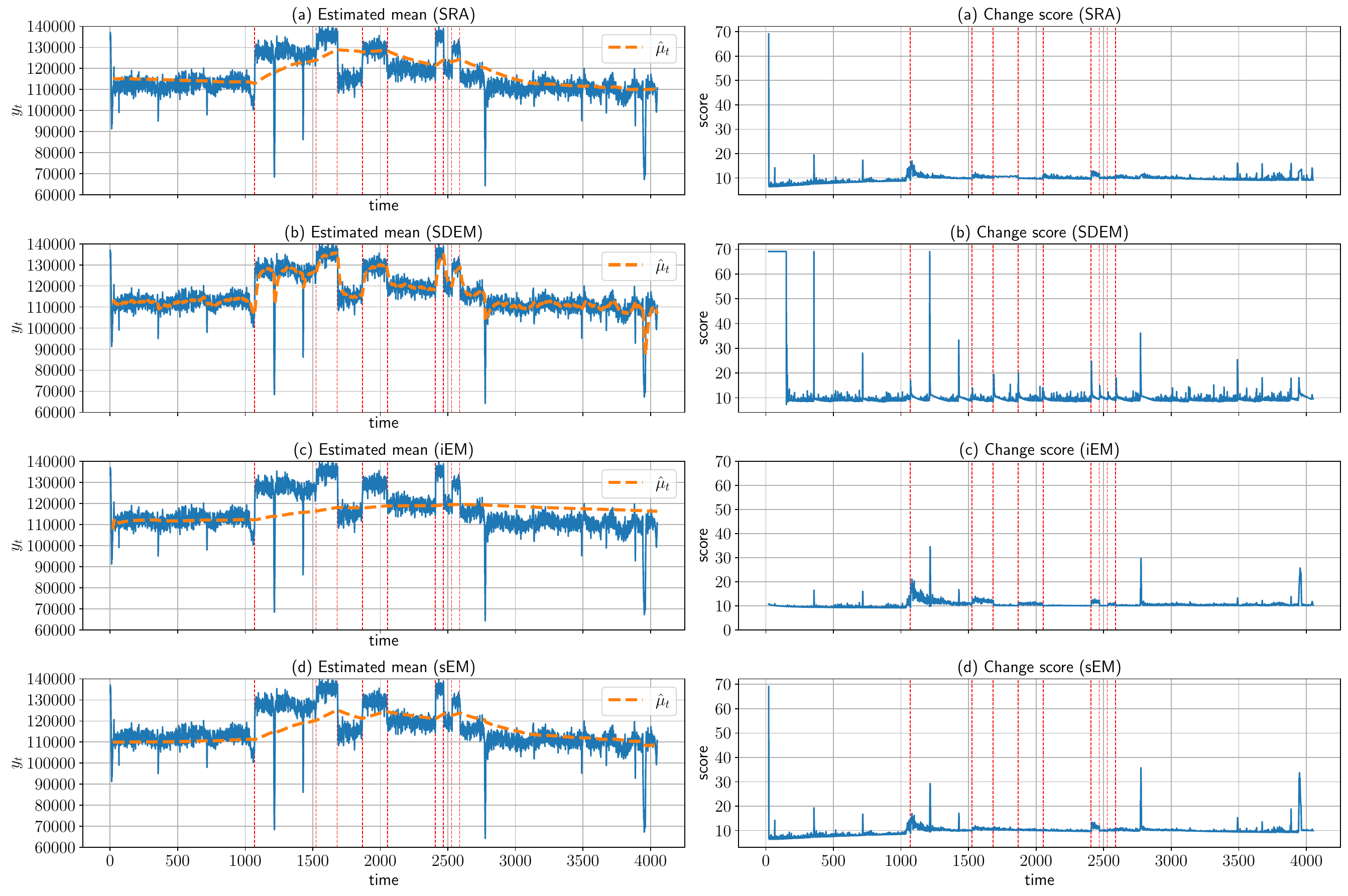}    
\caption{Estimated mean $\hat{\mu}_{t}$ and change scores for the Well-log dataset. 
The left and right panels display 
the estimated mean and change score 
for each algorithm, 
respectively. 
The red dashed lines indicate 
the change points for Annotation~2.}
\label{fig:estimated_mean_and_scores_for_the_well-log_dataset}
\end{center}
\end{figure}

We further investigated how well each algorithm estimated the change scores. 
We chose SRA, SDEM, iEM, and sEM 
because we could estimate the parameters for these online learning algorithms. 
Figure~\ref{fig:estimated_mean_and_scores_for_the_well-log_dataset} displays 
the estimated mean in the left panel 
and change scores in the right panel, 
respectively. 
The red dashed lines indicate 
the change points for Annotation~2. 
The hyperparameters were set as the best combinations for each algorithm 
for Annotation~2: $r_{\mathrm{SDEM}} = 0.05$ for SDEM, 
$r_{\mathrm{sEM}} = 0.003$ for sEM, 
and 
$(\gamma$, $M$, $\beta) = (2 \times 10^{4}$, 
$4 \times 10^{4}$, 
$2 \times 10^{2} )$~($\rho = 0.0039$) for SRA. 
As for the robustness and adaptivity, 
we observe the following points 
from the left panel of Figure~\ref{fig:estimated_mean_and_scores_for_the_well-log_dataset}:
\begin{itemize}
    \item SRA and sEM were robust to changes. 
    \item SDEM was adaptive to changes in the underlying data generating mechanism. 
    Therefore, it was overfitted. 
    \item iEM was robust (and not adaptive) to changes. 
\end{itemize}
As for the estimated change scores, 
we also observe from the right panel of Figure~\ref{fig:estimated_mean_and_scores_for_the_well-log_dataset} 
that SDEM, iEM, and sEM were prone to outliers, 
in particular after $t=2500$, 
whereas SRA was less influenced by outliers. 
On the other hand, 
for time points between $t=300$ and $t=1100$, 
it is hard to distinguish between the outliers and essential changes 
based only on the change scores estimated with SRA. 

\subsubsection{SKAB dataset}
\label{subsubsection:real_dataset_change_detection_skab}

The SKAB dataset is composed of 
34 eight-dimensional data streams~\citep{Katser2020}, 
which were collected from a testbed of a water circulation system. 
This dataset is available at \url{https://www.kaggle.com/dsv/1693952}. 

Anomalies and 
change points are annotated 
for each sequence. 
Here, 
the anomalies and 
change points refer to 
time points 
at which the machines operated abnormally 
and 
time points 
at which the mode of abnormality changed, 
respectively. 

\begin{figure}[tb]
\begin{center}
\includegraphics[width=\textwidth] {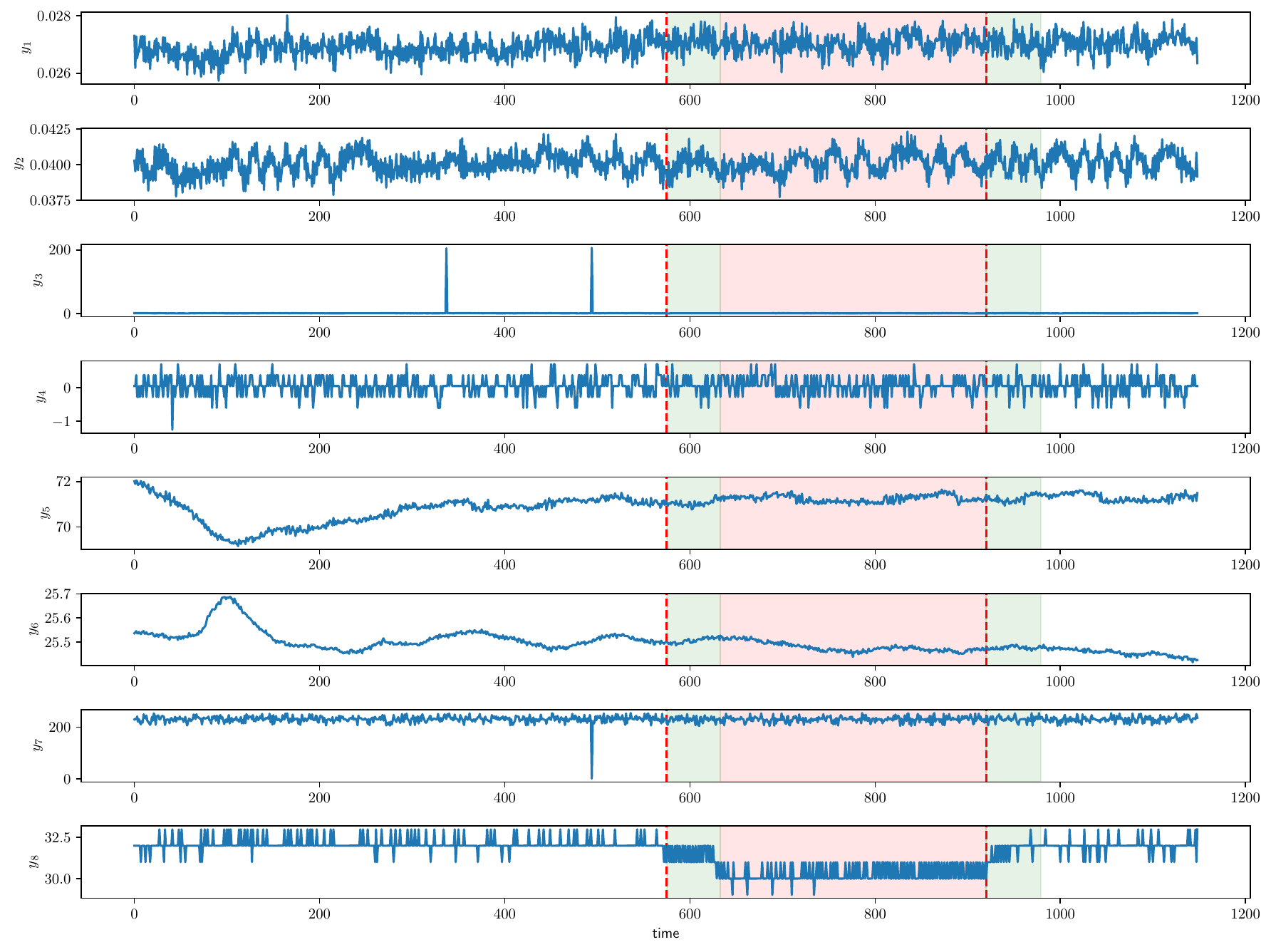}
\caption{A sample sequence of the SKAB dataset. 
The machine changed gradually from normal mode at $t=574$ 
and reached abnormal mode at $t=632$
~(the first span indicated by a light green color). 
Next, 
the machine remained at abnormal mode 
between $t=633$ and $918$
~(the span indicated by the light red color). 
Finally, 
the machine changed gradually from abnormal mode 
at $t=919$ 
and reached normal mode at $t=978$
~(the second span indicated by a light green color). 
Therefore, 
we annotated the change points at $t=745$ and $924$, 
as shown with red dashed lines. 
}
\label{fig:plot_skab_valve1_0}
\end{center}
\end{figure}

We chose 16 sequences obtained from the experiments, 
where the valve at the outlet of the flow from the pump was closed. 
Figure~\ref{fig:plot_skab_valve1_0} shows a sample sequence 
of the SKAB dataset. 
The machine started to change gradually at $t=574$ 
from normal mode 
and reached abnormal mode at $t=632$
~(the first span indicated by a green color). 
In contrast, 
the machine started to change gradually at $t=919$ 
from abnormal mode 
and reached normal mode at $t=978$
~(the second span indicated by green color). 
The machine remained in abnormal mode 
between $t=633$ and $t=918$
~(the span indicated by red). 
Therefore, 
we defined the change points 
used for evaluating SRA and the other algorithms 
at $t=574$ and $919$
~(the vertical red dashed lines). 
We clearly observe that 
there are anomaly points of 
$y_{3}$ at $t=338$ and $t=495$, 
and 
$y_{7}$ at $t=495$. 

We compared the performance of SRA 
with 
those of 
SDEM~\citep{Yamanishi2004}, 
iEM~\citep{Neal1999}, 
sEM~\citep{Cappe2009}, 
ADWIN~\citep{Bifet2007}, 
KSWIN~\citep{Raab2020}, 
and PH~\citep{Page1954}. 
We calculated the change score $s_{t}$ for each algorithm 
in the same way as in  Section~\ref{subsubsection:evaluation_metrics_univariate_synthetic_datasets}. 

We chose the hyperparameters of 
each algorithm with AUC scores between $t=20$ and 
the end time point. 
The hyperparameters were selected as follows: 
for SRA, 
$\gamma \in \{ 80$, $90$, $100 \}$, 
$\beta \in \{ 1$, $5$, $10 \}$, 
and 
$M \in \{ 10000$, $20000$, $30000 \}$. 
For SDEM, 
$r_{\mathrm{SDEM}} \in \{ 0.01$, $0.03$, $0.05$, $0.1$, $0.3$, $0.5 \}$.  
For sEM, 
$r_{\mathrm{sEM}} \in \{ 0.01$, $0.03$, $0.05$, $0.1$, $0.3$, $0.5 \}$.  
For ADWIN, 
$\delta_{\mathrm{ADWIN}} \in \{ 0.001$, $0.002$, $0.005$, 
$0.01$, $0.02$, $0.05$, $0.1$, $0.2$, $0.5 \}$.  
For KSWIN, 
$\alpha_{\mathrm{KSWIN}} \in \{ 0.001$, $0.005$, $0.01$, 
$w_{\mathrm{KSWIN}} \in \{ 20$, $30 \}$, 
and 
$r_{\mathrm{KSWIN}} \in \{ 10$, $15 \}$.  
For PH, 
$\delta_{\mathrm{PH}} \in \{ 0.01$, $0.05$, $0.1$, $0.5$, 
$1$, $5$, $10 \}$. 
For ADWIN, KSWIN, and PH, 
we calculated AUCs for each univariate sequence 
and selected the one which gave the best AUC score. 
We set the number of clusters $K=1$ 
for SRA, SDEM, iEM, and sEM. 

We used eight sequences labeled odd indexes 
as the training dataset 
and the remaining eight sequences labeled even indexes 
as the test dataset. 
We converted each variable 
$y_{t}^{i} \leftarrow 
 (y_{t}^{i} - \hat{\mu}^{i})/\hat{\sigma}^{i}$ \, 
 ($i=1, \dots, 8$), 
where 
$\hat{\mu}^{i}$ 
and 
$\hat{\sigma}^{i}$ 
were the estimated 
mean and standard deviation 
using the first 100 data points. 
We initialized the parameters of 
SRA, SDEM, iEM, and sEM using data points between 
$t = 20$ and $40$. 
We repeated this procedure 5 times. 

Table~\ref{table:auc_scores_on_skab_dataset} shows the AUCs 
on the test dataset 
for all the algorithms. 
We set the maximum tolerance delay 
$T_{\mathrm{b}}=20$ and $50$ 
in Equation~\eqref{eq:def_benefit}. 
We observe that SRA is superior to the other algorithms 
for both the values of $T_{\mathrm{b}}$.  
The best combinations of hyperparameters were selected 
as follows: 
$(\gamma$, $\beta$, $M) = (80$, $5$, $50000)$ 
for SRA, 
$r_{\mathrm{SDEM}} = 0.03$ 
for SDEM, 
$r_{\mathrm{sEM}} = 0.03$ 
for sEM, 
$\delta_{\mathrm{ADWIN}} = 0.1$ 
and the fifth variable 
for ADWIN, 
$(\alpha_{\mathrm{KSWIN}}$, 
$w_{\mathrm{KSWIN}}$, 
$r_{\mathrm{KSWIN}}) = (0.005$, 
$30$, 
$10)$ 
and the eighth variable 
for KSWIN, 
and 
$\delta_{\mathrm{PH}}=10$ 
and the third variable for PH.

\begin{table}[tb]
\begin{center}
\caption{AUC scores on the test dataset of the SKAB dataset. $T_{\mathrm{b}}$ is the maximum tolerant delay defined in Equation~\eqref{eq:def_benefit}.}
{\tabcolsep=0.5\tabcolsep
\begin{tabular}{p{3em}p{6.5em}p{6.5em}}
\toprule
  & 
  \multicolumn{1}{c}{$T_{\mathrm{b}}=20$}
  &
  \multicolumn{1}{c}{$T_{\mathrm{b}}=50$} \\
\midrule
\multicolumn{1}{c}{SRA}  & 
  \multicolumn{1}{c}{$\mathbf{0.551 \pm 0.043}$} &
  \multicolumn{1}{c}{$\mathbf{0.580 \pm 0.085}$} \\
\multicolumn{1}{c}{SDEM} & 
     \multicolumn{1}{c}{$0.543 \pm 0.135$} &
     \multicolumn{1}{c}{$0.565 \pm 0.119$} \\
\multicolumn{1}{c}{iEM}  & 
  \multicolumn{1}{c}{$0.529 \pm 0.125$} &
  \multicolumn{1}{c}{$0.544 \pm 0.078$} \\
\multicolumn{1}{c}{sEM}  & 
  \multicolumn{1}{c}{$0.537 \pm 0.078$}  &
  \multicolumn{1}{c}{$0.498 \pm 0.055$}  \\
\multicolumn{1}{c}{ADWIN} &
  \multicolumn{1}{c}{$0.505 \pm 0.012$}  &
  \multicolumn{1}{c}{$0.512 \pm 0.005$}  \\
\multicolumn{1}{c}{KSWIN} &
  \multicolumn{1}{c}{$0.536 \pm 0.085$}  &
  \multicolumn{1}{c}{$0.459 \pm 0.059$}  \\
\multicolumn{1}{c}{PH} &
  \multicolumn{1}{c}{$0.422 \pm 0.116$}  &
  \multicolumn{1}{c}{$0.518 \pm 0.166$}  \\
\bottomrule
\end{tabular}
}
\label{table:auc_scores_on_skab_dataset}
\end{center}
\end{table}

In summary, 
SRA outperformed the other online learning 
algorithms 
and 
concept drift detection algorithms 
for a univariate data stream with abrupt changes 
as well as 
multivariate data streams  
with gradual changes. 
Although this empirical study 
shows the performance of SRA, 
it remains future work 
to present 
theoretical analysis of SRA for 
data streams with gradual changes 
beyond the one presented in Section~\ref{section:convergence_analysis_of_bounded_stochastic_update}.

\subsection{Real dataset:~Anomaly detection}
\label{subsection:real_dataset_anomaly_detection} 

We applied SRA to anomaly detection in two real datasets: 
the SMTP and THYROID dataset. 
Both datasets are publicly available at \url{http://odds.cs.stonybrook.edu/smtp-kddcup99-dataset}. 
Table~\ref{table:summary_of_real_datasets} 
summarizes each dataset. 

\begin{figure}[htbp]
    \centering
    \includegraphics[width=\textwidth]{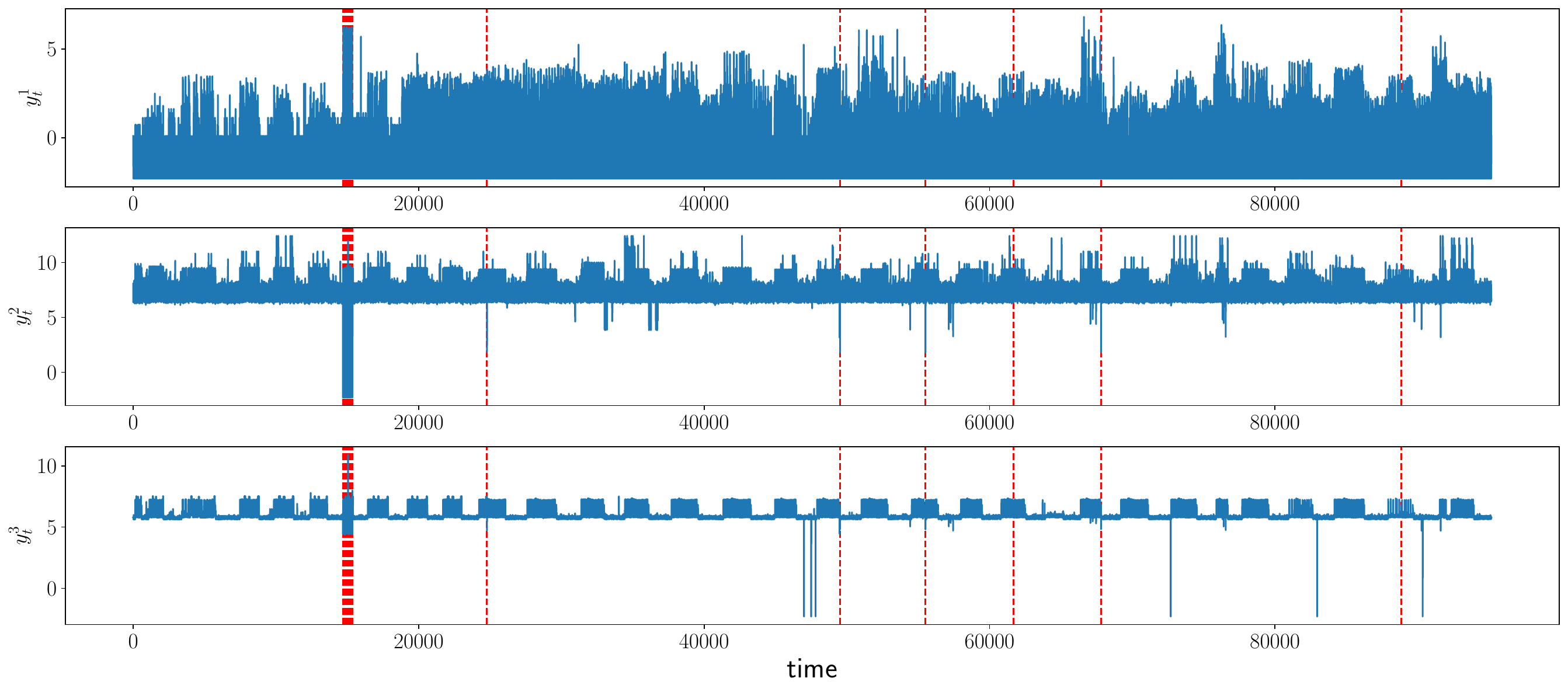}
    \caption{Plot of the SMTP dataset. The red dashed lines indicate anomalies. 
}
\label{fig:plot_smtp}
\end{figure}

Figure~\ref{fig:plot_smtp} 
shows 
the data stream of the SMTP dataset. 
We observe that:  
(i) Anomalies are concentrated around $t=15000$:  
$t=14692$, 
$14742$, 
$14789$, 
$14833$, 
$14888$, 
$14967$, 
$15016$, 
$15043$, 
$15099$,
$15165$, 
$15221$, 
$15283$, 
and $15366$. 
(ii) Although we see that 
all the variables deviate from 
these normal values 
at anomaly points around $t=15000$, 
it is not the case for anomalies after $t=49000$. 
Therefore, 
the mechanism of anomaly is 
also thought of as complicated. 
    
    \begin{table}[tb]
    \begin{center}
    \caption{Summary of the real datasets for anomaly detection.}
    \label{table:summary_of_real_datasets}
    {\tabcolsep=0.8\tabcolsep
    \begin{tabular}{rrr}
     \toprule
     \multicolumn{1}{c}{} & 
     \multicolumn{1}{c}{SMTP} &
     \multicolumn{1}{c}{THYROID} \\
     \midrule
     \multicolumn{1}{l}{Sequence length} & 95156 & 3772 \\
     \multicolumn{1}{l}{Number of attributes} & 3 & 6 \\
     \multicolumn{1}{l}{Number of outliers} & 30 & 93 \\
     \multicolumn{1}{l}{Ratio of outliers} & 0.03\% & 2.5\% \\
     \bottomrule
    \end{tabular}
    }
    \end{center}
    \end{table} 
    
Figure~\ref{fig:plot_smtp} 
shows 
the data stream of the THYROID dataset. 
We observe that 
some anomalies are visible. 
For example, 
the second variable $y_{t}^{2}$ deviates 
from normal values  
at 
$t=518$, $520$, $640$, $687$, 
$861$, and $965$. 
However, 
this observation is not applicable to all the anomalies. 
Therefore, 
the mechanism of anomaly is thought of as complicated. 
\begin{figure}[htbp]
\centering
\includegraphics[width=\textwidth]{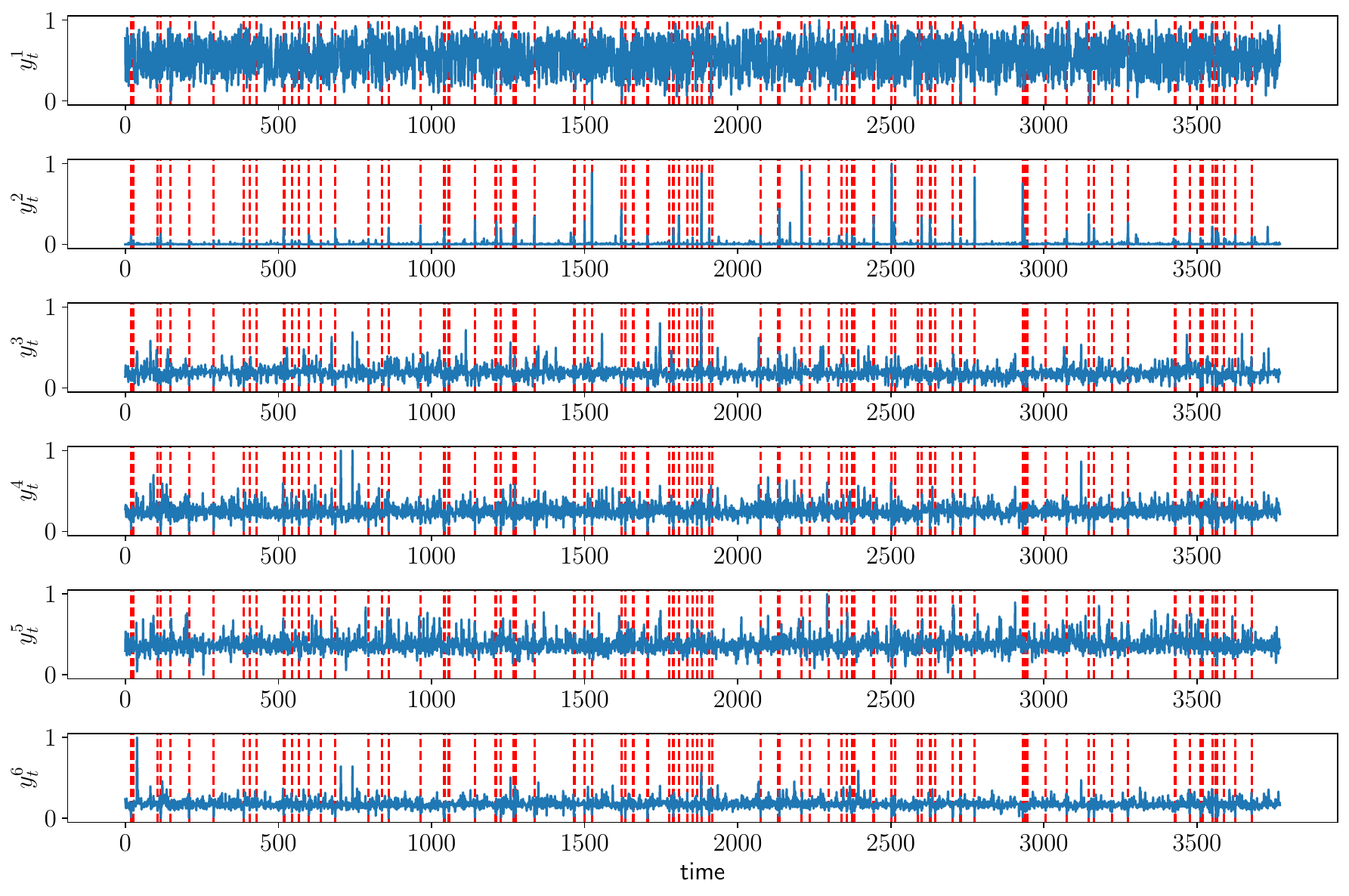}
\caption{Plot of THYROID dataset. 
         The red dashed lines indicate anomalies. 
         }
\label{fig:plot_thyroid}
\end{figure}

We used the first 40000 and 2000 data points 
as the training dataset for the SMTP and THYROID dataset, respectively, 
and the remaining data points as the test dataset. 
We chose SDEM~\citep{Yamanishi2004}, 
iEM~\citep{Neal1999}, 
sEM~\citep{Cappe2009}, 
ADWIN~\citep{Bifet2007}, 
KSWIN~\citep{Raab2020}, 
and PH~\citep{Page1954} 
for comparison.  
For SRA, SDEM, iEM, and sEM, 
each algorithm used GMM, 
and the number of components was selected 
among $K \in \{ 1, 2, 3 \}$.  

We calculated the anomaly score as 
$ s_{t} \mydef -\log{ f(x_{t}; \hat{\theta}_{t-1} )} $, 
where $\hat{\theta}_{t-1}$ is the parameter estimated at $t-1$. 
Then, we chose the best combinations with 
the AUC scores 
between $t=10000$ and $t=40000$ for SMTP 
and between $t=1000$ and $t=2000$ for THYROID. 
The AUC score was calculated based on the anomaly scores, 
and the ground truth labels associated with each data point, 
which indicate whether a data point is an anomaly or not. 
Note that the AUC score here is different from the one in Section~\ref{subsection:real_dataset_change_detection}. 

For both training datasets, 
we chose the hyperparameters among 
$\gamma \in \{ 5, 10, 15 \}$, 
$\beta = (d_{0}+1) / L(1-\alpha) \in \{ 0.1, 0.5, 1 \}$, 
and $M \in \{ 1, 5, 10 \}$ for SRA, 
$r_{\mathrm{SDEM}} \in \{ 0.1, 0.3, 0.5 \}$ for SMTP 
and $r_{\mathrm{SDEM}} \in \{ 0.01, 0.03, 0.05, 0.1 \}$ for THYROID, 
$r_{\mathrm{sEM}} \in \{ 0.01, 0.03, 0.05, 0.1 \} $ for SMTP and 
$r_{\mathrm{sEM}} \in \{ 0.0001, 0.001, 0.005 \}$ for THYROID, 
$\delta_{\mathrm{ADWIN}} \in \{ 0.001$, 
$0.002$, $0.005$, $0.01$, $0.02$, $0.05$, 
$0.1$, $0.2$, $0.5 \}$, 
$\alpha_{\mathrm{KSWIN}} \in \{ 0.001$, 
$0.005$, $0.01\}$, 
$w_{\mathrm{KSWIN}} \in \{ 0.001$, 
$0.005$, $0.01 \}$, 
$r_{\mathrm{KSWIN}} \in \{ 10$, $15\}$, 
and 
$\delta_{\mathrm{PH}} \in \{ 0.01$, $0.05$, 
$0.1$, $0.5$, $1$, $5$, $10 \}$.

As a result, the following parameters were chosen: 
for the SMTP dataset, 
$(\gamma$, $\beta$, $M$, $K) = (10$, $0.5$, $5$, $2)$ for SRA, 
$(r_{\mathrm{SDEM}}$, $K) = (0.3$, $3)$ for SDEM, 
$(r_{\mathrm{sEM}}$, $K) = (0.03$, $3)$ for sEM, 
$\delta_{\mathrm{ADWIN}} = 0.1$ and the second variable for ADWIN, 
$(\alpha_{\mathrm{KSWIN}}$, 
$w_{\mathrm{KSWIN}}$, 
$r_{\mathrm{KSWIN}})=(0.005$, 
$20$, 
$10)$ 
and the third variable for KSWIN, 
and 
all the values of hyperparameters 
and the third variable for PH. 
For the THYROID dataset, 
$(\gamma$, $\beta$, $M$, $K) = (10$, $0.5$, $5$, $3)$ for SRA, 
$(r_{\mathrm{SDEM}}$, $K) = (0.01$, $3)$ for SDEM, 
$(r_{\mathrm{sEM}}$, $K) = (0.001$, $3)$ for sEM, 
all the values of the hyperparameters and the first variable for ADWIN, 
$(\alpha_{\mathrm{KSWIN}}$, 
$w_{\mathrm{KSWIN}}$, 
$r_{\mathrm{KSWIN}})=(0.005$, 
$20$, 
$10)$ 
and the third variable for KSWIN, 
and 
all the values of the hyperparameters and 
the second variable for PH. 

To initialize the parameters or sufficient statistics, we drew 20 initial points for each algorithm. 
Each coordinate of the points was drawn from a uniform distribution with range set in the same way as in Section~\ref{subsection:real_dataset_change_detection}. 
We repeated this procedure 10 times and selected the combination of parameters that yielded the best performance on average. 

Table~\ref{table:auc_scores_for_smtp_dataset} shows 
the AUC scores on the test datasets 
for SMTP and THYROID. 
We observe that SRA outperforms 
other online learning algorithms 
and 
concept drift detection algorithms 
in anomaly detection. 

\begin{table}[tb]
\begin{center}
\caption{AUC scores on the SMTP dataset and THYROID dataset.}
\label{table:auc_scores_for_smtp_dataset}
{\tabcolsep=0.7\tabcolsep
\begin{tabular}{rrr}
 \toprule
 & \multicolumn{1}{c}{SMTP} & \multicolumn{1}{c}{THYROID} \\
 \midrule
 \multicolumn{1}{c}{SRA} & $\mathbf{0.874 \pm 0.001}$ & $\mathbf{0.972 \pm 0.000}$ \\
 \multicolumn{1}{c}{SDEM} & $0.773 \pm 0.000$ & $0.933 \pm 0.000$ \\
 \multicolumn{1}{c}{iEM} &  $0.744 \pm 0.000$ & $0.935 \pm 0.000$ \\
 \multicolumn{1}{c}{sEM} & $0.773 \pm 0.000$ & $0.968 \pm 0.000$ \\
 \multicolumn{1}{c}{ADWIN} & $0.498 \pm 0.000$ & $0.500 \pm 0.000$ \\
 \multicolumn{1}{c}{KSWIN} & $0.493 \pm 0.051$ & $0.478 \pm 0.027$  \\
 \multicolumn{1}{c}{PH} & $0.424 \pm 0.000$ & $0.709 \pm 0.000$ \\
 \bottomrule
\end{tabular}
}
\end{center}
\end{table}

\subsection{Concluding Remarks}
\label{subsection:concluding_remarks_experiments}

In Section~
\ref{subsection:univariate_synthetic_datasets} 
and 
\ref{subsection:multivariate_synthetic_datasets}, 
we confirmed that 
SRA is robust to outliers 
and adaptive to changes, 
in comparison with the other 
online learning algorithms 
and 
concept drift detection algorithms, 
for both synthetic univariate and multivariate datasets 
with abrupt and gradual changes. 
SRA outperformed the other online learning algorithms 
with MSE. 
SRA also outperformed the online learning algorithms 
and concept drift detection algorithms 
with AUC. 
However, 
we also confirmed that SRA took more total time 
to tune hyperparameters than 
the other algorihms 
because SRA has more hyperparameters. 
SRA also took more time for a set of hyperparameters 
because SRA requires the calculation of 
probability densities of parametric distributions 
in calculating the change scores. 

In 
Section~\ref{subsection:real_dataset_change_detection}, 
we confirmed that 
SRA also worked for real datasets 
in change detection. 
For both univariate and multivariate datasets, 
SRA was better than the other algorithms 
with AUC. 
We also visualized estimated parameters and 
change scores 
and 
concluded that 
SRA was excellent in both robustness and adaptivity, 
whereas 
other online learning algorithms 
either overfitted the changes 
or were not adaptive to changes. 

In Section~\ref{subsection:real_dataset_anomaly_detection}, 
we confirmed that 
SRA was also superior to the other online learning algorithms 
and 
concept drift detection algorithms 
in anomaly detection. 

In summary, 
SRA is excellent in 
estimation of parameters, 
change detection,
and anomaly detection, 
compared to the other 
online learning algorithms 
and 
concept drift detection algorithms. 
SRA has a sound theoretical background 
for quantitatively characterizing the tradeoff 
between the robustness and adaptivity. 
However, 
SRA took much more time to tune hyperparameters. 
Furthermore, 
although SRA is formulated based on datasets with 
abrupt changes, 
empirical studies show that it also works 
on datasets with gradual changes 
for both the synthetic datasets in 
Section~\ref{subsubsection:result_comparison_with_other_algorithms_synthetic_1D} 
and real datasets 
in Section~\ref{subsubsection:real_dataset_change_detection_skab}. 
Future work 
will verify the theoretical performance of SRA  
on datasets with gradual changes.

%% file: 6_Conclusion.tex
\section{Conclusion}

In this study, 
we quantitatively evaluate the tradeoff between  
the robustness and 
adaptivity of online learning algorithms. 
We proposed a novel algorithm, 
called SRA, to consider this tradeoff. 
SRA updates 
parameters of distribution or sufficient statistics in an online fashion, 
using the SA scheme \citep{Robbins1951}. 
No update is done when the norm of the stochastic update exceeds the threshold. 
We showed the upper bound of the expectation of 
the mean field of the stochastic update in the SA scheme. 
We further explicitly derived the relation between two parameters: 
1) step size of the SA scheme 
and 2) threshold parameter of the stochastic update. 
The empirical experiments for the synthetic datasets 
demonstrated that 
the dependencies on the parameters of SRA 
are consistent with the theoretical analysis, 
and that SRA is superior to previous 
online learning algorithms 
and 
concept drift detection algorithms. 
The experiments on real datasets 
also demonstrated that SRA is superior to other algorithms 
in change detection and anomaly detection. 

Future work includes 
extension of SRA to gradual changes of parameters. 
Another interesting line of research lies in 
a theoretical analysis as well as empirical experiments 
in a setting where the step size is determined adaptively 
and reset after a change point.

%% file: A_Proofs.tex
\section{ Proofs for Section~\ref{section:convergence_analysis_of_bounded_stochastic_update}}  \label{section:proofs_in_online_robust_and_adaptive_learning_from_data_streams}

In this section, 
we describe all the proofs for Section~\ref{section:convergence_analysis_of_bounded_stochastic_update}. 

\subsection{ Proof of Lemma~\ref{lemma:inequality_expectation_lyapunov_function_gradient_noise_vector} }
\label{subsection:proof_lemma_inequality_expectation_lyapunov_function_gradient_noise_vector}

\begin{proof}
It is easily shown that 
\begin{align}
\mathbb{E}[ 
  -\langle
    \nabla V(\theta_{k}) | \xi_{k+1} 
  \rangle
  \mid \mathcal{F}_{k}
] 
&= \mathbb{E}[
  -\langle
    \nabla V(\theta_{k}) | 
    G_{\theta_{k}}(Y_{k+1}) - h(\theta_{k})
  \rangle
  \mid \mathcal{F}_{k}
] \\
&= \mathbb{E}[
  -\langle
    \nabla V(\theta_{k}) | 
    G_{\theta_{k}}(Y_{k+1}) - \mathbb{E} [ H_{\theta_{k}}(Y_{k+1}) \mid \mathcal{F}_{k} ]
  \rangle  
  \mid \mathcal{F}_{k} 
]  \\
&\leq [
  \langle 
    \nabla V(\theta_{k}) | 
    \mathbb{E} [ \| G_{\theta_{k}}(Y_{k+1}) - H_{\theta_{k}}(Y_{k+1}) \| ]
  \rangle
  \mid \mathcal{F}_{k}
]  \\
&\leq 
  \| \nabla V(\theta_{k}) \| 
  \mathbb{E} [ 
    \| G_{\theta_{k}}(Y_{k+1}) - H_{\theta_{k}}(Y_{k+1}) \|
  \mid \mathcal{F}_{k}
] \\
&\leq
  \| \nabla V(\theta_{k}) \|
  \int_{0}^{\infty}
    P[ \| G_{\theta_{k}}(Y_{k+1}) - H_{\theta_{k}}(Y_{k+1}) \| \geq z ] \, 
    \d{z}
\\
&\leq 
  \| \nabla V(\theta_{k}) \|
  \int_{\gamma}^{\infty} 
    P[ \| G_{\theta_{k}}(Y_{k+1}) - H_{\theta_{k}}(Y_{k+1}) \| \geq z ] \, 
    \d{z}
  \\
&=
  \| \nabla V(\theta_{k}) \|
  \int_{\gamma}^{\infty} 
    P[ \| H_{\theta_{k}}(Y_{k+1}) \| \geq z ] \, 
    \d{z}. 
\end{align}
When $\mathbb{E}[ \| e_{k+1} \|^{2} | \mathcal{F}_{k} ] < \infty$ holds, 
there exists $M > 0$ such that 
the following inequality holds:
\begin{align}
P[ \| H_{\theta_{k}}(Y_{k+1}) \| \geq z ] \leq \exp{ \left( -\frac{z^{2}}{M^{2}} \right) }. 
\end{align}
Then, we have 
\begin{align} 
\mathbb{E} [ 
  -\langle 
    \nabla V(\theta_{k}) | \xi_{k+1} 
  \rangle \mid \mathcal{F}_{k}
] &\leq 
  \| \nabla V(\theta_{k}) \|
  \int_{\gamma}^{\infty} 
    \exp{ \left( -\frac{z^{2}}{M^{2}} \right) } \, \d{z}.
\end{align}
\end{proof}

\subsection{ Proof of Lemma~\ref{lemma:inequality_expectation_square_norm_noise_vector}}
\label{subsection:proof_of_lemma_inequality_expectation_square_norm_noise_vector}

\begin{proof}
First, 
when $Y_{k+1} \sim f(Y_{k+1}; \theta_{k})$, 
we have
\begin{align}
\mathbb{E} [ \| \xi_{k+1} \|^{2} | \mathcal{F}_{k} ]  
&= P ( \| H_{\theta_{k}}(Y_{k+1}) \| \geq \gamma) 
   \, \mathbb{E}[ \| h (\theta_{k}) \|^{2}
                   \mid 
                   \mathcal{F}_{k}, 
                   \| H_{\theta_{k}}(Y_{k+1}) \| \geq \gamma ]  \\
&\quad +P ( \| H_{\theta_{k}}(Y_{k+1}) \| < \gamma) 
    \mathbb{E}[ \| H_{\theta_{k}}(Y_{k+1}) - h (\theta_{k}) \|^{2}
                   \mid 
                   \mathcal{F}_{k}, 
                   \| H_{\theta_{k}}(Y_{k+1}) \| < \gamma ]  \\
&\leq P (\| H_{\theta_{k}}(Y_{k+1}) \| \geq \gamma) 
      \, \mathbb{E} [ \| h(\theta_{k}) \|^{2} \mid \mathcal{F}_{k} ]  \\
&\quad 
      +P (\| H_{\theta_{k}}(Y_{k+1}) \| < \gamma) \,
         (\sigma_{0}^{2} + 
          \sigma_{1}^{2} \mathbb{E} [ 
            \| h(\theta_{k}) \|^{2}  
            \mid \mathcal{F}_{k} 
          ] )  \\
&\leq 
 \sigma_{0}^{2} 
 + (\sigma_{1}^{2} + 1) 
   \mathbb{E}[ \| h(\theta_{k}) \|^{2} \mid \mathcal{F}_{k} ]. 
\label{eq:L_smooth_lyapunov_function_expectation_noise}
\end{align}
Next, when $Y_{k+1} \sim f_{\mathrm{noise}}$, 
we have
\begin{align}
\mathbb{E}[ \| \xi_{k+1} \|^{2}  \mid \mathcal{F}_{k} ]
\leq \min( dU^{2}, \gamma^{2}). 
\end{align}
Then, we have
\begin{align}
\mathbb{E} [ \| \xi_{k+1} \|^{2} \mid \mathcal{F}_{k} ] 
&\leq 
 \alpha \left\{
   \sigma_{0}^{2} + 
   (\sigma_{1}^{2} + 1) 
   \mathbb{E} [ \| h(\theta_{k}) \|^{2} | \mathcal{F}_{k} ]
 \right\} 
 + (1-\alpha) \min(dU^{2}, \gamma^{2}). 
\end{align}

\end{proof}

\subsection{ Proof of Theorem~\ref{theorem:tradeoff_between_discounting_factor_and_gradient_of_stochastic_update}}
\label{section:proof_of_theorem_tradeoff_between_discounting_factor_and_gradient_of_stochastic_update}

\begin{proof}
As the Lyapunov function $V(\theta)$ is $L$-smooth,
we obtain
\begin{align}
V(\theta_{k+1})
&\leq V(\theta_{k})
      - \rho_{k+1}
        \langle
          \nabla V(\theta_{k}) | G_{\theta_{k}} (Y_{k+1})
        \rangle
      + \frac{ L \rho_{k+1}^{2} }{2}
         \| G_{ \theta_{k} } (Y_{k+1}) \|^{2} \\
&= V(\theta_{k})
   -\rho_{k+1}
    \langle
      \nabla V(\theta_{k}) |
      h({\theta_{k}}) + \xi_{k+1}
    \rangle \\
&\quad
    +\frac{ L \rho_{k+1}^{2} }{2}
    ( \| h(\theta_{k}) \|^{2}
      + 2 \langle h(\theta_{k}) | \xi_{k+1} \rangle
      + \| \xi_{k+1} \|^{2} )  \\
&\leq V(\theta_{k})
   -\rho_{k+1}
    \langle
      \nabla V(\theta_{k}) | h(\theta_{k}) + \xi_{k+1}
    \rangle
   +L \rho_{k+1}^{2} ( \| h(\theta_{k}) \|^{2} +
                       \| \xi_{k+1} \|^{2}
    ).
\label{eq:L_smooth_lyapunov_function}
\end{align}
The equality in the last equation in 
Equation~\eqref{eq:L_smooth_lyapunov_function}
holds when $\xi_{k+1} = h(\theta_{k})$.
Rearranging terms yields
\begin{align}
\rho_{k+1} \langle
  \nabla V(\theta_{k}) | h(\theta_{k})
\rangle 
&\leq V(\theta_{k}) - V(\theta_{k+1})
      -\rho_{k+1} \langle
         \nabla V(\theta_{k}) | \xi_{k+1}
      \rangle 
      +L \rho_{k+1}^{2} ( \| h(\theta_{k}) \|^{2} +
                       \| \xi_{k+1} \|^{2}
    ).
\end{align}
As
$\langle
   \nabla V(\theta_{k}) | h(\theta_{k})
 \rangle
 \geq \frac{1}{c_{1}} ( \| h(\theta_{k}) \|^{2} - c_{0})
$,
we get
\begin{align}
\frac{ \rho_{k+1} }{ c_{1} }
( \| h(\theta_{k}) \|^{2} - c_{0})
&\leq
 V(\theta_{k}) - V(\theta_{k+1})  \\
&\quad
 -\rho_{k+1} \langle
    \nabla V(\theta_{k}) | \xi_{k+1}
  \rangle
 +L \rho_{k+1}^{2} ( \| h(\theta_{k}) \|^{2} +
                     \| \xi_{k+1} \|^{2}
  )  \\
\Longleftrightarrow
\frac{ \rho_{k+1} }{ c_{1} } (1 - c_{1} L \rho_{k+1})
\| h(\theta_{k}) \|^{2}
&\leq
 \frac{ c_{0} }{ c_{1} } \rho_{k+1}
 + V(\theta_{k}) - V(\theta_{k+1})  \\
&\quad
 -\rho_{k+1} \langle \nabla V(\theta_{k}) | \xi_{k+1} \rangle
 + L \rho_{k+1}^{2} \| \xi_{k+1} \|^{2}.
\label{eq:L_smooth_lyapunov_function_transformed}
\end{align}
Let us sum up both sides in
Equation~\eqref{eq:L_smooth_lyapunov_function_transformed}
from $k=0$ to $k=n$
and rearrange terms,
then we get
\begin{align}
\sum_{k=0}^{n}
  \frac{ \rho_{k+1}}{ c_{1} }
  (1 - c_{1} L \rho_{k+1}) \| h(\theta_{k}) \|^{2}
&\leq
  \frac{ c_{0} }{ c_{1} } \sum_{k=0}^{n} \rho_{k+1}
  + V(\theta_{0}) - V(\theta_{n+1})  \\
&\quad -\sum_{k=0}^{n}
     \rho_{k+1} \langle
       \nabla V(\theta_{k}) | \xi_{k+1}
     \rangle   
     + L \sum_{k=0}^{n}
       \rho_{k+1}^{2} \| \xi_{k+1} \|^{2}.
\label{eq:L_smooth_lyapunov_function_sumup}
\end{align}
Taking expectation in both sides of 
Equation~\eqref{eq:L_smooth_lyapunov_function_sumup}
gives
\begin{align}
\sum_{k=0}^{n}
  \frac{ \rho_{k+1} }{c_{1}}
  (1 - c_{1} L \rho_{k+1})
  \mathbb{E} [ \| h(\theta_{k}) \|^{2} | \mathcal{F}_{k} ]
&\leq
 \frac{ c_{0} }{ c_{1} } \sum_{k=0}^{n} \rho_{k+1}
 + V_{0, n}  \\
&\quad - \sum_{k=0}^{n}
     \rho_{k+1}
       \mathbb{E} [
         \langle
           \nabla V(\theta_{k}) | \xi_{k+1}
         \rangle  \mid \mathcal{F}_{k}
       ]  \\
&\quad + L \sum_{k=0}^{n} \rho_{k+1}^{2}
    \mathbb{E} [ \| \xi_{k+1} \|^{2} \mid \mathcal{F}_{k} ].
\label{eq:L_smooth_lyapunov_function_expectation}
\end{align}
Substituting
Equation~\eqref{eq:inequality_expectation_lyapunov_function_gradient_noise_vector}
and 
\eqref{eq:inequality_expectation_square_norm_noise_vector}
to the second and third term in
Equation~\eqref{eq:L_smooth_lyapunov_function_expectation},
we have the following inequality:
\begin{align}
&\sum_{k=0}^{n}
  \frac{ \rho_{k+1} }{c_{1}}
  (1 - c_{1} L \rho_{k+1} )
  \mathbb{E}[ \| h(\theta_{k}) \|^{2} | \mathcal{F}_{k} ] \\
&\leq
 \frac{ c_{0} }{ c_{1} } \sum_{k=0}^{n} \rho_{k+1}
 + V_{0, n}
 + \sum_{k=0}^{n}
     \rho_{k} \| \nabla V(\theta_{k}) \|
     \int_{\gamma}^{\infty}
       \exp{ \left( -\frac{z^{2}}{M^{2}} \right) } \, \d{z} \\
&\quad
 +L \sum_{k=0}^{n}
   \rho_{k+1}^{2}
   \left\{
     \alpha
     ( \sigma_{0}^{2} +
       (\sigma_{1}^{2} + 1)
       \mathbb{E} [ \| h(\theta_{k}) \|^{2} \mid \mathcal{F}_{k} ]
     )
     +
     (1-\alpha) \min(dU^{2}, \gamma^{2})
   \right\}  \\
&\leq
  \frac{ c_{0} }{ c_{1} } \sum_{k=0}^{n} \rho_{k+1}
 + V_{0, n}
 + \sum_{k=0}^{n}
     \rho_{k} (d_{0} + d_{1} \| h(\theta_{k}) \|)
     \int_{\gamma}^{\infty}
       \exp{ \left(-\frac{z^{2}}{M^{2}} \right) } \, \d{z} \\
&\quad
 +L \sum_{k=0}^{n}
   \rho_{k+1}^{2}
   \left\{
     \alpha
     ( \sigma_{0}^{2} +
       (\sigma_{1}^{2} + 1)
       \mathbb{E} [ \| h(\theta_{k}) \|^{2} \mid \mathcal{F}_{k} ]
     )
     +
     (1-\alpha) \min(dU^{2}, \gamma^{2})
   \right\}  \\
&\leq
  \frac{ c_{0} }{ c_{1} } \sum_{k=0}^{n} \rho_{k+1}
 + V_{0, n}
 + \sum_{k=0}^{n}
     \rho_{k+1}
     \left(
       d_{0} + d_{1} (\| h(\theta_{k}) \|^{2} + 1)
     \right)
     \int_{\gamma}^{\infty}
       \exp{ \left( -\frac{ z^{2} }{ M^{2} } \right) } \, \d{z} \\
&\quad
 +L \sum_{k=0}^{n}
   \rho_{k+1}^{2}
   \left\{
     \alpha
     ( \sigma_{0}^{2} +
       (\sigma_{1}^{2} + 1)
       \mathbb{E} [ \| h(\theta_{k}) \|^{2} \mid \mathcal{F}_{k} ]
     )
     +(1-\alpha) \min(dU^{2}, \gamma^{2})
   \right\}.
\end{align}
As a result, we have
\begin{align}
&\sum_{k=0}^{n}
  \frac{ \rho_{k+1} }{c_{1}}
  \left\{
    1 -
      c_{1}
        d_{1} \int_{\gamma}^{\infty}
        \exp{ \left( -\frac{ z^{2} }{ M^{2} } \right) } \, \d{z}
    -c_{1} L (\sigma_{1}^{2}+2) \rho_{k+1}
  \right\}
  \mathbb{E}[ \| h(\theta_{k}) \|^{2} \mid \mathcal{F}_{k} ]  \\
&\leq
 V_{0, n}
 + \left(
   \frac{c_{0}}{c_{1}}
   +
   (d_{0} + 1)
   \int_{\gamma}^{\infty}
     \exp{ \left( -\frac{z^{2}}{M^{2}} \right) } \, \d{z}
 \right)
 \sum_{k=0}^{n} \rho_{k+1} \\
&\quad
 + L
   (\alpha \sigma_{0}^{2} + (1-\alpha) \min( dU^{2}, \gamma^{2}) )
   \sum_{k=0}^{n} \rho_{k+1}^{2}.
\label{eq:inequality_expected_mean_field}
\end{align}
When $\rho_{k+1}$ satisfies
\begin{align}
\rho_{k+1}
  < \frac{
      1 - 2 c_{1} d_{1} \int_{\gamma}^{\infty}
      \exp{
        \left(
          -\frac{z^{2}}{M^{2}}
        \right)
      } \d{z}
    }{
      2 c_{1} L (\sigma_{1}^{2} + 2)
    },
\end{align}
then
\begin{align}
\mathbb{E}[ \| h(\theta_{N}) \|^{2} ]
&= \frac{
     \sum_{k=0}^{n}
       \rho_{k+1} \mathbb{E}[ \| h(\theta_{k}) \|^{2} | \mathcal{F}_{k}]
   }{
     \sum_{k=0}^{n} \rho_{k+1}
   }  \\
&\leq
  2 c_{0}
  +
  2 c_{1} (d_{0} + 1)
  \int_{\gamma}^{\infty}
    \exp{ \left( - \frac{z^{2}}{M^{2}} \right) } \d{z} \\
&\quad
    + 2 c_{1} \frac{
        V_{0, n}
        +
        L ( \alpha \sigma_{0}^{2} +
          (1-\alpha) \min(dU^{2}, \gamma^{2})
        )
        \sum_{k=0}^{n} \rho_{k+1}^{2}
    }{
      \sum_{k=0}^{n} \rho_{k+1}
    }.
\end{align}

\end{proof}

\subsection{ Proof of Corollary~\ref{corollary:rho_determined_by_gamma}}
\label{subsection:corollary_rho_determined_by_gamma} 

\begin{proof}
When $\rho_{k} = \rho = \mathrm{const.}$, 
the right hand side of Equation~\eqref{eq:expectation_mean_field_upper_bound} is written as a function of 
$\rho$ and $\gamma$ given the other variables as 
\begin{align}
b(\rho, \gamma; \alpha, \sigma_{0}, c_{0}, c_{1}, d_{1}, U)
&= 2 \left(
     c_{0} + 
     c_{1} (d_{0} + 1) \int_{\gamma}^{\infty} 
       \exp{ \left(- \frac{z^{2}}{M^{2}}\right) } \d{z}
   \right)  \\
&\quad  + 2 c_{1} 
     \frac{ V_{0, n} + L( \alpha \sigma_{0}^{2} + (1-\alpha) \min(dU^{2}, \gamma^{2})) (n+1) \rho^{2}
     }{
        (n+1) \rho
     }. 
\end{align}

When $\gamma^{2} < dU^{2}$, 
$b=b(\rho, \gamma; \alpha, \sigma_{0}, c_{0}, c_{1}, d_{1}, U)$ is
 minimized if the following equations hold:
\begin{align}
\frac{ \partial b}{ \partial \gamma} 
&= -2c_{1} ( d_{0} + 1) \exp{\left ( -\frac{\gamma^{2} }{ M^{2}} \right) } 
   + 2(1-\alpha) \gamma \rho = 0, \label{eq:deriv_b_gamma} \\
\frac{ \partial b}{ \partial \rho }
&= -\frac{ 2c_{1} V_{0,n}}{(n+1) \rho^{2}}
   + (1-\alpha) \gamma^{2} = 0.  
\end{align}
Then, 
from Equation~\eqref{eq:deriv_b_gamma}, 
we have the following equation:
\begin{align}
\rho = \frac{ 
         c_{1} (d_{0} + 1) \exp{ \left(-\frac{\gamma^{2}}{M^{2}}\right) } 
       }{ 
         2L(1-\alpha) \gamma
       }. 
\end{align} 
\end{proof}

\subsection{ Proof of Corollary~\ref{corollary:tradeoff_between_discounting_factor_and_gradient_of_stochastic_update_gamma_infty}} 
\label{subsection:proof_corollary_tradeoff_between_discounting_factor_and_gradient_of_stochastic_update_gamma_infty} 

\begin{proof}
We easily obtain
\begin{align}
\lim_{\gamma \rightarrow \infty} 
  \mathbb{E} [ \| h(\theta_{N}) \|^{2} ] 
&\leq 2c_{0} + \frac{ 2c_{1} V_{0,n}}{\rho(n+1)} + 
      2c_{1} \rho L \alpha \sigma_{0}^{2} \\
&\quad + \lim_{\gamma \rightarrow \infty} \left\{ 
           2c_{1} (d_{0} + 1) \int_{\gamma}^{\infty} 
             \exp{ \left(-\frac{z^{2}}{M^{2}} \right) } \d{z} 
           + 2c_{1} \rho L (1-\alpha) \min(dU^{2}, \gamma^{2})
         \right\} \\
&= 2c_{0} + \frac{ 2c_{1} V_{0,n} }{ \rho(n+1) }
   + 2c_{1} \rho L (\alpha \sigma_{0}^{2}  + (1-\alpha) dU^{2} ). 
\end{align}
\end{proof}

\subsection{ Proof of Corollary~\ref{corollary:difference_between_upper_bounds} }
\label{subsection:corollary_difference_between_upper_bounds}

\begin{proof}
The difference between the upper bounds in Equation~\eqref{eq:tradeoff_between_discounting_factor_and_gradient_of_stochastic_update_gamma_infty} 
and \eqref{eq:expectation_mean_field_upper_bound} 
is easily calculated as 
\begin{align}
g(\gamma)
&= 
   2c_{0}
   +2c_{1} \frac{ 
     V_{0,n} + 
     L(\alpha \sigma_{0}^{2} + (1-\alpha) dU^{2}) 
     \sum_{k=0}^{n} \rho_{k+1}^{2} 
   }{ \sum_{k=0}^{n} \rho_{k+1} }  \\
&\quad 
   -2 \left(
     c_{0} + c_{1} (d_{0} + 1) \int_{\gamma}^{\infty}
       \exp{ \left(-\frac{z^{2}}{M^{2}} \right) } \d{z}
   \right)  \\
&\quad
   -2 c_{1} 
   \frac{ V_{0,n} + 
          L(\alpha \sigma_{0}^{2} + (1-\alpha) \min(dU^{2}, \gamma^{2}) ) \sum_{k=0}^{n} \rho_{k+1}^{2} 
   }{ \sum_{k=0}^{n} \rho_{k+1} } \\
&= -2c_{1} (d_{0}+1) 
   \int_{\gamma}^{\infty} 
     \exp{ \left( -\frac{z^{2}}{M^{2}} \right) } \, \d{z}
   +2c_{1} \frac{ L(1-\alpha) \max(dU^{2} - \gamma^{2}, 0) \sum_{k=0}^{n} \rho_{k+1}^{2} }{ \sum_{k=0}^{n} \rho_{k+1} }. 
\end{align}
\end{proof}

%% file: B_Dependency_on_Hyper_Parameters.tex
\section{Dependency on Hyperparameters}
\label{section:dependency_on_hyperparameters}

In this section, 
we show detailed results of dependencies on hyperparameters 
for SRA 
in the experiment described in Section~\ref{subsubsection:result_comparison_with_other_algorithms_synthetic_1D}.

Table~\ref{table:dependency_of_auc_on_hyperparameters_for_sra} 
shows average AUCs of SRA for synthetic univariate datasets with abrupt and gradual changes, 
varying the hyperparameters $\gamma$, $\beta$, and $M$. 
We observe that the AUCs of SRA are dependent on 
the values of the hyperparamters. 
Therefore, 
we conclude that the values of the hyperparameters should be carefully chosen. 

\renewcommand{\arraystretch}{0.9}

\begin{table}[tb]
\begin{footnotesize}
\caption{Average AUCs of SRA for synthetic univariate datasets with abrupt and gradual changes. $\gamma$, $\beta$, and $M$ are the hyperparameters of SRA. }
\label{table:dependency_of_auc_on_hyperparameters_for_sra}
\begin{center}
\begin{tabular}{lllrr}
\toprule
\multicolumn{1}{c}{$\gamma$} & 
\multicolumn{1}{c}{$\beta$} & 
\multicolumn{1}{c}{$M$} & 
\multicolumn{1}{c}{Abrupt Change} &  
\multicolumn{1}{c}{Gradual Change} \\
\midrule
1  & 0.1 & 5  & $0.715 \pm 0.080$ & $0.616 \pm 0.046$ \\
   &     & 10 & $0.629 \pm 0.171$ & $0.623 \pm 0.067$ \\
   &     & 15 & $0.657 \pm 0.126$ & $0.639 \pm 0.104$ \\
   & 0.2 & 5  & $0.663 \pm 0.033$ & $0.633 \pm 0.043$ \\
   &     & 10 & $0.660 \pm 0.035$ & $0.643 \pm 0.017$ \\
   &     & 15 & $0.660 \pm 0.034$ & $0.642 \pm 0.017$ \\
   & 0.3 & 5  & $0.609 \pm 0.029$ & $0.644 \pm 0.034$ \\
   &     & 10 & $0.615 \pm 0.044$ & $0.665 \pm 0.060$ \\
   &     & 15 & $0.618 \pm 0.044$ & $0.665 \pm 0.061$ \\
   & 0.5 & 5  & $0.574 \pm 0.035$ & $\mathbf{0.702 \pm 0.023}$ \\
   &     & 10 & $0.569 \pm 0.032$ & $0.689 \pm 0.029$ \\
   &     & 15 & $0.572 \pm 0.033$ & $0.697 \pm 0.025$ \\
3  & 0.1 & 5  & $0.513 \pm 0.044$ & $0.422 \pm 0.217$ \\
   &     & 10 & $\mathbf{0.717 \pm 0.021}$ & $0.518 \pm 0.234$ \\
   &     & 15 & $0.496 \pm 0.042$ & $0.463 \pm 0.240$ \\
   & 0.2 & 5  & $0.667 \pm 0.155$ & $0.561 \pm 0.229$ \\
   &     & 10 & $0.622 \pm 0.068$ & $0.491 \pm 0.119$ \\
   &     & 15 & $0.633 \pm 0.019$ & $0.487 \pm 0.073$ \\
   & 0.3 & 5  & $0.594 \pm 0.096$ & $0.475 \pm 0.066$ \\
   &     & 10 & $0.615 \pm 0.023$ & $0.491 \pm 0.054$ \\
   &     & 15 & $0.614 \pm 0.028$ & $0.486 \pm 0.053$ \\
   & 0.5 & 5  & $0.598 \pm 0.034$ & $0.497 \pm 0.053$ \\
   &     & 10 & $0.585 \pm 0.022$ & $0.503 \pm 0.051$ \\
   &     & 15 & $0.583 \pm 0.022$ & $0.500 \pm 0.052$ \\
5  & 0.1 & 5  & $0.538 \pm 0.057$ & $0.489 \pm 0.221$ \\
   &     & 10 & $0.541 \pm 0.196$ & $0.426 \pm 0.225$ \\
   &     & 15 & $0.626 \pm 0.174$ & $0.519 \pm 0.191$ \\
   & 0.2 & 5  & $0.562 \pm 0.207$ & $0.418 \pm 0.232$ \\
   &     & 10 & $0.580 \pm 0.178$ & $0.492 \pm 0.289$ \\
   &     & 15 & $0.591 \pm 0.153$ & $0.343 \pm 0.106$ \\
   & 0.3 & 5  & $0.522 \pm 0.260$ & $0.418 \pm 0.226$ \\
   &     & 10 & $0.638 \pm 0.041$ & $0.421 \pm 0.087$ \\
   &     & 15 & $0.607 \pm 0.054$ & $0.438 \pm 0.095$ \\
   & 0.5 & 5  & $0.561 \pm 0.199$ & $0.528 \pm 0.166$ \\
   &     & 10 & $0.626 \pm 0.052$ & $0.505 \pm 0.099$ \\
   &     & 15 & $0.603 \pm 0.020$ & $0.480 \pm 0.072$ \\
10 & 0.1 & 5  & $0.537 \pm 0.115$ & $0.551 \pm 0.143$ \\
   &     & 10 & $0.607 \pm 0.109$ & $0.334 \pm 0.137$ \\
   &     & 15 & $0.532 \pm 0.171$ & $0.399 \pm 0.215$ \\
   & 0.2 & 5  & $0.486 \pm 0.048$ & $0.522 \pm 0.139$ \\
   &     & 10 & $0.530 \pm 0.213$ & $0.290 \pm 0.088$ \\
   &     & 15 & $0.582 \pm 0.151$ & $0.409 \pm 0.190$ \\
   & 0.3 & 5  & $0.554 \pm 0.218$ & $0.412 \pm 0.136$ \\
   &     & 10 & $0.596 \pm 0.173$ & $0.374 \pm 0.163$ \\
   &     & 15 & $0.602 \pm 0.136$ & $0.402 \pm 0.179$ \\
   & 0.5 & 5  & $0.421 \pm 0.146$ & $0.432 \pm 0.081$ \\
   &     & 10 & $0.673 \pm 0.123$ & $0.451 \pm 0.193$ \\
   &     & 15 & $0.674 \pm 0.080$ & $0.512 \pm 0.153$ \\
15 & 0.1 & 5  & $0.477 \pm 0.106$ & $0.402 \pm 0.138$ \\
   &     & 10 & $0.558 \pm 0.152$ & $0.557 \pm 0.149$ \\
   &     & 15 & $0.563 \pm 0.109$ & $0.278 \pm 0.024$ \\
   & 0.2 & 5  & $0.512 \pm 0.054$ & $0.470 \pm 0.146$ \\
   &     & 10 & $0.533 \pm 0.164$ & $0.389 \pm 0.167$ \\
   &     & 15 & $0.632 \pm 0.079$ & $0.379 \pm 0.207$ \\
   & 0.3 & 5  & $0.544 \pm 0.175$ & $0.457 \pm 0.263$ \\
   &     & 10 & $0.555 \pm 0.130$ & $0.346 \pm 0.137$ \\
   &     & 15 & $0.584 \pm 0.019$ & $0.283 \pm 0.017$ \\
   & 0.5 & 5  & $0.478 \pm 0.083$ & $0.422 \pm 0.087$ \\
   &     & 10 & $0.621 \pm 0.089$ & $0.330 \pm 0.192$ \\
   &     & 15 & $0.577 \pm 0.017$ & $0.318 \pm 0.107$ \\
\bottomrule
\end{tabular}
\end{center}
\end{footnotesize}
\end{table}